\documentclass[10pt,journal,compsoc]{IEEEtran}
\usepackage{amsmath,amsfonts}
\usepackage{algorithmic}
\usepackage{algorithm}
\usepackage{array}
\usepackage[caption=false,font=normalsize,labelfont=sf,textfont=sf]{subfig}
\usepackage{textcomp}
\usepackage{stfloats}
\usepackage{url}
\usepackage{verbatim}
\usepackage{graphicx}
\graphicspath{{./}}
\usepackage{cite}
\newcommand\eg{\textit{e.g.}}
\newcommand\ie{\textit{i.e.}}
\usepackage{multirow}
\usepackage[table]{xcolor}
\usepackage{booktabs}
\usepackage{tabularx}
\usepackage{adjustbox}
\usepackage{placeins}
\definecolor{color3}{HTML}{F0F0F0}
\definecolor{color-ours}{HTML}{E8F7FF}
\DeclareRobustCommand{\added}[1]{\begingroup#1\endgroup}
\definecolor{ieeeblue}{HTML}{00629B}
\usepackage[colorlinks=true,allcolors=ieeeblue]{hyperref}
\begin{document}
\raggedbottom

\title{DVFace: Spatio-Temporal Dual-Prior Diffusion\\ for Video Face Restoration}

\author{Zheng Chen\IEEEauthorrefmark{2},
        Bowen Chai\IEEEauthorrefmark{2},
        Rongjun Gao,
        Mingtao Nie,
        Xi Li,\\
        Bingnan Duan,
        Jianping Fang,
        Xiaohong Liu,
        Linghe Kong,~\IEEEmembership{Fellow,~IEEE},
        Yulun Zhang\IEEEauthorrefmark{1}
\IEEEcompsocitemizethanks{
\IEEEcompsocthanksitem Zheng Chen, Bowen Chai, Rongjun Gao, Mingtao Nie, Xiaohong Liu, Linghe Kong, and Yulun Zhang are with Shanghai Jiao Tong University.
\IEEEcompsocthanksitem Xi Li, Bingnan Duan, and Jianping Fang are with Meituan Inc.\IfFormatAtLeastTF{2024-06-01}{\protect\vspace{\baselineskip}}{}}
\thanks{\IEEEauthorrefmark{2}\,Zheng Chen and Bowen Chai contributed equally to this work.\protect\newline
\IEEEauthorrefmark{1}\,Corresponding author: Yulun Zhang (email: yulun100@gmail.com).}
}

\markboth{IEEE Transactions on Pattern Analysis and Machine Intelligence}{Chen \MakeLowercase{\textit{et al.}}: DVFace: Spatio-Temporal Dual-Prior Diffusion for Video Face Restoration}

\IEEEpubid{0000--0000/00\$00.00~\copyright~2021 IEEE}

\IEEEtitleabstractindextext{
\begin{abstract}
Video face restoration aims to enhance degraded face videos into high-quality results with realistic facial details, stable identity, and temporal coherence. Recent diffusion-based methods have brought strong generative priors to restoration and enabled more realistic detail synthesis. However, existing approaches for face videos still rely heavily on generic diffusion priors and multi-step sampling, which limit both facial adaptation and inference efficiency. These limitations motivate the use of one-step diffusion for video face restoration, yet achieving faithful facial recovery alongside temporally stable outputs remains challenging. In this paper, we propose, \textbf{DVFace}, a one-step diffusion framework for real-world video face restoration. Specifically, we introduce a spatio-temporal dual-codebook design to extract complementary spatial and temporal facial priors from degraded videos. We further propose an asymmetric spatio-temporal fusion module to inject these priors into the diffusion backbone according to their distinct roles. Evaluation on various benchmarks shows that DVFace delivers superior restoration quality, temporal consistency, and identity preservation compared to recent methods. Code:~\url{https://github.com/zhengchen1999/DVFace}.
\end{abstract}

}

\maketitle

\IEEEdisplaynontitleabstractindextext
\IEEEpeerreviewmaketitle

\section{Introduction}

\IEEEPARstart{V}{ideo} face restoration (VFR) aims to recover high-quality (HQ) face videos from low-quality (LQ) inputs degraded by various factors (\eg, blur and noise)~\cite{rota2023video,li2025survey}. Compared with single-image face restoration, VFR is more difficult because it must address both spatial and temporal degradation~\cite{zhou2022towards,yang2021gan,yue2024difface}. A desirable model should recover clear facial structures, faithful textures, and natural details in each frame. At the same time, it should maintain a consistent identity, appearance, and motion across neighboring frames \added{and over the full video sequence}. This requirement makes real-world VFR highly challenging~\cite{ren2019face,feng2024kalman}.

Several solutions extend general restoration models from images or videos to face videos. These methods can exploit powerful reconstruction backbones or temporal aggregation mechanisms, but they are not designed for facial content. As a result, they often fail to recover fine facial details under severe degradations~\cite{wang2019edvr,chan2022basicvsr++,chan2022investigating}. To better solve VFR, some methods develop specialized VFR frameworks by introducing facial priors, such as geometric cues, generative priors, or codebook-based representations~\cite{ren2019face,wang2023restoreformer++,zhu2022blind,feng2024kalman,chen2025dicface}. These methods improve facial plausibility and temporal stability \added{by incorporating face-specific constraints} to some extent. However, their restoration quality is still limited in \added{complex and diverse} real-world scenarios. Several methods \added{therefore continue to} struggle to jointly achieve high perceptual realism and stable temporal consistency.

\begin{figure}[t]
\centering
\includegraphics[width=\linewidth]{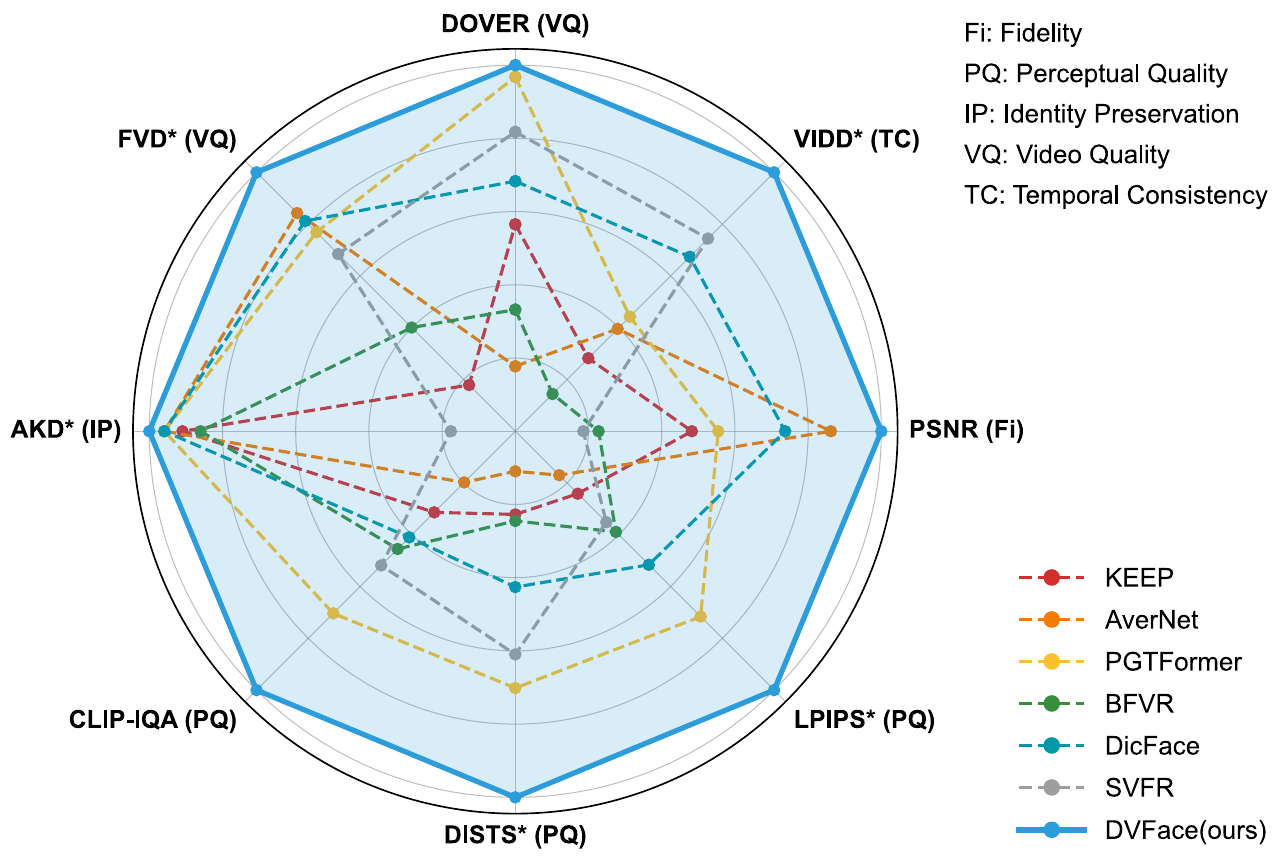}
\caption{Performance comparison on the VFHQ-Test~\cite{xie2022vfhq} dataset. Metrics marked with $*$ indicate lower is better. \added{DVFace achieves the best results on most of the evaluated metrics in the overall comparison.}}
\vspace{-1.mm}
\label{fig:radar}
\end{figure}

Recently, diffusion models have shown strong potential for image and video restoration because of their powerful generative priors~\cite{ho2020denoising,rombach2022high,blattmann2023align}. They are especially attractive for face restoration, where realistic detail synthesis is crucial. However, most existing diffusion-based restoration methods rely on multi-step sampling. Their repeated denoising process leads to heavy computation and high inference latency~\cite{wang2025seedvr,xie2025star,yeh2024diffir2vr,lin2024diffbir,li2025diffvsr}. This issue becomes more severe for videos, where multiple frames must be restored jointly. 

To improve efficiency, one-step diffusion has emerged as an attractive alternative~\cite{wu2024one,li2024unleashing,dong2025tsd}. By compressing the denoising trajectory into a single step, it greatly reduces inference cost. Promising results have been reported in generic image generation, image restoration, and video restoration~\cite{wang2025osdface,chen2025dove,sun2025one}. However, extending one-step diffusion to video face restoration remains highly challenging. \textbf{First}, video face restoration places much stricter demands on identity preservation than generic restoration tasks. Minor structural deviations can lead to obvious identity inconsistency. \textbf{Second}, temporal coherence is crucial. Without progressive refinement, one-step models are more likely to produce unstable details and temporal artifacts~\cite{zhuang2025flashvsr,zhang2025infvsr,li2025asymmetric}. \textbf{Third}, existing video face restoration methods mainly rely on implicit or entangled representations~\cite{zhang2025vividface,wang2025svfr}. Such representations are not explicitly model reusable facial appearance priors and temporal dynamics, limiting restoration.

\begin{figure*}[t]
\footnotesize
\centering

    \begin{adjustbox}{valign=t}
    \begin{tabular}{ccccccccc}
    \hspace{-1.5mm}
    \includegraphics[width=0.136\textwidth]{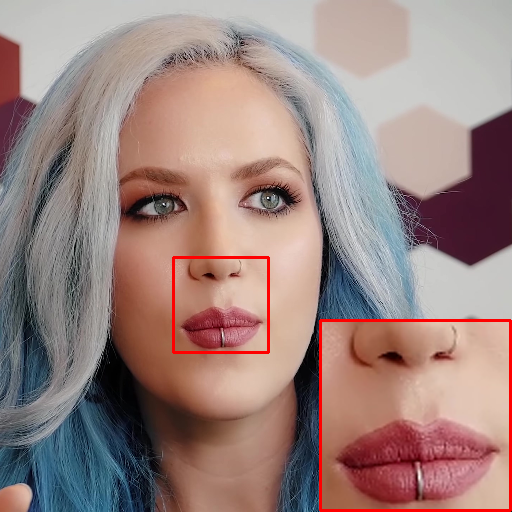} \hspace{-3.mm} &
    \includegraphics[width=0.136\textwidth]{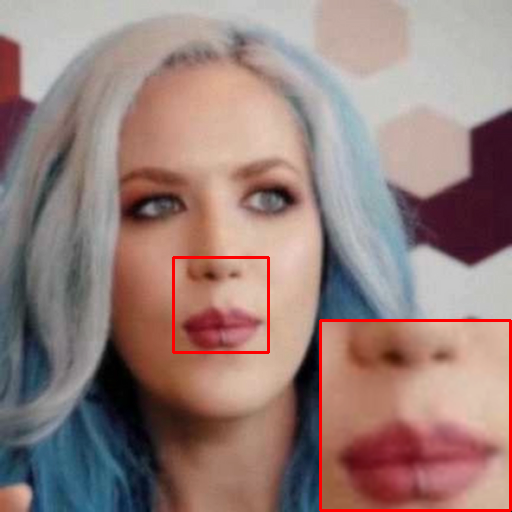} \hspace{-3.mm} &
    \includegraphics[width=0.136\textwidth]{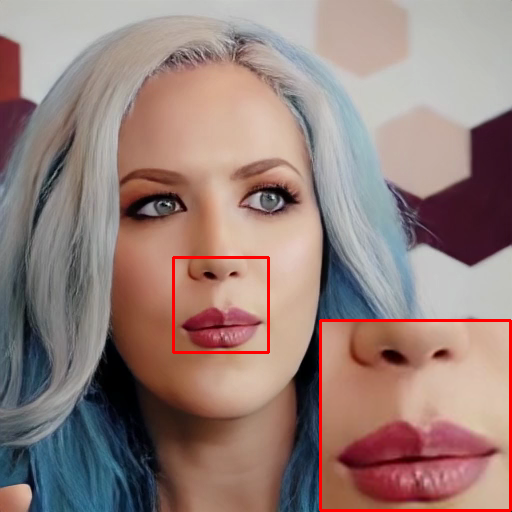} \hspace{-3.mm} &
    \includegraphics[width=0.136\textwidth]{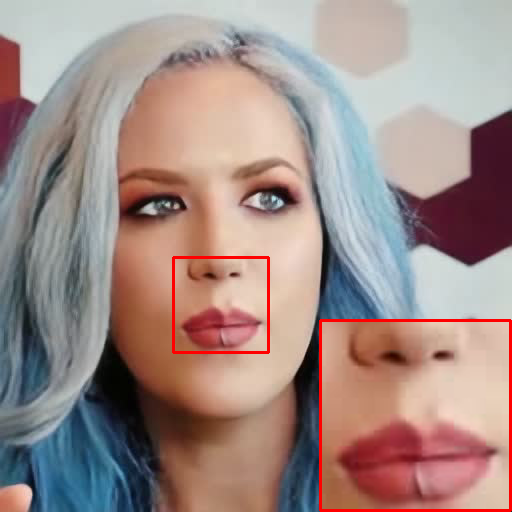} \hspace{-3.mm} &
    \includegraphics[width=0.136\textwidth]{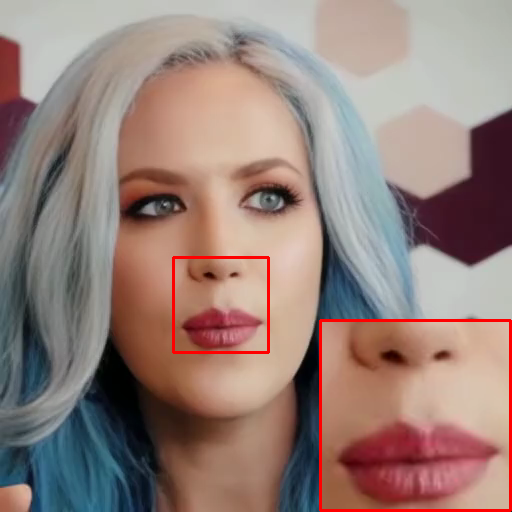} \hspace{-3.mm} &
    \includegraphics[width=0.136\textwidth]{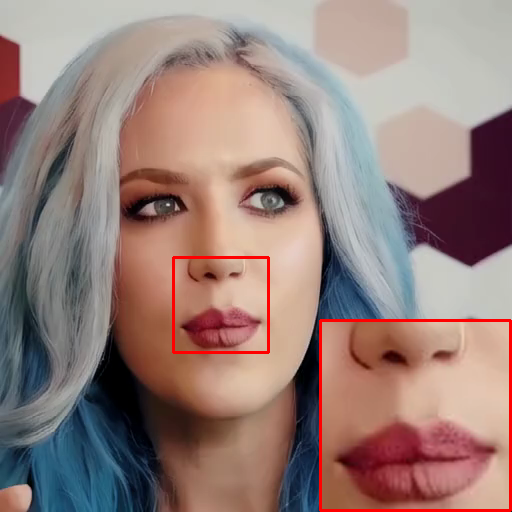} \hspace{-3.mm} &
    \includegraphics[width=0.136\textwidth]{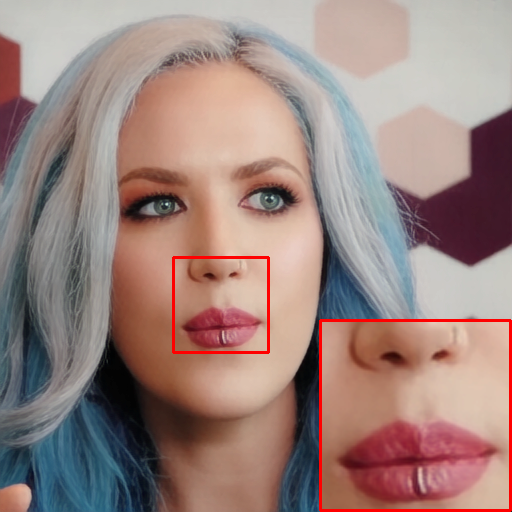} \hspace{-3.mm} &
    \\
    HQ (synthetic)  \hspace{-3.mm} &
    LQ  \hspace{-3.mm} &
    PGTFormer~\cite{xu2024beyond} \hspace{-3.mm} &
    BFVR~\cite{wang2025efficient} \hspace{-3.mm} & 
    DicFace~\cite{chen2025dicface} \hspace{-3.mm} &
    SVFR~\cite{wang2025svfr} \hspace{-3.mm} &
    DVface (ours) \hspace{-3.mm}
    \\
    \end{tabular}
    \end{adjustbox}
    \\
    
    \begin{adjustbox}{valign=t}
    \begin{tabular}{ccccccccc}
    \hspace{-1.5mm}
    \includegraphics[width=0.136\textwidth]{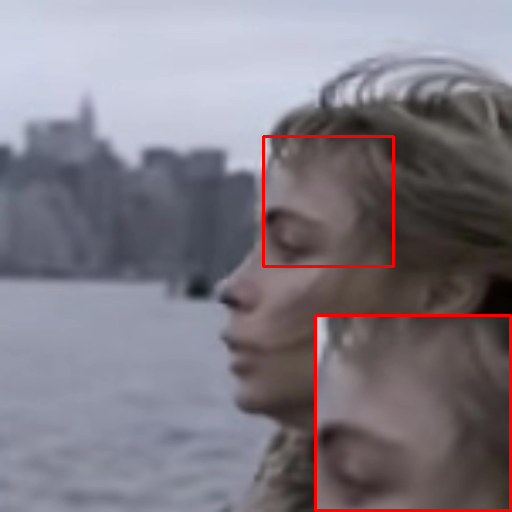} \hspace{-3.mm} &
    \includegraphics[width=0.136\textwidth]{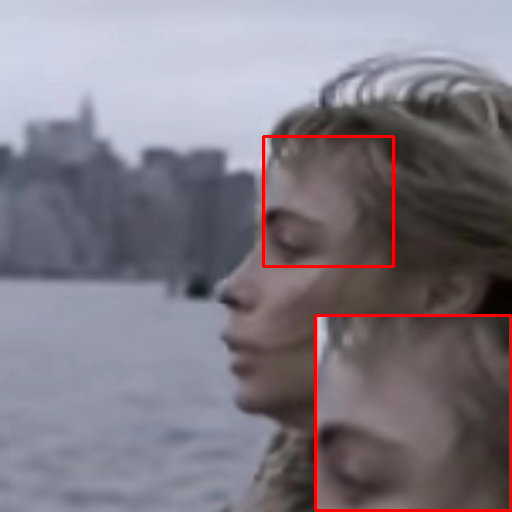} \hspace{-3.mm} &
    \includegraphics[width=0.136\textwidth]
    {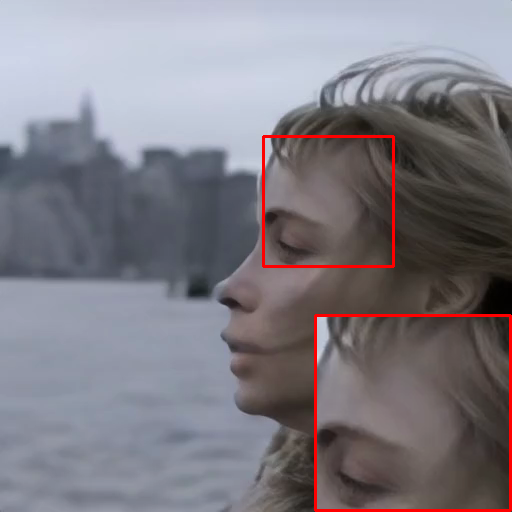} \hspace{-3.mm} &
    \includegraphics[width=0.136\textwidth]{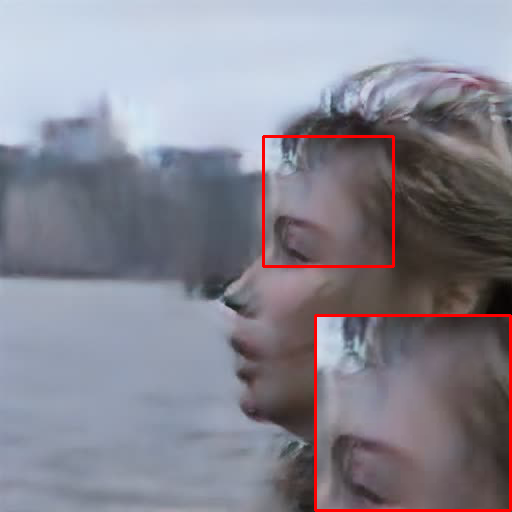} \hspace{-3.mm} &
    \includegraphics[width=0.136\textwidth]{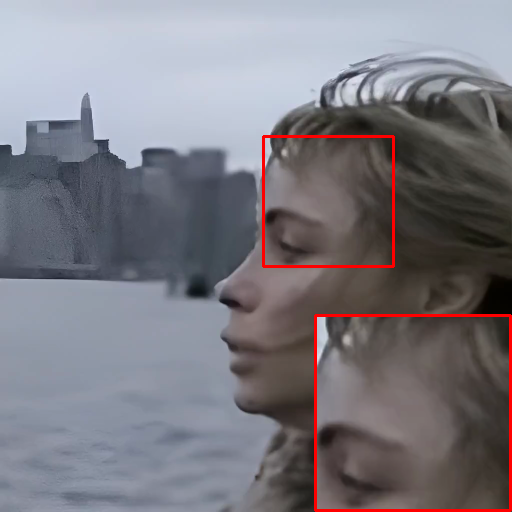} \hspace{-3.mm} &
    \includegraphics[width=0.136\textwidth]{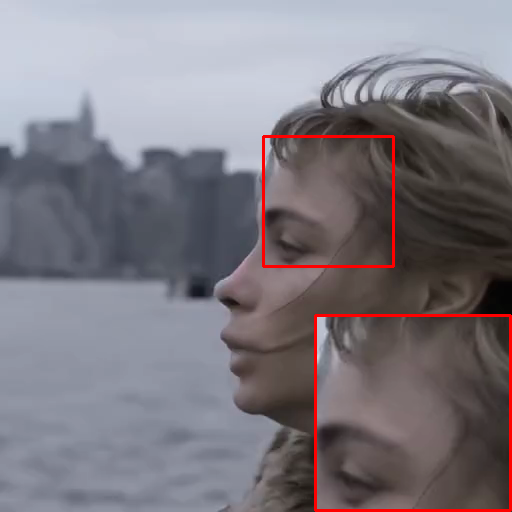} \hspace{-3.mm} &
    \includegraphics[width=0.136\textwidth]{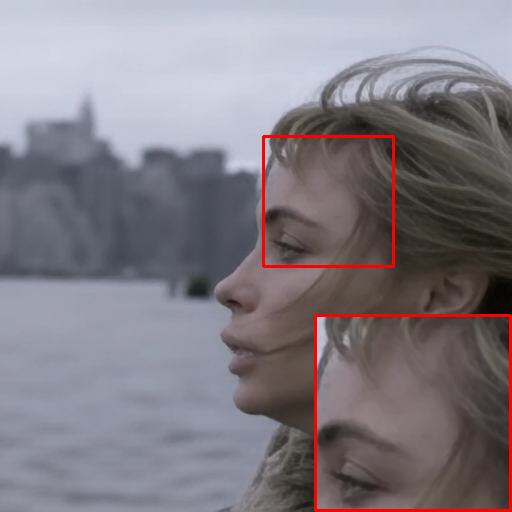} \hspace{-3.mm} &
    \\
    LQ (real-world)  \hspace{-3.mm} &
    AverNet~\cite{zhao2024avernet}  \hspace{-3.mm} &
    PGTFormer~\cite{xu2024beyond} \hspace{-3.mm} &
    BFVR~\cite{wang2025efficient} \hspace{-3.mm} & 
    DicFace~\cite{chen2025dicface} \hspace{-3.mm} &
    SVFR~\cite{wang2025svfr} \hspace{-3.mm} &
    DVface (ours) \hspace{-3.mm}
    \end{tabular}
    \end{adjustbox}

\caption{Visual comparisons on synthetic and real-world datasets. Please zoom in for details. For the real-world example, the reference image (HQ) is unavailable. DVFace restores faithful videos with sharp details.}
\label{fig:performance}
\end{figure*}

In this paper, we propose \textbf{DVFace}, a one-step diffusion framework for real-world video face restoration. Our method is built upon a pretrained diffusion~\cite{wan2025wan} and is designed to explicitly exploit facial priors from degraded input videos. The overall framework \added{therefore} performs single-step restoration while conditioning the diffusion backbone on structured facial priors. To this end, we introduce a \textbf{spatio-temporal dual-codebook} design. The spatial codebook captures stable facial structures and fine appearance details. The temporal codebook models motion variation and dynamic evolution across frames. These two priors are \added{highly} complementary and provide explicit guidance for both facial fidelity and temporal consistency.

Based on the dual-codebook priors, we further propose an \textbf{asymmetric spatio-temporal fusion} module. This design is motivated by the distinct roles of spatial and temporal priors. Temporal priors are more global and mainly promote temporal coherence. Spatial priors provide facial structures and textures. Thus, injecting them in the same way is suboptimal. In our design, temporal priors act as global modulation signals, while spatial priors are injected as local residual details under temporal guidance. 

As revealed in Fig.~\ref{fig:radar}, our approach maintains a favorable balance among restoration quality, temporal consistency, and identity preservation. 
The visual results in Fig.~\ref{fig:performance} further indicate that DVFace produces more realistic details and more temporally stable results than existing methods. The main contributions \added{are summarized as follows}:

\begin{itemize}
    \item We propose a one-step diffusion framework, DVFace, tailored for real-world video face restoration, enabling efficient inference \added{in a single denoising step} while achieving high-quality reconstruction.
    \item We introduce the spatio-temporal dual-codebook prior design and the asymmetric spatio-temporal fusion module, which explicitly utilize \added{spatial and temporal information from} complementary facial priors.
    \item Extensive experiments \added{across multiple benchmarks} show that DVFace achieves superior restoration quality, stronger temporal consistency, and better identity consistency than existing methods.
\end{itemize}

\section{Related Work}
\vspace{1.5mm}
\subsection{Video Face Restoration}
Video face restoration (VFR) faces the dual challenge of recovering spatial details while strictly maintaining temporal consistency~\cite{feng2024kalman,wang2025efficient,li2025interlcm}. Applying single-image restoration models frame-by-frame inevitably introduces severe identity flickering~\cite{zhou2022towards,gu2022vqfr,wang2021towards,yang2021gan,yue2024difface}. Consequently, contemporary research has shifted exclusively towards specialized VFR methodologies. Early video super-resolution networks utilized explicit temporal alignment techniques like optical flow or deformable convolutions~\cite{pan2021deep,chan2022investigating,chen2022videoinr}. For instance, methods like EDVR~\cite{wang2019edvr} and BasicVSR++~\cite{chan2022basicvsr++} excel at temporal smoothing but lack specific facial priors. As a result, they may produce over-smoothed textures under severe real-world blind degradations. To tackle this, subsequent approaches incorporated facial priors. Initial methods integrated geometric priors to guide the structural restoration~\cite{ren2019face,li2022faceformer,hu2020face,zhu2022blind}. For example, PGTFormer~\cite{xu2024beyond} utilizes facial parsing maps as contextual clues to reduce artifacts and temporal jitters. Later works shifted to utilizing robust generative and codebook priors to recover fine-grained details ~\cite{wang2023restoreformer++,zhou2022towards}. To illustrate, KEEP~\cite{feng2024kalman} adopts a Kalman-inspired latent propagation mechanism to maintain stable face priors across frames. DicFace~\cite{chen2025dicface} relaxes discrete codebooks into continuous Dirichlet distributions to ensure temporal coherence. Most recently, advanced methods construct unified spatiotemporal frameworks to balance spatial fidelity and temporal coherence~\cite{zhang2025vividface,wang2025svfr}. For instance, SVFR~\cite{wang2025svfr} leverages stable video diffusion priors within a unified architecture to handle video face restoration. Despite these advancements, existing VFR methods still struggle to achieve high perceptual realism and stable temporal consistency in challenging real-world scenarios.

\subsection{Diffusion Models}
\vspace{.5mm}
Diffusion models are currently the dominant generative paradigm for image and video restoration~\cite{ho2020denoising,rombach2022high,blattmann2023align,zhang2023adding}. They provide unprecedented priors for recovering complex details. However, their real-world application is bottlenecked by the strict trade-off between generative fidelity and sampling efficiency~\cite{wang2025osdface}. Multi-step diffusion models adapt pretrained priors to hallucinate realistic textures ~\cite{wang2025seedvr,xie2025star}. For example, SeedVR~\cite{wang2025seedvr} relies on iterative denoising and shifted window attention to process real-world videos. Despite impressive visual quality, this iterative process requires tens to hundreds of costly network evaluations~\cite{yeh2024diffir2vr,lin2024diffbir}. Consequently, when applied to videos, methods like Upscale-A-Video~\cite{zhou2024upscale} and STAR~\cite{xie2025star} suffer from prohibitive computational overhead and severe inference latency. Furthermore, the inherent stochasticity of multi-step diffusion paths introduces unexpected temporal discontinuities and texture flickering across frames~\cite{li2025diffvsr,wang2025svfr}.

Single-step diffusion models have recently been explored to alleviate this severe computational burden. They leverage knowledge distillation or adversarial training to compress the sampling trajectory into a single step ~\cite{wu2024one,li2024unleashing,dong2025tsd}. For static images, methods such as OSDFace~\cite{wang2025osdface} successfully achieve one-step inference for face restoration. Extending this efficiency to video restoration, however, is highly challenging due to exorbitant multi-frame training costs and strict fidelity demands. Recently, a few one-step video restoration methods have emerged to accelerate temporal inference~\cite{chen2025dove,sun2025one,wang2025seedvr,wang2025seedvr2,chen2026diffst}. For instance, DOVE introduces a latent-pixel training strategy to achieve single-step video super-resolution~\cite{chen2025dove}. DLoRAL~\cite{sun2025one} employs dual LoRA learning to accomplish the same highly efficient inference. However, these models struggle to preserve precise facial identities and temporal coherence under severe real-world degradations~\cite{zhuang2025flashvsr,zhang2025infvsr,li2025asymmetric}. In general, current diffusion-based restoration methods remain insufficient for video face restoration, as they cannot effectively exploit facial and temporal priors \added{within a unified restoration framework.}

\begin{figure*}[t]
    \centering
    \scriptsize
    \includegraphics[width=\linewidth]{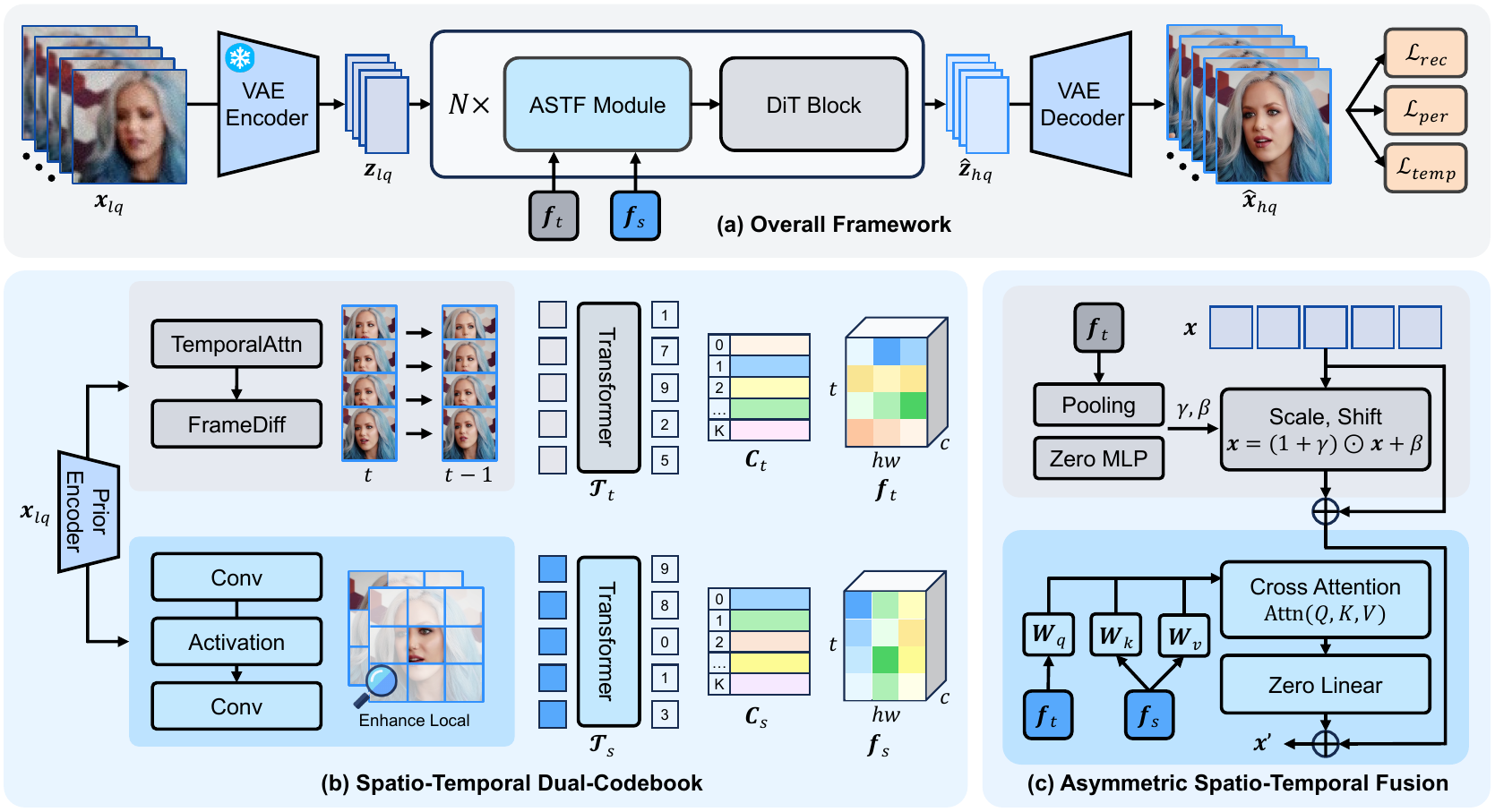}
    \vspace{-4.mm}
    \caption{Overview of \textbf{DVFace}. (a) Overall Framework: DVFace restores low-quality face videos with one-step diffusion and facial priors. (b) Spatio-Temporal Dual-Codebook (STDC): spatial and temporal codebooks extract complementary facial priors. (c) Asymmetric Spatio-Temporal Fusion (ASTF): injecting temporal priors globally and spatial priors locally.}
    \label{fig:method}
    \vspace{-2.mm}
\end{figure*}

\vspace{1.mm}
\section{Method}
\vspace{1.mm}
This section presents the proposed video face restoration model. First, we outline the overall framework, which leverages a pretrained text-to-video (T2V) model and incorporates spatio-temporal dual-codebook priors. Then, we focus on \added{the detailed design of} two core components that are designed for prior extraction and integration, respectively. Finally, we describe the full training pipeline.

\vspace{1.mm}
\subsection{Overall Framework}
\vspace{1.mm}
\label{sec:overall_framework}

Our video face restoration framework adopts a one-step diffusion formulation built upon a pretrained T2V model (\ie, Wan 2.1~\cite{wan2025wan}). While the pretrained backbone provides strong generative priors, our approach focuses on extracting and leveraging facial priors from low-quality (LQ) face videos to enhance the restoration. Specifically, we construct spatial and temporal codebooks to capture reusable facial representations, and augment each diffusion transformer (DiT) block with a lightweight fusion module. These components enable the backbone to be \added{consistently and} continuously conditioned on the proposed spatio-temporal dual-codebook priors during denoising, resulting in high-fidelity restoration and improved temporal coherence.

Fig.~\ref{fig:method} shows the overall framework of our method. Given a LQ video $\boldsymbol{x}_{lq}$, we encode it into a latent representation $\boldsymbol{z}_{lq}$ using the VAE encoder $\mathcal{E}$. Following previous works~\cite{wu2023seesr,wu2024one,wang2025osdface,chen2025dove}, we treat $\boldsymbol{z}_{lq}$ as the noisy latent $\boldsymbol{z}_{t}$ at a diffusion timestep $t$, written as:
\begin{equation}
    \boldsymbol{z}_t = t \boldsymbol{\epsilon} + (1 - t) \boldsymbol{z}_{hq}, \quad
    \boldsymbol{z}_{lq} = \boldsymbol{z}_{t^*}, \quad t,\,t^* \in [0, 1],
\end{equation}
where $\boldsymbol{\epsilon}$ is the Gaussian noise, $\boldsymbol{z}_{hq}$ is the encoded high-quality (HQ) video, and $t^*$ denotes a fixed timestep. In parallel, we extract spatial and temporal facial priors from $\boldsymbol{x}_{lq}$ via the well-trained dual codebooks, which are denoted as $\boldsymbol{f}_{s}$ and $\boldsymbol{f}_t$ respectively. These priors are injected into the DiT to predict a velocity field $\boldsymbol{v}_{\theta}(\cdot)$ for denoising, producing the restored HQ latent $\hat{\boldsymbol{z}}_{hq}$. Accordingly, the complete one-step diffusion denoising process is formulated as:
\begin{equation}
    \hat{\boldsymbol{z}}_{hq} = \boldsymbol{z}_{lq} - t^* \boldsymbol{v}_{\theta}(\boldsymbol{z}_{lq}, t^*, \boldsymbol{c}_{text}, \boldsymbol{f}_{s}, \boldsymbol{f}_t).
\end{equation}
\added{Here, $\boldsymbol{c}_{text}$ denotes the textual condition of the pretrained T2V backbone. We use an empty text prompt throughout restoration, including both training and inference.}

Finally, the latent $\hat{\boldsymbol{z}}_{hq}$ is decoded by the VAE decoder $\mathcal{D}$ to obtain the output $\hat{\boldsymbol{x}}_{hq}$ as the restoration result.

\vspace{1.mm}
\subsection{Spatio-Temporal Dual-Codebook}
\vspace{1.mm}
\label{sec:dual_codebook}
We adapt a generative model for video face restoration by using the low-quality (LQ) inputs as the starting point and performing diffusion denoising guided by an empty or fixed text prompt. However, this paradigm does not fully exploit the rich \added{facial} priors contained in the LQ video, limiting its ability to restore face videos with high fidelity, identity consistency, and temporal coherence. Previous studies~\cite{zhou2022towards,gu2022vqfr,wang2025osdface} introduce \added{learned} codebooks to learn reusable facial representations, which can be decoded into high-quality outputs \added{with realistic facial details}. Inspired by this, we argue that the representations learned by \added{the} codebook can serve as strong priors for the restoration process.

\noindent\textbf{Codebook Preliminary.} 
A codebook learns a finite set of representative latent embeddings $\mathcal{C}=\{\boldsymbol{c}_k\}_{k=1}^{K}$, where each code entry $\boldsymbol{c}_k \in \mathbb{R}^{d}$ corresponds to a prototypical facial pattern. 
In general, the codebook is learned through high-quality (HQ) reconstruction, so that the discrete latent space captures clean and reusable facial structures and textures. Given an HQ input \(x_h\), an encoder maps it to latent features \(\boldsymbol{z}_h = E_H(x_h)\). For each spatial location \((i,j)\) in the latent feature map, the corresponding latent token \(\boldsymbol{z}_h^{(i,j)}\) is quantized by nearest-neighbor lookup in the codebook:
\begin{equation}
\boldsymbol{z}_q^{(i,j)} = \underset{\boldsymbol{c}_k \in \mathcal{C}}{\arg\min} \|\boldsymbol{z}_h^{(i,j)}-\boldsymbol{c}_k\|_2, \quad \boldsymbol{s}^{(i,j)} = \underset{k}{\arg\min} \| \boldsymbol{z}_h^{(i,j)} - \boldsymbol{c}_k \|_2 .
\end{equation}
The quantized latent representation $\hat{\boldsymbol{z}}_c$ is then decoded to reconstruct \added{the original} HQ target, which encourages the codebook to organize common facial patterns into a compact \added{and reusable} discrete latent vocabulary.

During restoration, the learned codebook serves as a clean prior space for recovering faithful and realistic facial details. 
However, degradations in low-quality (LQ) inputs may shift their latent features away from the HQ manifold, making direct code matching unreliable. 
To mitigate this issue, practical restoration frameworks often employ a Transformer to model global interrelations among latent tokens for better code prediction.

Nevertheless, existing codebook-based methods mainly model spatial information in individual images. When extended to video restoration, this image-centric paradigm cannot fully exploit the temporal information contained in LQ sequences, such as motion continuity, expression evolution, and identity dynamics across neighboring frames, which are crucial for temporally consistent restoration. Motivated by this, we extend the conventional single-codebook paradigm to a spatio-temporal dual-codebook design.

\noindent\textbf{Priors Extraction.}
Given an LQ face video $\boldsymbol{x}_{lq} \in \mathbb{R}^{T \times H \times W \times 3}$, an encoder first embeds it into latent features $\boldsymbol{z}_l \in \mathbb{R}^{t \times h \times w \times d}$. From $\boldsymbol{z}_l$, we further derive \added{two complementary} temporal and spatial latents to query the corresponding codebooks. Specifically, the temporal latents $\boldsymbol{z}_{t}$ are obtained via a \emph{temporal interaction module}:
\begin{equation}
\boldsymbol{z}_{t} = \mathrm{TemporalAttn}(\boldsymbol{z}_l) + \mathrm{FrameDiff}(\boldsymbol{z}_l),
\end{equation}
where $\mathrm{TemporalAttn}(\cdot)$ performs self-attention along the temporal dimension by treating the latent features as token sequences in $\mathbb{R}^{(h \cdot w) \times t \times d}$, thereby capturing long-range temporal dependencies across frames at each spatial location. $\mathrm{FrameDiff}(\cdot)$ computes adjacent-frame feature differences to encode temporal variations and motion trajectories. In parallel, the spatial latents $\boldsymbol{z}_s$ are obtained through two convolutional blocks, which capture localized appearance patterns and structural details from $\boldsymbol{z}_l$:
\begin{equation}
    \boldsymbol{z}_s = \mathrm{Conv}_2\left(\mathrm{Activation}(\mathrm{Conv}_1(\boldsymbol{z}_l))\right).
\end{equation}

Since degraded LQ features may deviate from the correct HQ codes and be mistakenly assigned to nearby entries, we use two independent Transformers, $\mathcal{T}_s$ and $\mathcal{T}_t$, to \added{separately} process the spatial and temporal latents, respectively. Specifically, the spatial latents $\boldsymbol{z}_s$ and temporal latents $\boldsymbol{z}_t$ obtained above are reshaped into token sequences in $\mathbb{R}^{(t \cdot h \cdot w)\times c}$, and then fed into the Transformers to model global interrelations \added{across the complete sequence} for more accurate code prediction. Based on the predicted code indices, the corresponding entries are \added{then} selected from the learned spatial and temporal codebooks to form the quantized spatial and temporal features, which can be \added{compactly} written as:
\begin{equation}
\begin{aligned}
\boldsymbol{z}_{q,s}^{(i,j)} = \boldsymbol{c}_{\boldsymbol{s}_s^{(i,j)}, s},  \quad \boldsymbol{s}_s^{(i,j)} &= \underset{k}{\arg\min} \| \boldsymbol{z}_s^{(i,j)} - \boldsymbol{c}_{k,s} \|_2, \quad \\ 
\boldsymbol{z}_{q,t}^{(i,j)} = \boldsymbol{c}_{\boldsymbol{s}_t^{(i,j)}, t}, \quad  \boldsymbol{s}_t^{(i,j)} &= \underset{k}{\arg\min} \| \boldsymbol{z}_t^{(i,j)} - \boldsymbol{c}_{k,t} \|_2,
\end{aligned}
\end{equation}
where \((i,j)\) denotes the token location in the corresponding latent map. With well-learned codebooks, we can extract reliable spatial and temporal priors from degraded video inputs, where $\boldsymbol{f}_s := \boldsymbol{z}_{q,s}$ captures stable facial structures and appearance details, and $\boldsymbol{f}_t := \boldsymbol{z}_{q,t}$ provides complementary temporal cues such as motion variation and dynamic evolution across frames.

\subsection{Asymmetric Spatio-Temporal Fusion}
Although the proposed spatio-temporal dual-codebook priors provide complementary cues for video face restoration, directly injecting them into the diffusion backbone through naive operations such as simple addition or cross-attention is suboptimal. At the same time, spatial priors primarily capture fine-grained, pixel-level facial details, whereas temporal priors encode more global dynamic information, such as motion patterns and temporal style consistency across neighboring frames. Treating these two types of priors in an identical manner would fail to fully exploit their distinct properties. To this end, we design a lightweight spatio-temporal fusion module that purposefully and asymmetrically integrates spatial and temporal priors into each individual DiT block \added{throughout the backbone}, enabling high-fidelity and temporally coherent video restoration.

The proposed fusion module is illustrated in Fig.~\ref{fig:method}(c). For the temporal prior $\boldsymbol{f}_t \in \mathbb{R}^{b\times t \times h \times w \times c}$, we first apply spatio-temporal pooling to obtain a compact global descriptor, which is then mapped by an MLP to the DiT feature space to predict a pair of \added{learnable} modulation parameters $(\gamma,\beta)$. These parameters are \added{then} used in the fusion module to scale and shift the input tokens $\boldsymbol{x}$:
\begin{equation}
(\gamma, \beta) = \mathrm{MLP}(\mathrm{Pool}(\boldsymbol{f}_t)), \quad
\tilde{\boldsymbol{x}} = (1+\gamma)\odot \boldsymbol{x} + \beta,
\end{equation}
where $\tilde{\boldsymbol{x}}$ denotes the temporally modulated features. Notably, the same modulation parameters are shared across different DiT blocks, rather than being predicted separately for each layer. In this way, the temporal prior provides a consistent global temporal bias across layers, which is well suited for modeling motion tendency and temporal coherence without introducing explicit local structures.

For the spatial prior $\boldsymbol{f}_s$, we inject it as token-level residual details. Considering that spatial priors extracted from degraded inputs may still contain unreliable local patterns, we use temporal cues to temporally pre-refine the spatial details before fusion \added{by exploiting complementary evidence across frames.} Specifically,
the temporal prior is linearly projected to the query, while the
spatial prior is projected to the key and value for cross-attention:
\begin{equation}
\begin{gathered}
\Delta \boldsymbol{x} = \mathrm{Attn}(Q=W_q\boldsymbol{f}_t,\, K=W_k\boldsymbol{f}_s,\, V=W_v\boldsymbol{f}_s), \\
\boldsymbol{x}' = \tilde{\boldsymbol{x}} + \mathrm{Proj}(\Delta \boldsymbol{x}),
\end{gathered}
\end{equation}
where $\boldsymbol{x}'$ is the output of the fusion module. For simplicity, we omit the flattening and unflattening operations before and after attention. In this way, the spatial prior contributes explicit fine-grained facial details for local refinement, while temporal guidance ensures that only temporally compatible and reliable details are injected into the backbone. As a result, the fused features are enriched with faithful appearance cues while avoiding degraded local artifacts, leading to more faithful and temporally consistent restoration.

To avoid the spatio-temporal priors disturbing the pretrained backbone at the early stage of training, we zero-initialize the last layer of all projection modules in the fusion branch, \added{which is denoted as Zero Linear}. \added{This initialization lets the fusion module start from an exact identity mapping and enables gradual, stable adaptation to the pretrained backbone without abrupt feature perturbations.}

\subsection{Training Objectives}
The whole training pipeline is divided into two stages. We first learn the spatio-temporal dual-codebook priors extraction module, and then adapt the diffusion backbone equipped with fusion modules. Specifically, the priors extraction module is trained by first learning the codebooks from HQ reconstruction and then adapting it to LQ inputs in practice. The diffusion model is trained with the prior extractor and VAE encoder fixed, while the DiT, VAE decoder, and fusion modules are jointly optimized. 

\noindent\textbf{Stage-1: Spatio-Temporal Dual-Codebook Priors Learning.}
For codebook learning, the HQ video $\boldsymbol{x}_{hq}$ is first encoded into latent features $\boldsymbol{z}_{h}$. Spatial and temporal latents are then constructed as described in Sec.~\ref{sec:dual_codebook} to query the two codebooks, producing the quantized features $\boldsymbol{z}_{q,s}$ and $\boldsymbol{z}_{q,t}$. The reconstructed HQ video $\hat{\boldsymbol{x}}_{hq}$ is obtained by decoding the quantized latent $\boldsymbol{z}_{q} = \boldsymbol{z}_{q,s} + \boldsymbol{z}_{q,t}$. Following common practice in codebook-based restoration, we train the autoencoder and codebooks using reconstruction, perceptual, code-level feature, and adversarial losses:
\begin{equation}
\mathcal{L}_{s1} = \mathcal{L}_{1} + \mathcal{L}_{per} + \mathcal{L}_{feat} + \lambda_{adv} \mathcal{L}_{adv}.
\end{equation}
In this objective, $\mathcal{L}_{1}$ computes the L1 loss between the reconstructed video $\hat{\boldsymbol{x}}_{hq}$ and the ground-truth video $\boldsymbol{x}_{hq}$, and $\mathcal{L}_{per}$ denotes the L2 loss between their features extracted by VGG19 $\Phi$. The \added{corresponding perceptual loss}, code-level feature loss, and adversarial loss are defined as:
\begin{equation}
\begin{aligned}
&\mathcal{L}_{per} = \mathcal{L}_2(\Phi(\hat{\boldsymbol{x}}_{hq}), \Phi(\boldsymbol{x})) \\
&\mathcal{L}_{feat} = \mathcal{L}_{mse} \left(\mathrm{sg}(\boldsymbol{z}_{h}), \boldsymbol{z}_{q}\right)
+ \beta \mathcal{L}_{mse}\left(\boldsymbol{z}_{h}, \mathrm{sg}(\boldsymbol{z}_{q})\right), \\
&\mathcal{L}_{adv} = \log \mathcal{D}_{adv}(\boldsymbol{x}_{hq}) + \log \left(1 - \mathcal{D}_{adv}(\hat{\boldsymbol{x}}_{hq})\right),
\end{aligned}
\end{equation}
where $\mathcal{L}_{mse}$ denotes the MSE loss, $\mathrm{sg}(\cdot)$ refers to the stop-gradient operator, $\mathcal{D}_{adv}$ represents the discriminator, and $\beta$ balances the update rates of the encoder and the codebooks.

Subsequently, we train the Transformer $\mathcal{T}_s$ and $\mathcal{T}_t$ and further finetune the encoder with LQ inputs, while keeping the decoder and dual codebooks $\mathcal{C}_s$ and $\mathcal{C}_t$ fixed. This step is designed to align clean-video codebook learning with practical degraded-video inference, so that the predicted code indices from LQ inputs remain highly consistent with those derived from HQ videos, even under severe degradations. Cross-entropy prediction loss and code-level feature loss are jointly used in this training stage:
\begin{equation}
\begin{aligned}
&\mathcal{L}_{ce, s} = \sum_{k=1}^{thw} - \boldsymbol{s}_{s}^{k} \log \left( \hat{\boldsymbol{s}}_{s}^{k} \right), \enspace
\mathcal{L}_{ce,t} = \sum_{k=1}^{thw} - \boldsymbol{s}_{t}^{k} \log \left( \hat{\boldsymbol{s}}_{t}^{k} \right), \\
&\mathcal{L}_{cf} = \mathcal{L}_{mse}\left(\boldsymbol{z}_{l}, \mathrm{sg}(\boldsymbol{z}_{q})\right), \enspace
\mathcal{L}_{s1}' = \mathcal{L}_{cf} + \lambda_{ce}\left(\mathcal{L}_{ce,s} + \mathcal{L}_{ce,t}\right).
\end{aligned}
\end{equation}
where $\hat{\boldsymbol{s}}_{s}$ and $\hat{\boldsymbol{s}}_{t}$ denote the spatial and temporal code sequences predicted by $\mathcal{T}_s$ and $\mathcal{T}_t$, respectively, while $\boldsymbol{s}_{s}$ and $\boldsymbol{s}_{t}$ are the corresponding ground-truth code sequences generated from HQ inputs using the fixed dual codebooks. Here, $thw$ is the number of latent tokens, $\boldsymbol{z}_{l}$ denotes the latent feature encoded from the LQ input, $\boldsymbol{z}_{q}$ is the corresponding quantized latent, $\mathrm{sg}(\cdot)$ refers to the stop-gradient operator, and $\lambda_{ce}$ is the weight of the cross-entropy loss.

\noindent\textbf{Stage-2: One-Step Diffusion Restoration Training.} The restoration process is described in Sec~\ref{sec:overall_framework}. We employ \added{three complementary} reconstruction, perceptual and temporal losses, which are formulated as follows:
\begin{equation}
    \mathcal{L}_{s2} = \mathcal{L}_{rec} + \mathcal{L}_{per} + \lambda_{temp}\mathcal{L}_{temp}.
\end{equation}
\added{The reconstruction objective promotes faithful content recovery by combining latent-level and pixel-level MSE losses:}
\begin{equation}
    \mathcal{L}_{rec} = \mathcal{L}_{mse}(\boldsymbol{\hat{z}}_{hq}, \boldsymbol{z}_{hq}) + \mathcal{L}_{mse}(\boldsymbol{\hat{x}}_{hq}, \boldsymbol{x}_{hq}),
\end{equation}
where $\mathcal{L}_{mse}$ represents the MSE loss, $\boldsymbol{\hat{x}}_{hq}$ and $\boldsymbol{x}_{hq}$ are the restored and ground-truth HQ videos, and $\boldsymbol{\hat{z}}_{hq}$ and $\boldsymbol{z}_{hq}$ are the corresponding latents. We adopt LPIPS as the perceptual loss \added{to further constrain visual similarity}, defined as:
\begin{equation}
    \mathcal{L}_{per} = \mathrm{LPIPS}(\boldsymbol{\hat{x}}_{hq}, \boldsymbol{x}_{hq})
\end{equation}
For the temporal loss, we use a warp-based temporal consistency loss that warps the restored frames using the \added{forward and backward} optical flow computed from the corresponding ground-truth HQ video and penalizes the mismatch between aligned temporally neighboring frames:
\begin{equation}
\begin{gathered}
\boldsymbol{\tilde{x}}_{hq}^{bw,i} = \mathrm{Warp}(\boldsymbol{\hat{x}}_{hq}^{i}, \boldsymbol{O}_{gt}^{bw,i}), \quad
\boldsymbol{\tilde{x}}_{hq}^{fw,i} = \mathrm{Warp}(\boldsymbol{\hat{x}}_{hq}^{i}, \boldsymbol{O}_{gt}^{fw,i}),\\
\mathcal{L}_{temp} =
\sum_{i=2}^{T-1}
\left(
\left\| \boldsymbol{\tilde{x}}_{hq}^{bw,i} - \boldsymbol{\hat{x}}_{hq}^{i+1} \right\|_1
+
\left\| \boldsymbol{\tilde{x}}_{hq}^{fw,i} - \boldsymbol{\hat{x}}_{hq}^{i-1} \right\|_1
\right),
\end{gathered}
\end{equation}
where $\mathrm{Warp}(\cdot,\cdot)$ denotes the warping operation, $\boldsymbol{O}_{gt}^{fw,i}$ and $\boldsymbol{O}_{gt}^{bw,i}$ are the forward and backward optical flows computed from the $x_{hq}$, and $T$ is the number of video frames.

\begin{table*}[t]
    \footnotesize
    \centering
    \caption{Quantitative comparison of video face restoration models on synthetic datasets (VFHQ-Test and HDTF). The best and second-best results are colored with \textcolor{red}{red} and \textcolor{blue}{blue}. $\uparrow$ indicates higher is better, and $\downarrow$ indicates lower is better.}
    \vspace{-2.mm}
    \label{tab:quant-synthetic}
        \setlength{\tabcolsep}{2.28mm}
        \begin{tabular}{l|l|c|c|c|c|c|c|c}
            \toprule[0.1em]
            \rowcolor{color-ours} Dataset & Metric & KEEP \cite{feng2024kalman} & AverNet \cite{zhao2024avernet} & PGTFormer \cite{xu2024beyond} & BFVR \cite{wang2025efficient} & DicFace \cite{chen2025dicface} & SVFR \cite{wang2025svfr} & DVFace (ours) \\
            \midrule[0.1em]
            \multirow{15}{*}{VFHQ-Test}
                & PSNR $\uparrow$       & 28.58 & \textcolor{blue}{30.95} & 29.03 & 26.99 & 30.17 & 26.73 & \textcolor{red}{31.81} \\
                & SSIM $\uparrow$       & 0.8735 & \textcolor{blue}{0.8981} & 0.8749 & 0.8625 & 0.8979 & 0.8255 & \textcolor{red}{0.9007} \\
                & LPIPS $\downarrow$    & 0.1815 & 0.1913 & \textcolor{blue}{0.1165} & 0.1361 & 0.1439 & 0.1366 & \textcolor{red}{0.0776} \\
                & DISTS $\downarrow$    & 0.1284 & 0.1367 & \textcolor{blue}{0.0990} & 0.1273 & 0.1161 & 0.1047 & \textcolor{red}{0.0805} \\
                & CLIP-IQA $\uparrow$   & 0.3588 & 0.3133 & \textcolor{blue}{0.5123} & 0.4145 & 0.3968 & 0.4394 & \textcolor{red}{0.6289} \\
                & MUSIQ $\uparrow$      & 58.93 & 49.76 & \textcolor{blue}{69.94} & 58.73 & 61.31 & 64.43 & \textcolor{red}{71.89} \\
                & NIQE $\downarrow$     & 6.20 & 6.33 & \textcolor{red}{4.97} & 5.53 & 6.57 & 5.82 & \textcolor{blue}{5.27} \\
                & MANIQA $\uparrow$     & 0.3568 & 0.2810 & \textcolor{red}{0.4195} & 0.3489 & 0.3463 & \textcolor{blue}{0.3773} & \textcolor{red}{0.4195} \\
                & LIQE $\uparrow$       & 2.70 & 1.43 & \textcolor{blue}{4.02} & 2.32 & 2.55 & 3.15 & \textcolor{red}{4.05} \\
                & AKD $\downarrow$      & 0.0037 & 0.0031 & \textcolor{blue}{0.0031} & 0.0043 & 0.0031 & 0.0126 & \textcolor{red}{0.0026} \\
                & IDS $\uparrow$ & 0.8809 & 0.9432 & 0.9442 & 0.9032 & 0.9482 & \textcolor{blue}{0.9511} & \textcolor{red}{0.9632} \\
                & FVD $\downarrow$      & 158.46 & \textcolor{blue}{78.91} & 87.81 & 131.88 & 82.61 & 97.89 & \textcolor{red}{60.11} \\
                & DOVER $\uparrow$      & 0.6428 & 0.4399 & \textcolor{blue}{0.8535} & 0.5209 & 0.7045 & 0.7749 & \textcolor{red}{0.8703} \\
                & $E^*_{warp}$ $\downarrow$    & 1.06 & 0.99 & 1.21 & 1.65 & \textcolor{blue}{0.94} & 1.03 & \textcolor{red}{0.91} \\
                & VIDD $\downarrow$     & 0.5024 & 0.4916 & 0.4871 & 0.5156 & 0.4650 & \textcolor{blue}{0.4583} & \textcolor{red}{0.4339} \\
            \midrule
            \multirow{15}{*}{HDTF}
                & PSNR $\uparrow$       & 22.08 & \textcolor{blue}{29.10} & 26.25 & 26.39 & 27.65 & 26.90 & \textcolor{red}{29.54} \\
                & SSIM $\uparrow$       & 0.6914 & \textcolor{red}{0.8826} & 0.8402 & 0.8501 & 0.8607 & 0.8548 & \textcolor{blue}{0.8791} \\
                & LPIPS $\downarrow$    & 0.3093 & 0.1797 & 0.1406 & 0.1361 & 0.1771 & \textcolor{blue}{0.1193} & \textcolor{red}{0.1074} \\
                & DISTS $\downarrow$    & 0.1509 & 0.1191 & 0.1061 & 0.1217 & 0.1247 & \textcolor{red}{0.0896} & \textcolor{blue}{0.0931} \\
                & CLIP-IQA $\uparrow$   & 0.3211 & 0.3072 & \textcolor{blue}{0.4281} & 0.3382 & 0.3373 & 0.3794 & \textcolor{red}{0.5972} \\
                & MUSIQ $\uparrow$      & 58.21 & 48.67 & \textcolor{blue}{65.96} & 56.28 & 57.81 & 60.56 & \textcolor{red}{72.58} \\
                & NIQE $\downarrow$     & 5.91 & 5.84 & \textcolor{red}{5.01} & 5.86 & 6.74 & 5.42 & \textcolor{blue}{5.18} \\
                & MANIQA $\uparrow$     & 0.3359 & 0.2528 & \textcolor{blue}{0.3482} & 0.2841 & 0.3037 & 0.3148 & \textcolor{red}{0.3985} \\
                & LIQE $\uparrow$       & 2.10 & 1.15 & \textcolor{blue}{2.96} & 1.58 & 1.64 & 2.38 & \textcolor{red}{3.71} \\
                & AKD $\downarrow$      & 0.0029 & \textcolor{blue}{0.0022} & 0.0023 & 0.0030 & \textcolor{blue}{0.0022} & \textcolor{blue}{0.0022} & \textcolor{red}{0.0020} \\
                & IDS $\uparrow$ & 0.8870 & 0.9430 & 0.9432 & 0.9163 & 0.9440 & \textcolor{blue}{0.9513} & \textcolor{red}{0.9594} \\
                & FVD $\downarrow$      & 160.77 & 84.55 & 100.54 & 133.89 & 107.90 & \textcolor{blue}{76.69} & \textcolor{red}{60.47} \\
                & DOVER $\uparrow$      & 0.5600 & 0.3263 & \textcolor{blue}{0.6988} & 0.3986 & 0.5354 & 0.5773 & \textcolor{red}{0.8085} \\
                & $E^*_{warp}$ $\downarrow$    & 0.6200 & 0.6500 & 0.6800 & 0.7600 & \textcolor{red}{0.4700} & 0.6300 & \textcolor{blue}{0.5100} \\
                & VIDD $\downarrow$     & 0.4188 & 0.4193 & 0.3945 & 0.4033 & 0.3823 & \textcolor{blue}{0.3720} & \textcolor{red}{0.3681} \\
            \bottomrule[0.1em]
        \end{tabular}
\end{table*}

\begin{figure*}[t]
\small
\centering
\begin{adjustbox}{valign=t}
    \setlength{\tabcolsep}{0.8mm}
    \footnotesize
    \begin{tabular}{ccccccc}
    \includegraphics[width=0.13\textwidth]{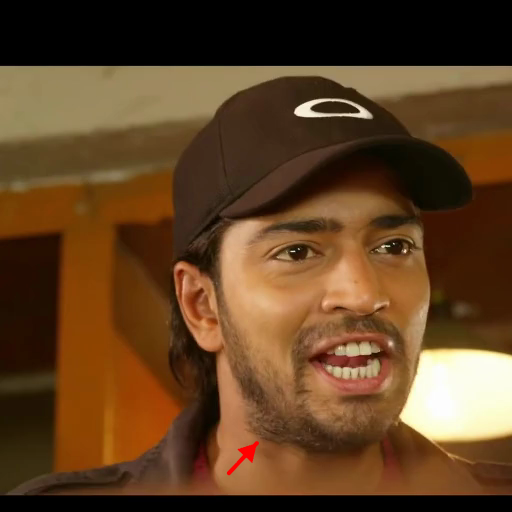} &
    \includegraphics[width=0.13\textwidth]{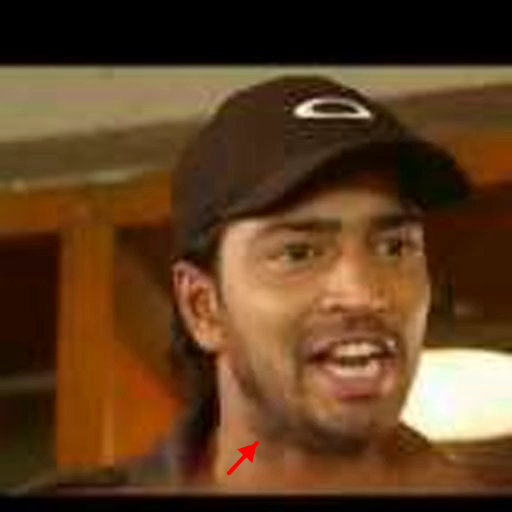} &
    \includegraphics[width=0.13\textwidth]{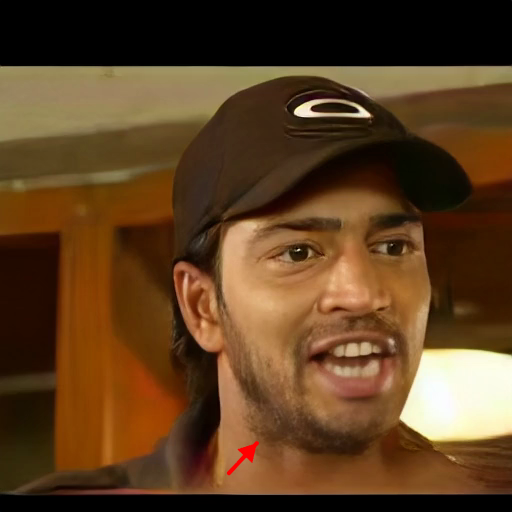} &
    \includegraphics[width=0.13\textwidth]{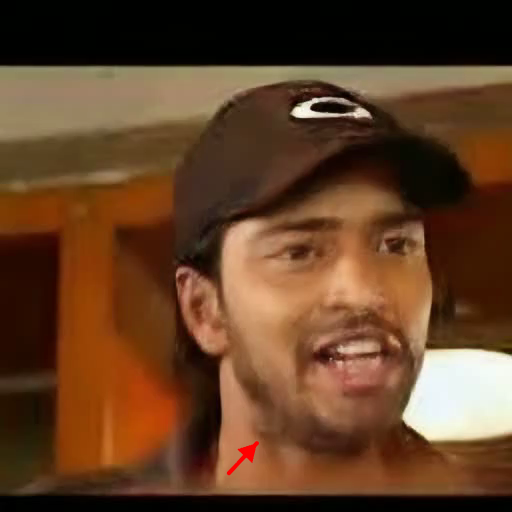} &
    \includegraphics[width=0.13\textwidth]{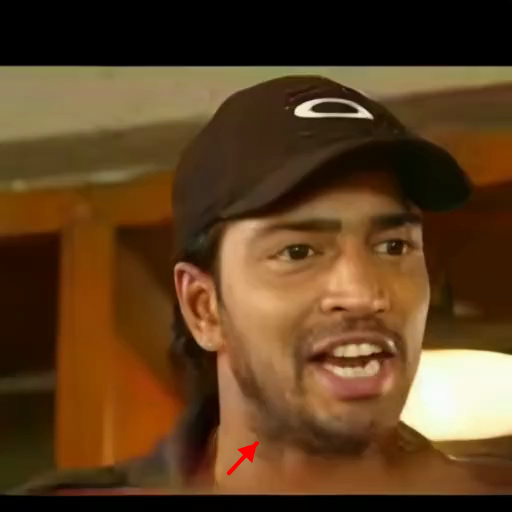} &
    \includegraphics[width=0.13\textwidth]{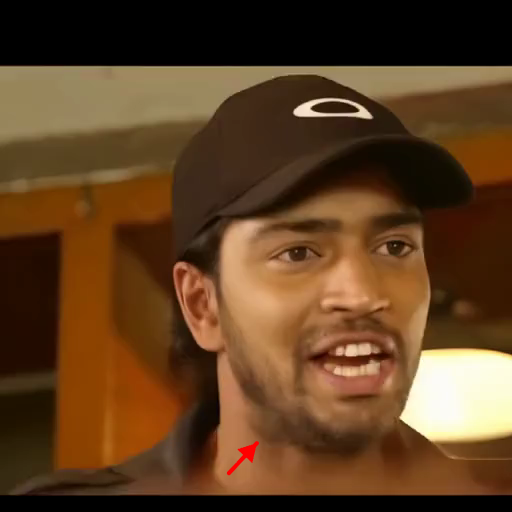} &
    \includegraphics[width=0.13\textwidth]{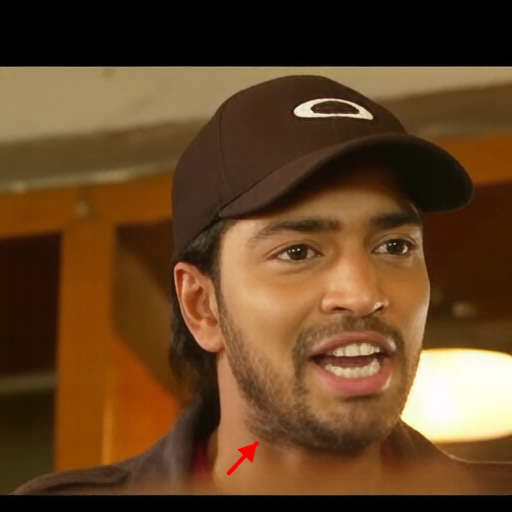} \\
    HQ (VFHQ-Test) & LQ & PGTFormer~\cite{xu2024beyond} &
    BFVR~\cite{wang2025efficient} & DicFace~\cite{chen2025dicface} &
    SVFR~\cite{wang2025svfr} & DVFace (ours) \\
    \end{tabular}
    \end{adjustbox}\par\vspace{.2mm}
\begin{adjustbox}{valign=t}
    \setlength{\tabcolsep}{0.8mm}
    \footnotesize
    \begin{tabular}{ccccccc}
    \includegraphics[width=0.13\textwidth]{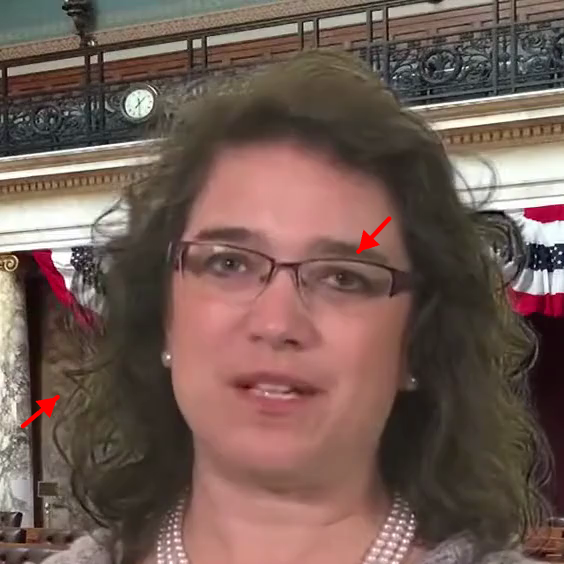} &
    \includegraphics[width=0.13\textwidth]{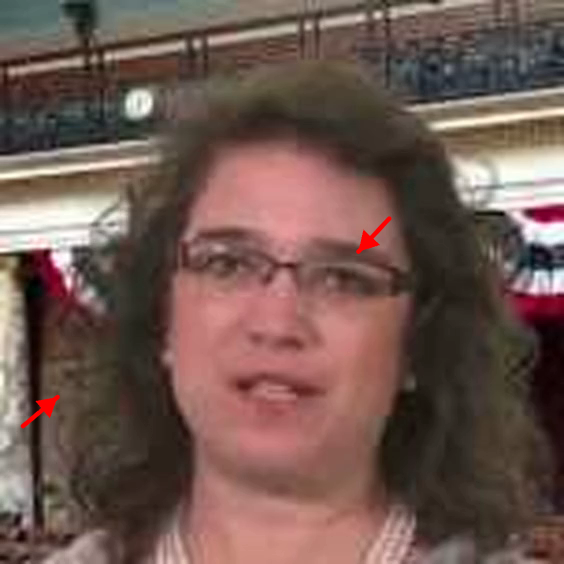} &
    \includegraphics[width=0.13\textwidth]{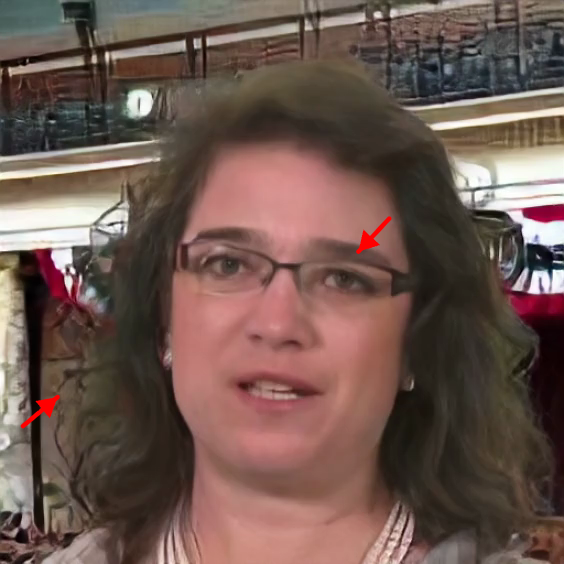} &
    \includegraphics[width=0.13\textwidth]{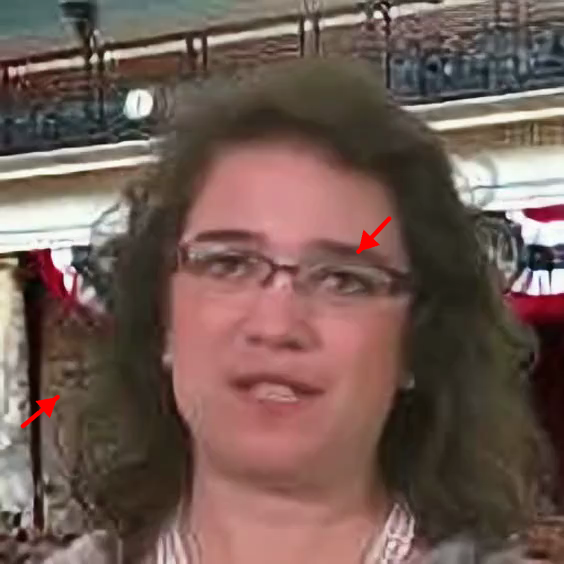} &
    \includegraphics[width=0.13\textwidth]{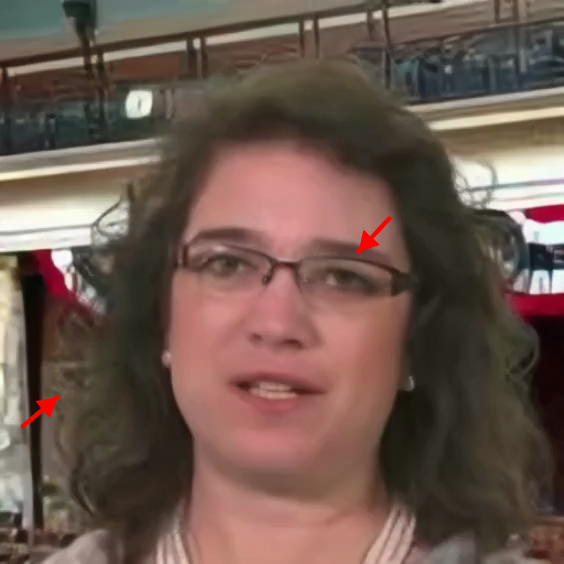} &
    \includegraphics[width=0.13\textwidth]{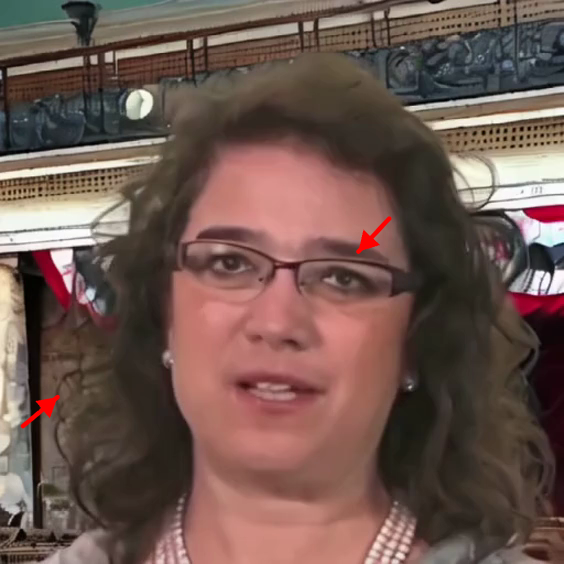} &
    \includegraphics[width=0.13\textwidth]{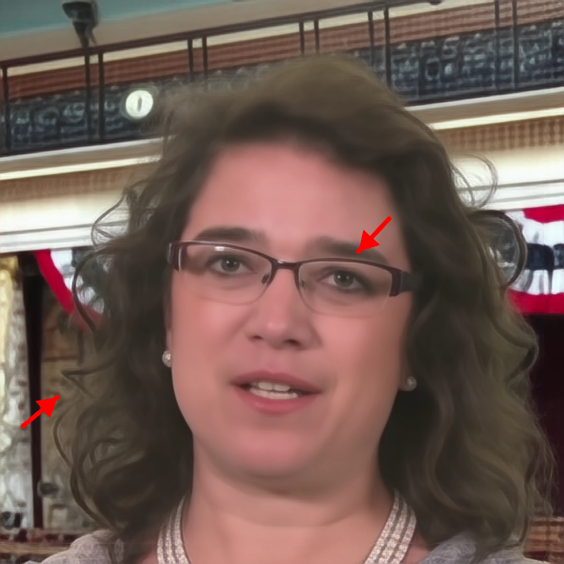} \\
    HQ (HDTF) & LQ & PGTFormer~\cite{xu2024beyond} &
    BFVR~\cite{wang2025efficient} & DicFace~\cite{chen2025dicface} &
    SVFR~\cite{wang2025svfr} & DVFace (ours) \\
    \end{tabular}
    \end{adjustbox}\par\vspace{.2mm}
\begin{adjustbox}{valign=t}
    \setlength{\tabcolsep}{0.8mm}
    \footnotesize
    \begin{tabular}{ccccccc}
    \includegraphics[width=0.13\textwidth]{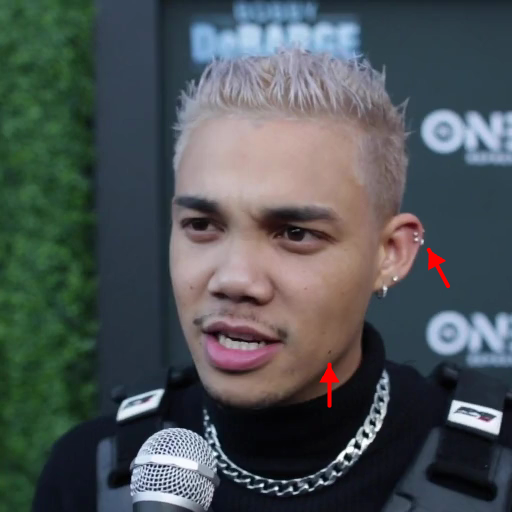} &
    \includegraphics[width=0.13\textwidth]{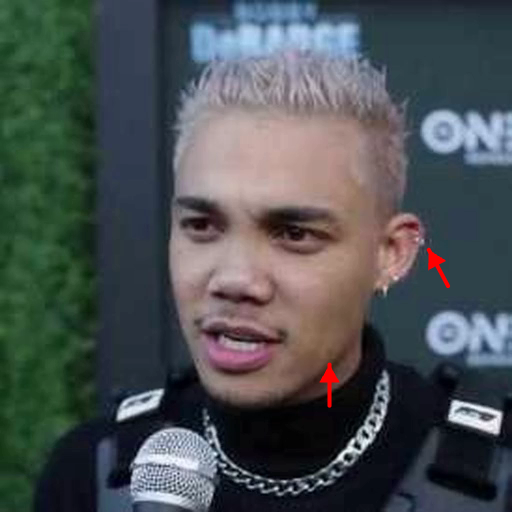} &
    \includegraphics[width=0.13\textwidth]{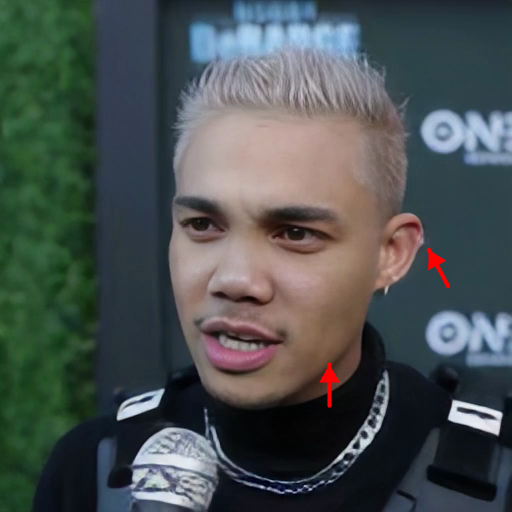} &
    \includegraphics[width=0.13\textwidth]{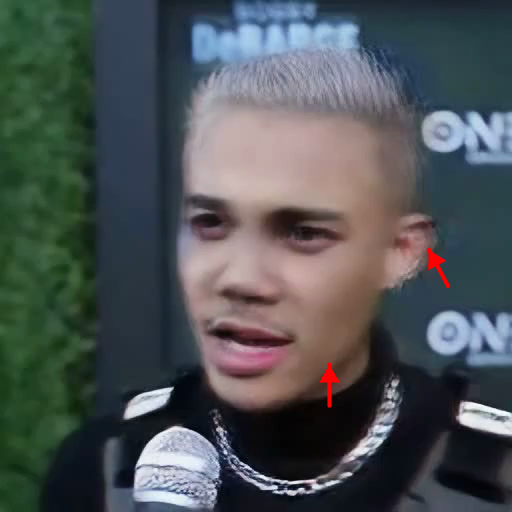} &
    \includegraphics[width=0.13\textwidth]{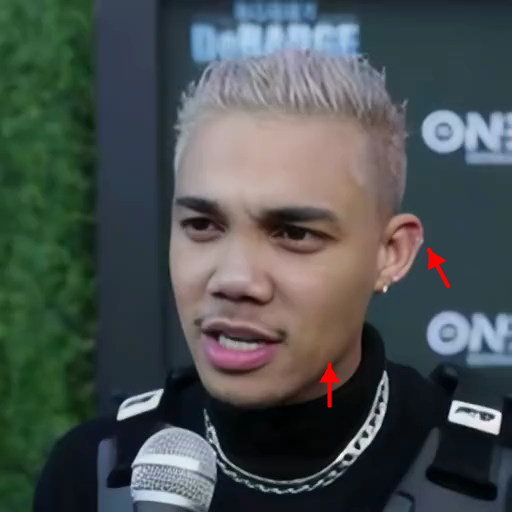} &
    \includegraphics[width=0.13\textwidth]{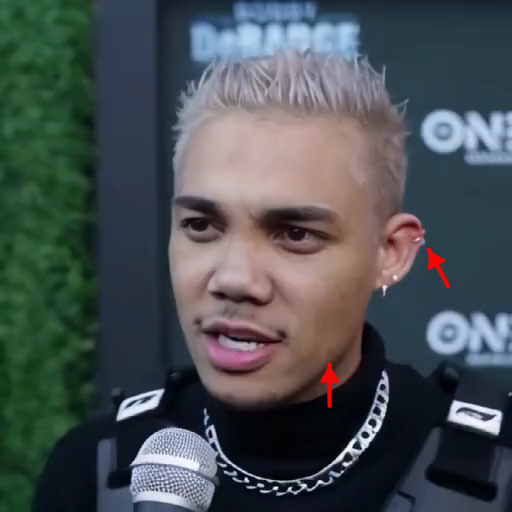} &
    \includegraphics[width=0.13\textwidth]{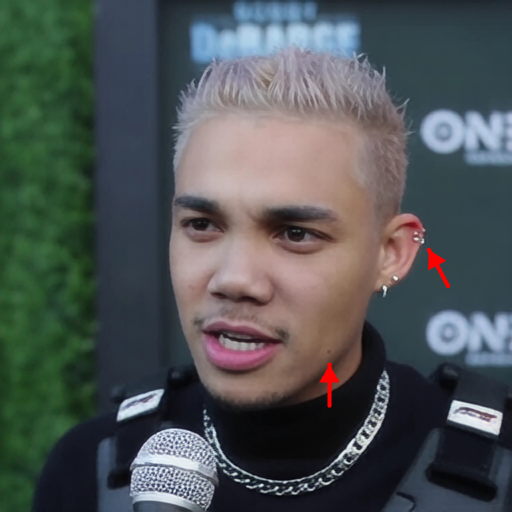} \\
    HQ (VFHQ-Test) & LQ & PGTFormer~\cite{xu2024beyond} &
    BFVR~\cite{wang2025efficient} & DicFace~\cite{chen2025dicface} &
    SVFR~\cite{wang2025svfr} & DVFace (ours) \\
    \end{tabular}
    \end{adjustbox}\par\vspace{.2mm}
\begin{adjustbox}{valign=t}
    \setlength{\tabcolsep}{0.8mm}
    \footnotesize
    \begin{tabular}{ccccccc}
    \includegraphics[width=0.13\textwidth]{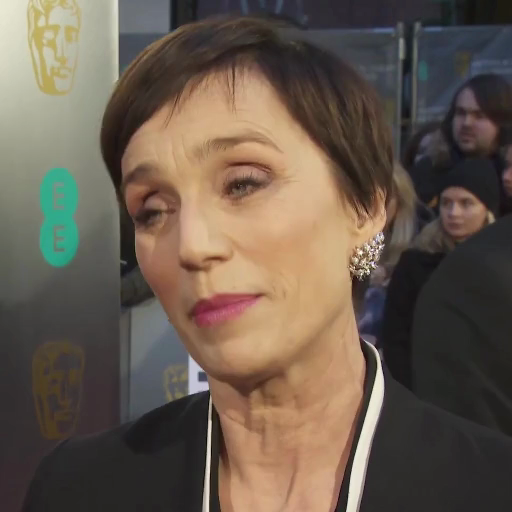} &
    \includegraphics[width=0.13\textwidth]{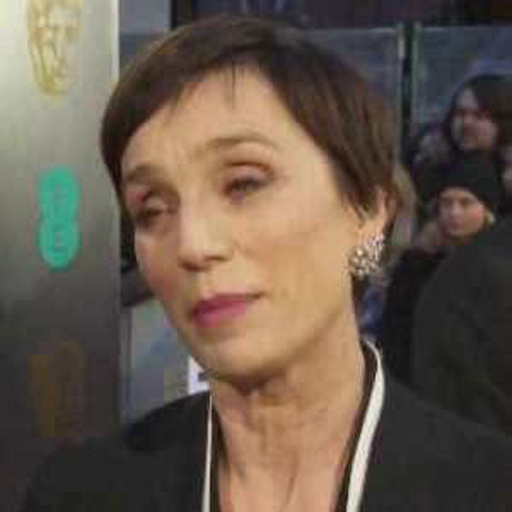} &
    \includegraphics[width=0.13\textwidth]{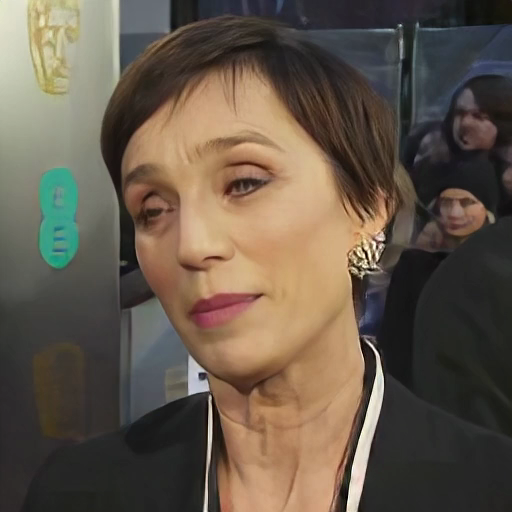} &
    \includegraphics[width=0.13\textwidth]{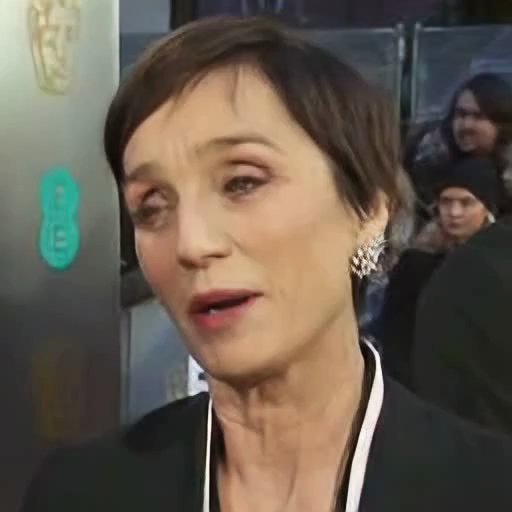} &
    \includegraphics[width=0.13\textwidth]{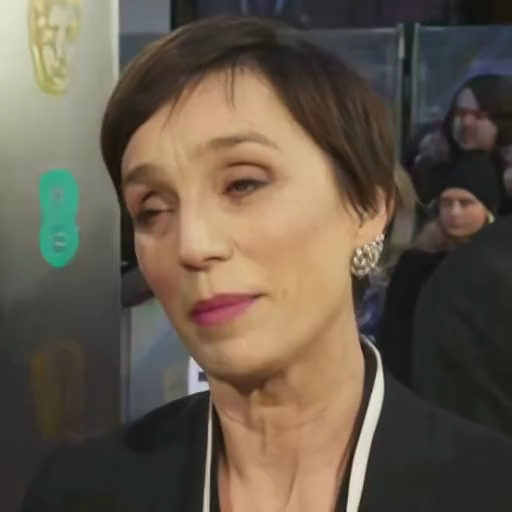} &
    \includegraphics[width=0.13\textwidth]{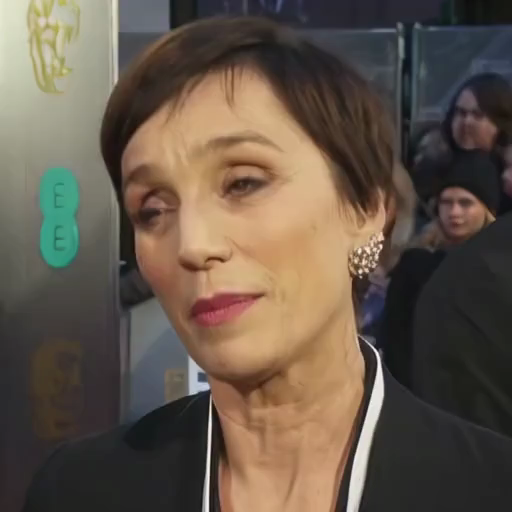} &
    \includegraphics[width=0.13\textwidth]{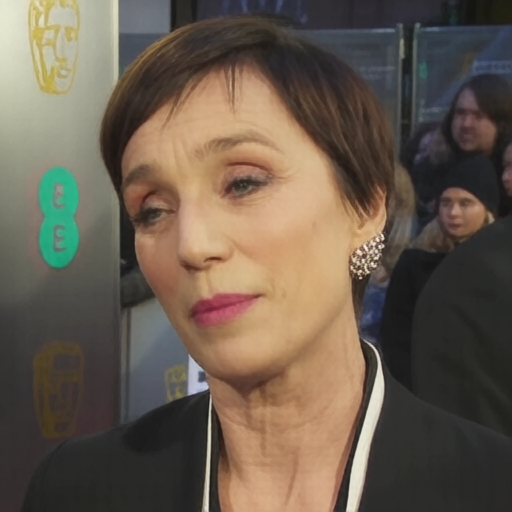} \\
    HQ (VFHQ-Test) & LQ & PGTFormer~\cite{xu2024beyond} &
    BFVR~\cite{wang2025efficient} & DicFace~\cite{chen2025dicface} &
    SVFR~\cite{wang2025svfr} & DVFace (ours) \\
    \end{tabular}
    \end{adjustbox}\par\vspace{.2mm}
\begin{adjustbox}{valign=t}
    \setlength{\tabcolsep}{0.8mm}
    \footnotesize
    \begin{tabular}{ccccccc}
    \includegraphics[width=0.13\textwidth]{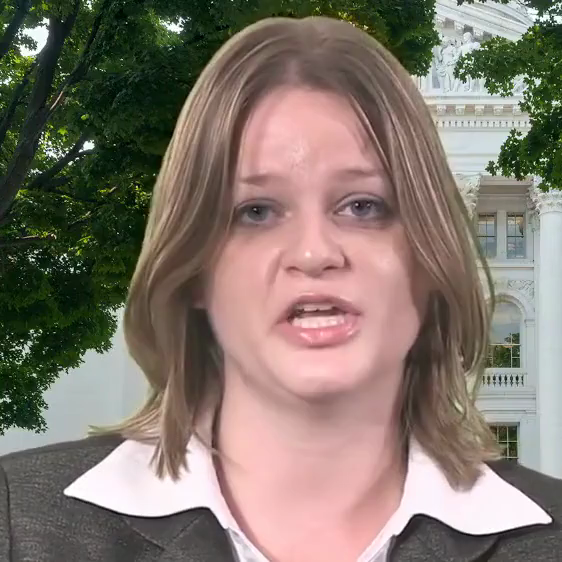} &
    \includegraphics[width=0.13\textwidth]{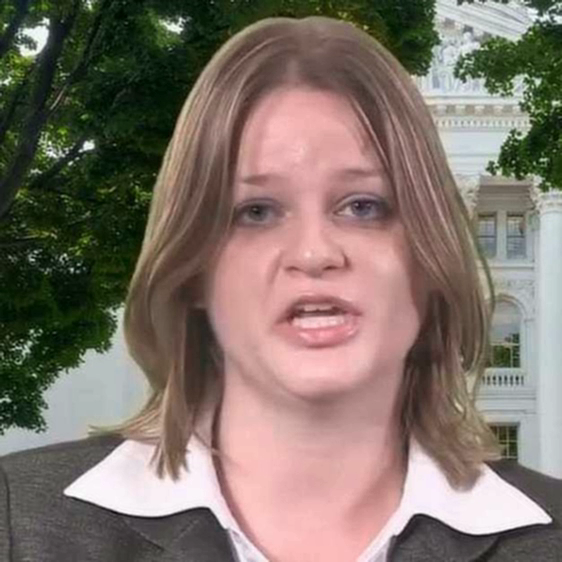} &
    \includegraphics[width=0.13\textwidth]{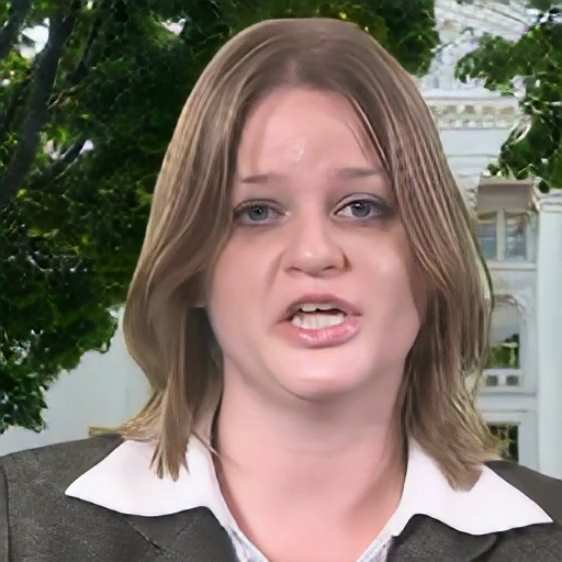} &
    \includegraphics[width=0.13\textwidth]{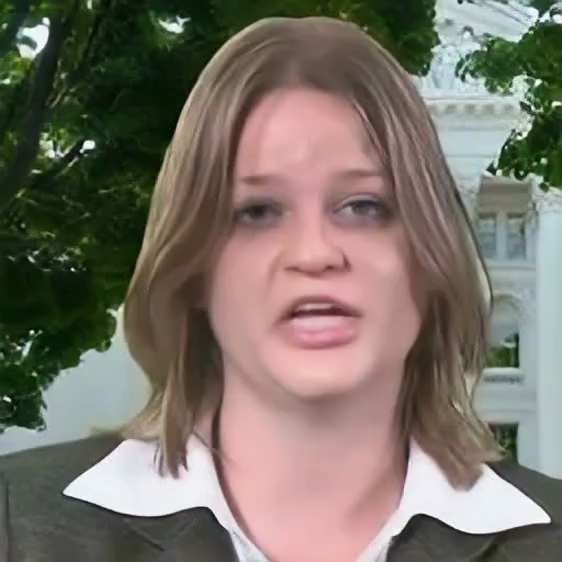} &
    \includegraphics[width=0.13\textwidth]{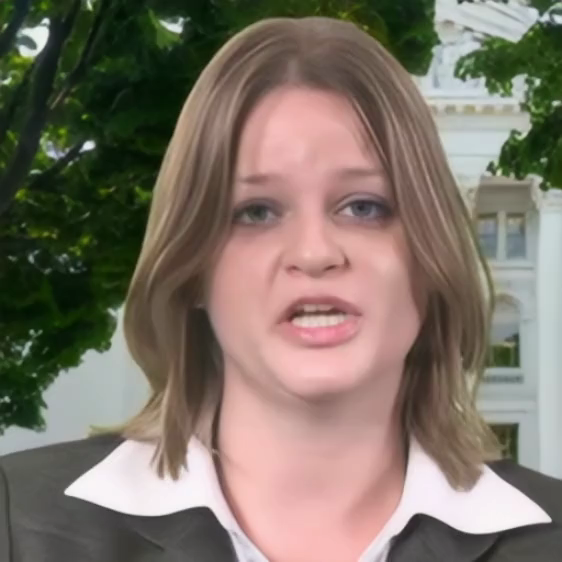} &
    \includegraphics[width=0.13\textwidth]{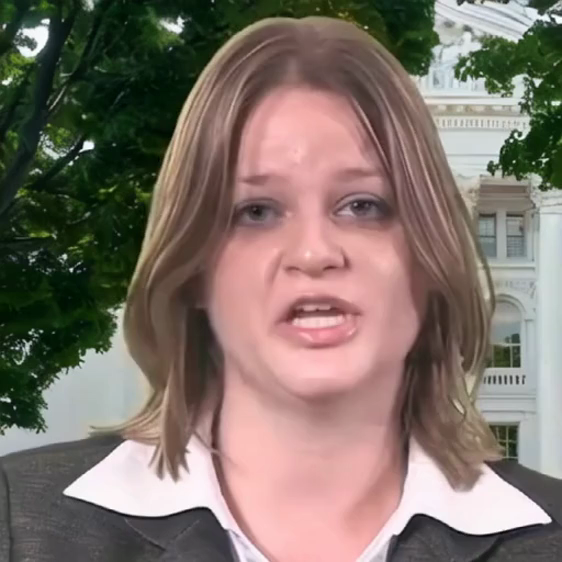} &
    \includegraphics[width=0.13\textwidth]{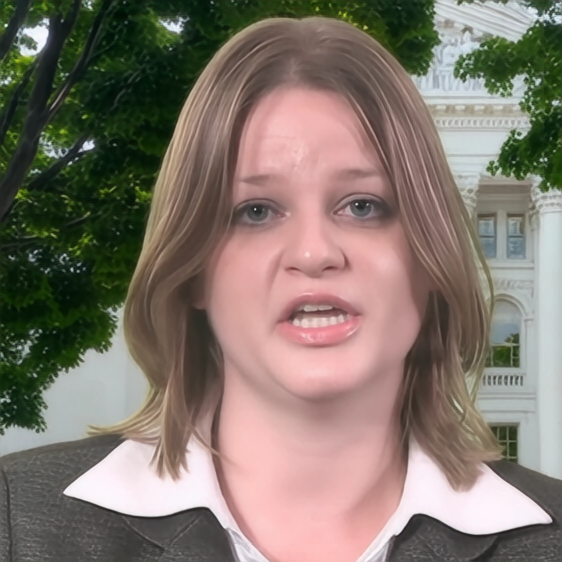} \\
    HQ (HDTF) & LQ & PGTFormer~\cite{xu2024beyond} &
    BFVR~\cite{wang2025efficient} & DicFace~\cite{chen2025dicface} &
    SVFR~\cite{wang2025svfr} & DVFace (ours) \\
    \end{tabular}
    \end{adjustbox}\par\vspace{.2mm}
\begin{adjustbox}{valign=t}
    \setlength{\tabcolsep}{0.8mm}
    \footnotesize
    \begin{tabular}{ccccccc}
    \includegraphics[width=0.13\textwidth]{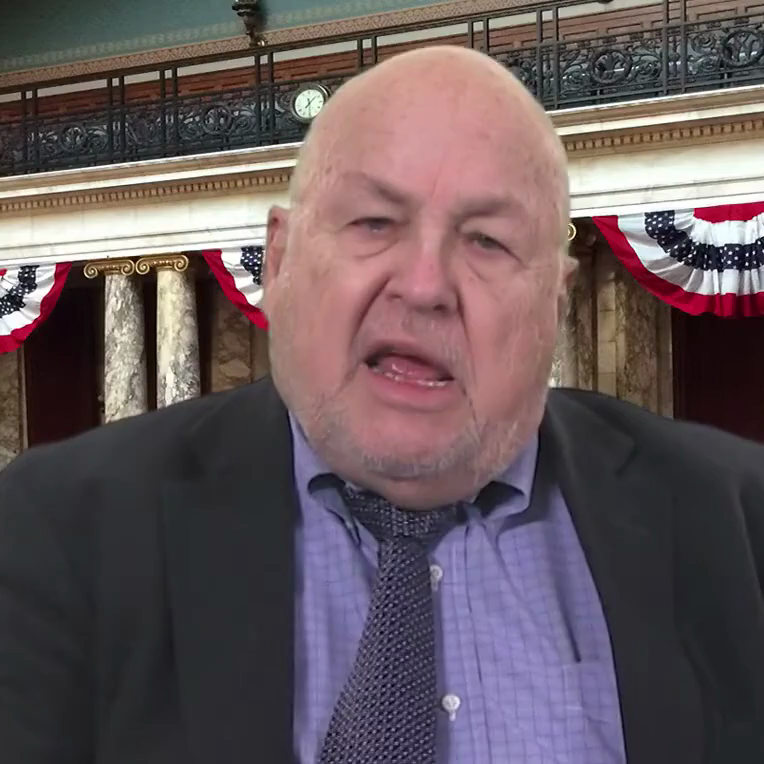} &
    \includegraphics[width=0.13\textwidth]{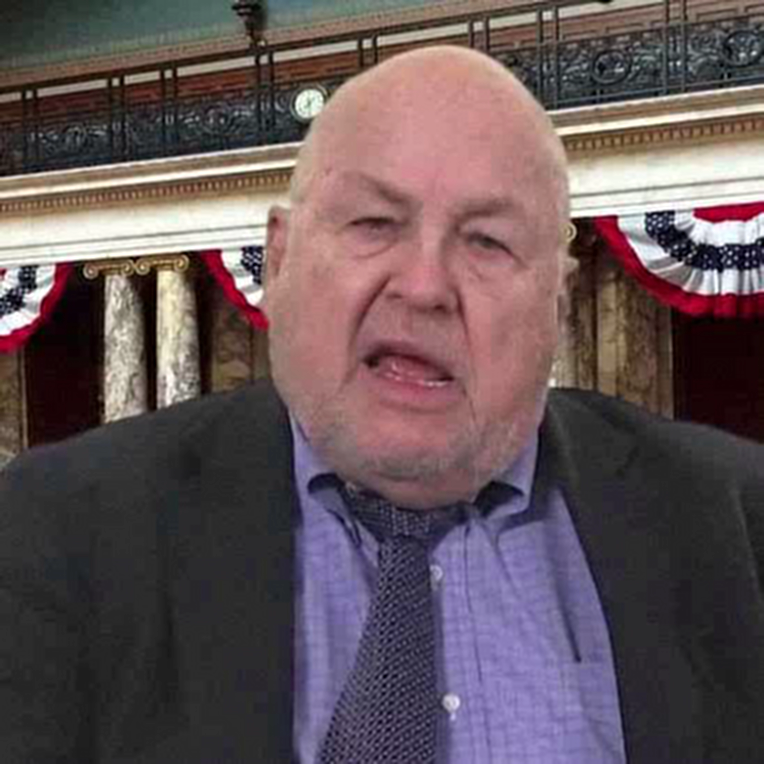} &
    \includegraphics[width=0.13\textwidth]{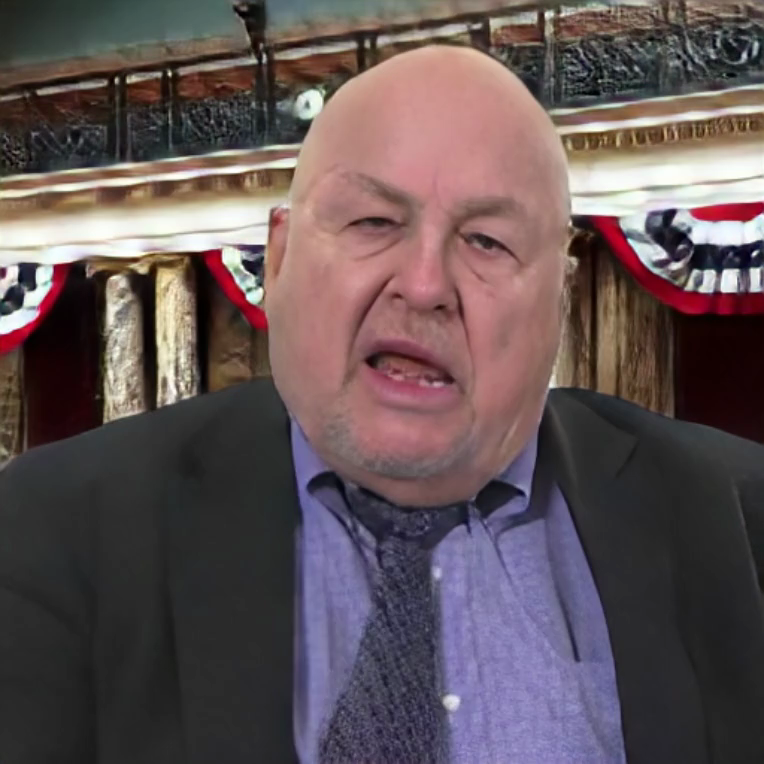} &
    \includegraphics[width=0.13\textwidth]{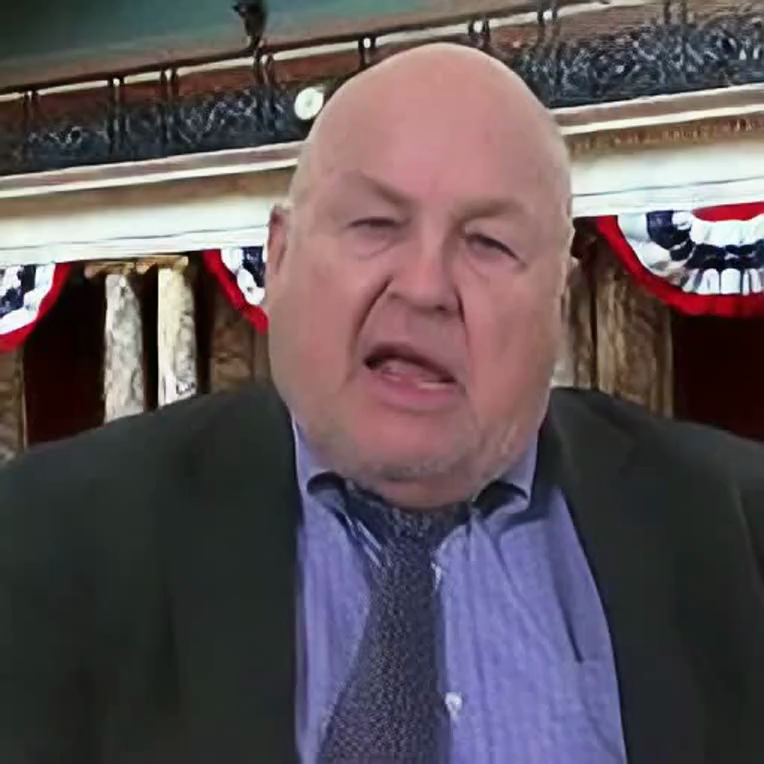} &
    \includegraphics[width=0.13\textwidth]{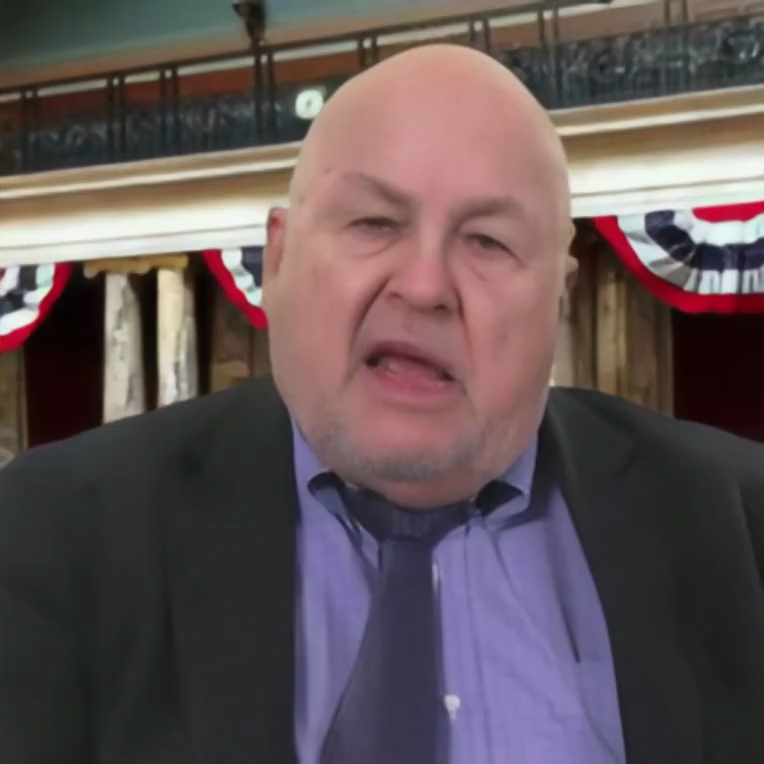} &
    \includegraphics[width=0.13\textwidth]{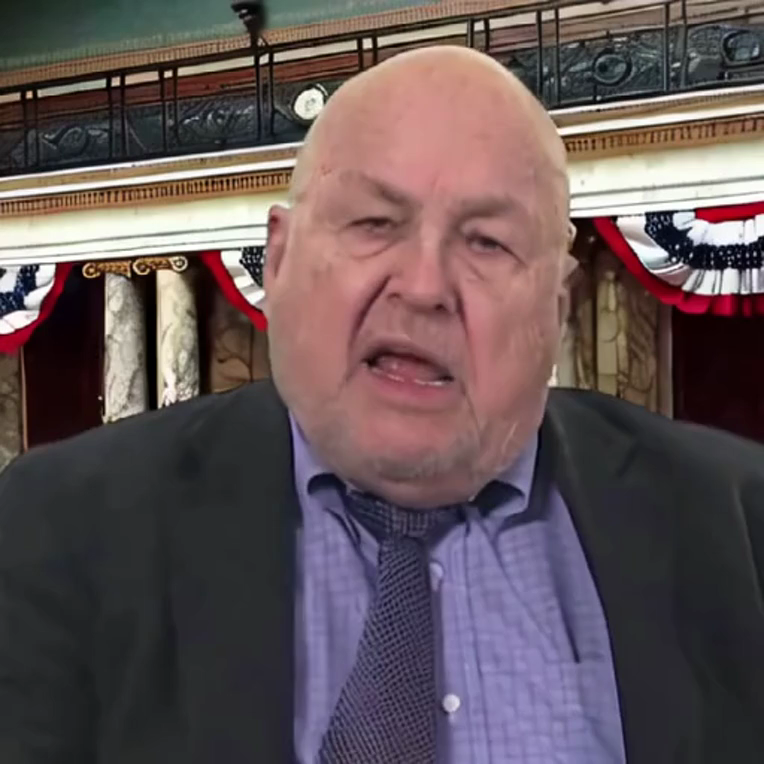} &
    \includegraphics[width=0.13\textwidth]{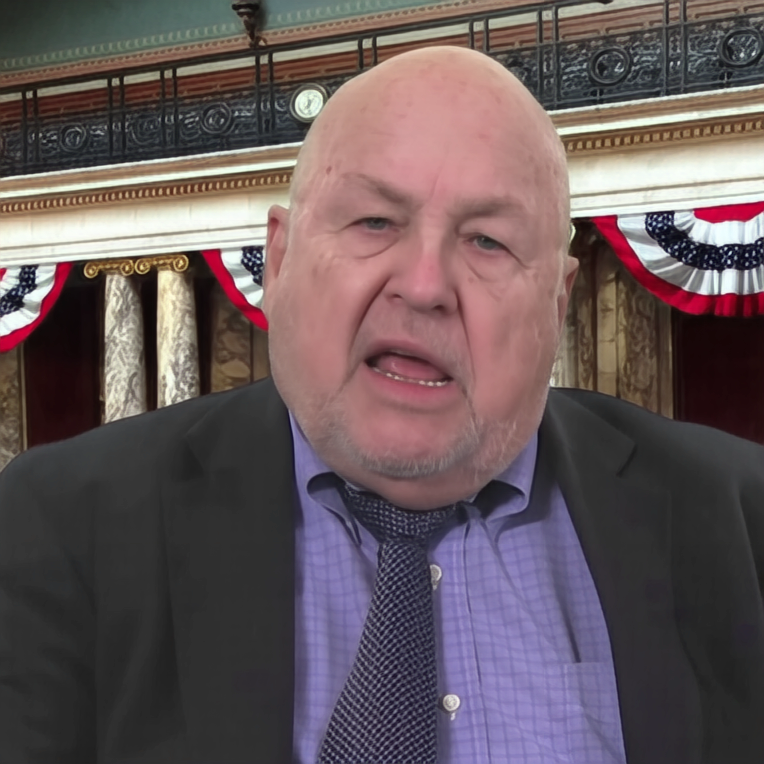} \\
    HQ (HDTF) & LQ & PGTFormer~\cite{xu2024beyond} &
    BFVR~\cite{wang2025efficient} & DicFace~\cite{chen2025dicface} &
    SVFR~\cite{wang2025svfr} & DVFace (ours) \\
    \end{tabular}
    \end{adjustbox}\par\vspace{.2mm}

\caption{
\added{Qualitative comparison on challenging synthetic cases. DVFace reconstructs facial structures and fine details consistent with the HQ references. Please zoom in for better visibility.}}
\vspace{-2.mm}
\label{fig:visual-synthetic}
\end{figure*}

\section{Experiments}
\vspace{1.mm}
\subsection{Experimental Settings}
\noindent \textbf{Dataset.} We train our model on the VFHQ~\cite{xie2022vfhq} dataset, which consists of 16,000 high-quality face videos. We conduct evaluations on both synthetic and real-world benchmarks. For efficient evaluation, we randomly sample a subset from each benchmark. The synthetic benchmarks include VFHQ-Test and 100 videos from HDTF~\cite{zhang2021flow}, to which we apply the same degradation pipeline used during training. For real-world evaluation, we use 50 videos from RFV-LQ~\cite{wang2024analysis} and 100 videos from VoxCeleb2~\cite{chung2018voxceleb2}.

\added{\noindent \textbf{Training Degradation.} Following OSDFace~\cite{wang2025osdface} and RealBasicVSR~\cite{chan2022investigating}, we synthesize LQ-HQ pairs with a two-stage degradation pipeline. For each HQ video, we sample 25 consecutive frames and crop them to $384\times384$. In the first stage, each frame is convolved with a randomly sampled blur kernel $k^{(1)}_{\sigma}$, downsampled by a factor $r^{(1)}$, corrupted by additive Gaussian noise $n^{(1)}_{\delta}$, and compressed by both JPEG and a video codec sampled from $\{\texttt{libx264},\texttt{h264},\texttt{mpeg4}\}$. The degraded sequence is then resized back to $384\times384$. We sample the degradation parameters as $\sigma^{(1)}\sim\mathcal{U}(21,31)$, $r^{(1)}\sim\mathcal{U}(4,10)$, $\delta^{(1)}\sim\mathcal{U}(0,10)$, JPEG quality $q^{(1)}\sim\mathcal{U}(40,90)$, and video bitrate $b^{(1)}\sim\mathcal{U}(10^{4},10^{5})$. In the second stage, we apply milder Gaussian noise, JPEG compression, and video compression, where $\delta^{(2)}\sim\mathcal{U}(0,3)$, $q^{(2)}\sim\mathcal{U}(60,100)$, and $b^{(2)}\sim\mathcal{U}(10^{4},10^{5})$. Here, $\mathcal{U}$ denotes a uniform distribution over the specified parameter range.}

\noindent \textbf{Evaluation Metrics.} We use a diverse set of video quality assessment metrics to evaluate our model from multiple perspectives. For \textbf{\textit{fidelity}}, we use PSNR and SSIM~\cite{wang2004image}. For \textbf{\textit{perceptual quality}}, we adopt LPIPS~\cite{zhang2018unreasonable}, DISTS~\cite{ding2020image}, CLIP-IQA~\cite{wang2023exploring}, MUSIQ~\cite{ke2021musiq}, NIQE~\cite{zhang2015feature}, MANIQA~\cite{yang2022maniqa}, and LIQE~\cite{zhang2023blind}. To measure \textbf{\textit{identity preservation}}, we use the average keypoint distance (AKD) \added{ and identity similarity score (IDS). IDS is the average cosine similarity between the 512-D ArcFace~\cite{deng2019arcface} features of restored and ground-truth frames}. For overall \textbf{video quality}, we adopt FVD~\cite{unterthiner2019fvd} and DOVER~\cite{wu2023exploring}. In addition, we evaluate \textbf{\textit{temporal consistency}} using $E^*_{warp}$~\cite{lai2018learning} and VIDD~\cite{chen2024towards}.

\noindent \textbf{Implementation Details.} DVFace is based on the T2V model Wan2.1~\cite{wan2025wan} and optimized in two stages. During the first stage, the prior extraction module undergoes 250k iterations for codebook learning, followed by 50k iterations of LQ inputs refinement, with the learning rate set to $lr = 8\times10^{-5}$. The loss coefficients are set to $\beta=0.25$, $\lambda_{adv}=0.8$, and $\lambda_{ce}=0.5$. During the second stage, we adapt the diffusion backbone for 15k iterations with $lr = 3\times10^{-5}$. The loss weight $\lambda_{temp}$ is fixed to 0.1. Training is conducted on 4 NVIDIA A6000 GPUs \added{throughout both stages} using AdamW with a total batch size of 4 \added{for all experiments}.

\subsection{Comparison with State-of-the-Art Methods}
\vspace{2.mm}
We evaluate DVFace \added{on several synthetic and real-world datasets} against several recent leading methods for video face restoration, including: PGTFormer~\cite{xu2024beyond}, KEEP~\cite{feng2024kalman}, AverNet~\cite{zhao2024avernet}, BFVR~\cite{wang2025efficient}, DicFace~\cite{chen2025dicface}, and SVFR~\cite{wang2025svfr}.

\vspace{0.5mm}
\noindent \textbf{Evaluation on Synthetic Datasets.}
We report the quantitative and qualitative comparisons on the synthetic datasets VFHQ and HDTF in Tab.~\ref{tab:quant-synthetic} and Fig.~\ref{fig:visual-synthetic}, respectively. Overall, DVFace shows the best or highly competitive performance \added{across both evaluation datasets} on most metrics. For instance, on VFHQ, DVFace \added{consistently} ranks first on all metrics except NIQE, obtaining PSNR/DOVER of 31.81/0.8703. This strong performance indicates that our method can effectively restore fine facial content while maintaining identity fidelity and temporal consistency. We can also observe from Fig.~\ref{fig:performance} that, in the first case, DVFace reconstructs clearer facial structures and details, while the competing methods either over-smooth details or introduce artifacts. Similar phenomena can be observed in the second case, \added{where DVFace likewise preserves finer facial details}.

\begin{table*}[t]
    \footnotesize
    \centering
    \caption{Quantitative comparison of video face restoration models on real-world datasets (RFV-LQ and Voxceleb2). The best and second-best results are colored with \textcolor{red}{red} and \textcolor{blue}{blue}. $\uparrow$ indicates higher is better, and $\downarrow$ indicates lower is better.}
    \vspace{-2.mm}
    \label{tab:quant-real}
        \setlength{\tabcolsep}{2.5mm}
        \begin{tabular}{l|l|c|c|c|c|c|c|c}
            \toprule[0.1em]
            \rowcolor{color-ours} Dataset & Metric & KEEP \cite{feng2024kalman} & AverNet \cite{zhao2024avernet} & PGTFormer \cite{xu2024beyond} & BFVR \cite{wang2025efficient} & DicFace \cite{chen2025dicface} & SVFR \cite{wang2025svfr} & DVFace (ours) \\
            \midrule[0.1em]
            \multirow{8}{*}{RFV-LQ}
                & CLIP-IQA $\uparrow$   & 0.3364 & 0.2969 & \textcolor{blue}{0.4880} & 0.3586 & 0.2824 & 0.4012 & \textcolor{red}{0.6256} \\
                & MUSIQ $\uparrow$      & 51.30 & 35.18 & \textcolor{blue}{59.55} & 44.94 & 42.66 & 52.66 & \textcolor{red}{63.69} \\
                & NIQE $\downarrow$     & 6.79 & 8.50 & \textcolor{red}{5.89} & {6.59} & 7.54 & 6.65 & \textcolor{blue}{6.03} \\
                & MANIQA $\uparrow$     & 0.2834 & 0.1854 & \textcolor{blue}{0.3092} & 0.2142 & 0.2045 & 0.2587 & \textcolor{red}{0.3422} \\
                & LIQE $\uparrow$       & 1.83 & 1.06 & \textcolor{blue}{2.34} & 1.26 & 1.03 & 1.65 & \textcolor{red}{2.80} \\
                & DOVER $\uparrow$      & 0.4638 & 0.2205 & \textcolor{blue}{0.6412} & 0.3528 & 0.4679 & 0.4742 & \textcolor{red}{0.7165} \\
                & $E^*_{warp}$ $\downarrow$    & 0.82 & \textcolor{red}{0.69} & 1.02 & 2.12 & \textcolor{blue}{0.81} & 1.11 & 0.89 \\
                & VIDD $\downarrow$     & 0.3681 & \textcolor{blue}{0.3165} & 0.3464 & 0.4110 & 0.3421 & 0.3355 & \textcolor{red}{0.3159} \\
            \midrule
            \multirow{8}{*}{Voxceleb2}
                & CLIP-IQA $\uparrow$   & 0.4704 & 0.3882 & 0.4881 & 0.4581 & 0.4538 & \textcolor{blue}{0.5003} & \textcolor{red}{0.6411} \\
                & MUSIQ $\uparrow$      & 46.27 & 33.61 & 50.48 & 39.37 & 44.49 & \textcolor{blue}{51.05} & \textcolor{red}{53.19} \\
                & NIQE $\downarrow$     & 7.15 & 7.48 & 7.07 & \textcolor{blue}{6.76} & 7.68 & 7.21 & \textcolor{red}{6.59} \\
                & MANIQA $\uparrow$     & 0.3814 & 0.2694 & 0.3871 & 0.2967 & 0.3436 & \textcolor{red}{0.3955} & \textcolor{blue}{0.3894} \\
                & LIQE $\uparrow$       & 2.98 & 1.82 & 3.35 & 1.99 & 2.84 & \textcolor{blue}{3.63} & \textcolor{red}{4.13} \\
                & DOVER $\uparrow$      & 0.4019 & 0.1934 & \textcolor{red}{0.5136} & 0.2599 & 0.3549 & \textcolor{blue}{0.4869} & 0.4697 \\
                & $E^*_{warp}$ $\downarrow$    & 1.66 & \textcolor{blue}{1.49} & 1.77 & 3.72 & 1.99 & 1.97 & \textcolor{red}{1.45} \\
                & VIDD $\downarrow$     & 0.6460 & 0.6235 & \textcolor{blue}{0.5760} & 0.7674 & 0.6296 & 0.5809 & \textcolor{red}{0.4795} \\
            \bottomrule[0.1em]
        \end{tabular}
\end{table*}

\begin{figure*}[t]
\small
\centering
\begin{adjustbox}{valign=t}
    \setlength{\tabcolsep}{0.8mm}
    \footnotesize
    \begin{tabular}{ccccccc}
    \includegraphics[width=0.13\textwidth]{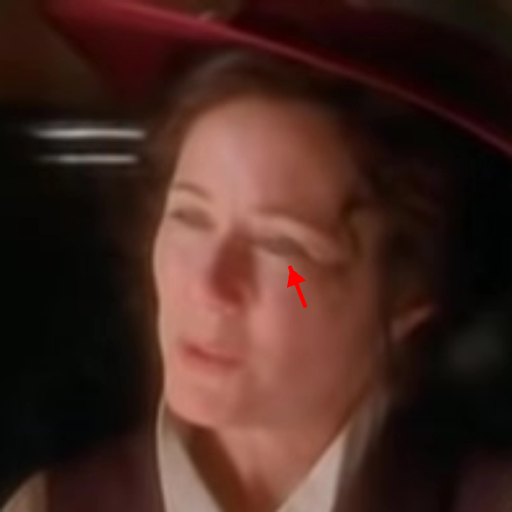} &
    \includegraphics[width=0.13\textwidth]{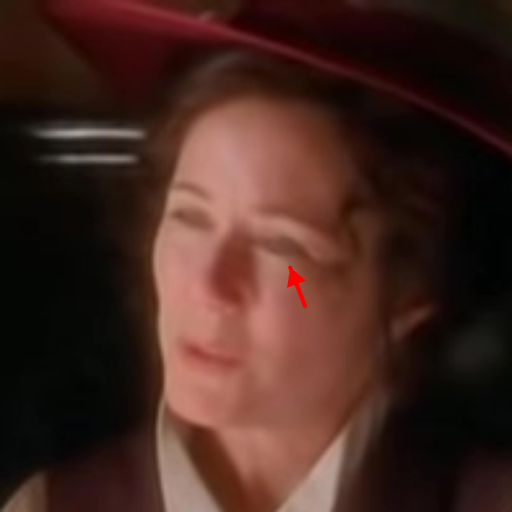} &
    \includegraphics[width=0.13\textwidth]{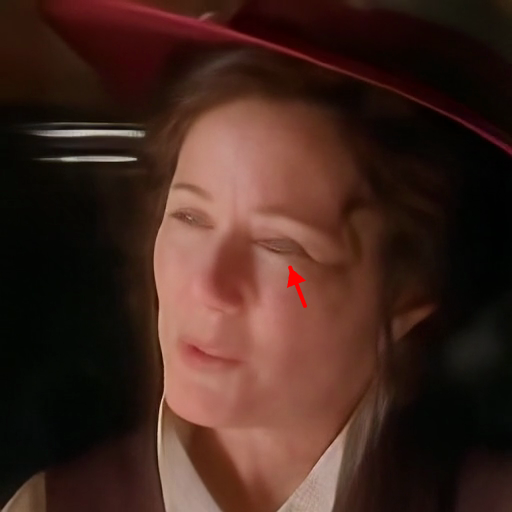} &
    \includegraphics[width=0.13\textwidth]{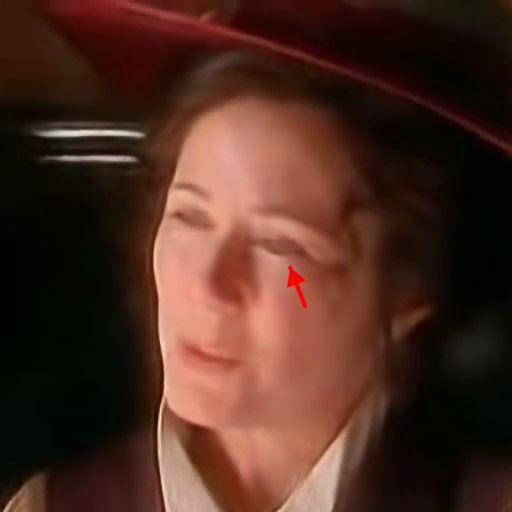} &
    \includegraphics[width=0.13\textwidth]{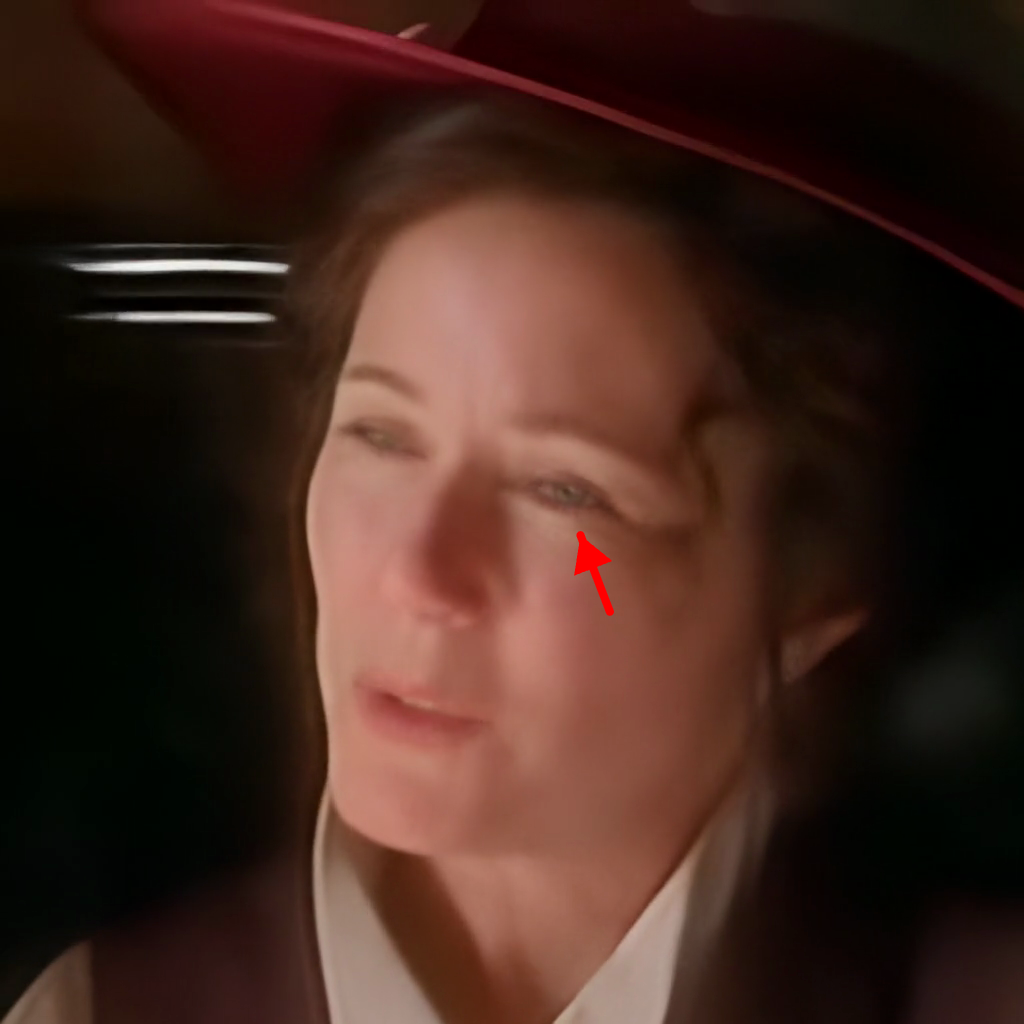} &
    \includegraphics[width=0.13\textwidth]{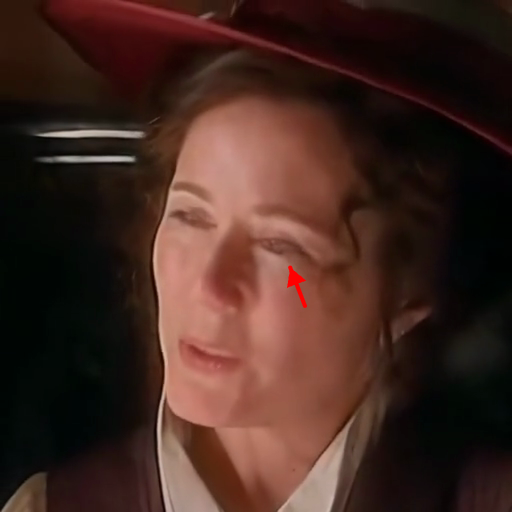} &
    \includegraphics[width=0.13\textwidth]{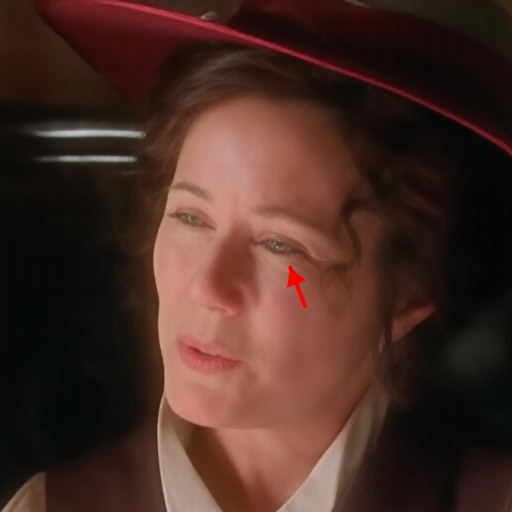} \\
    LQ (RFV-LQ) & AverNet~\cite{zhao2024avernet} & PGTFormer~\cite{xu2024beyond} &
    BFVR~\cite{wang2025efficient} & DicFace~\cite{chen2025dicface} &
    SVFR~\cite{wang2025svfr} & DVFace (ours) \\
    \end{tabular}
    \end{adjustbox}\par\vspace{0.2mm}
\begin{adjustbox}{valign=t}
    \setlength{\tabcolsep}{0.8mm}
    \footnotesize
    \begin{tabular}{ccccccc}
    \includegraphics[width=0.13\textwidth]{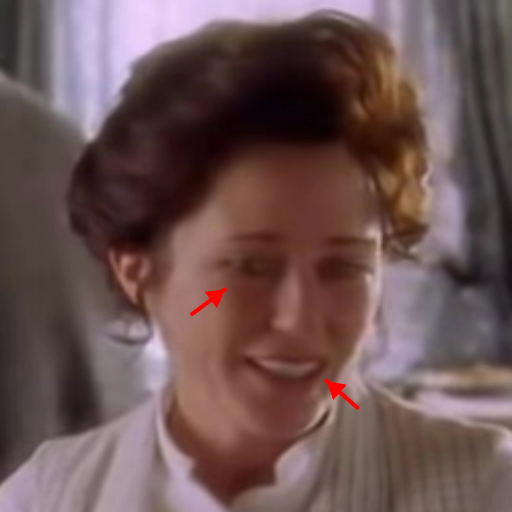} &
    \includegraphics[width=0.13\textwidth]{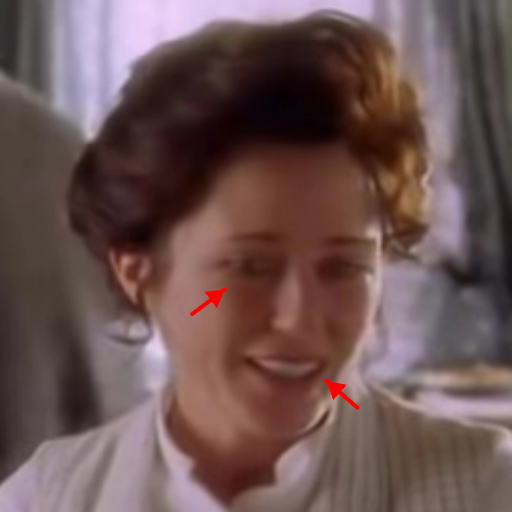} &
    \includegraphics[width=0.13\textwidth]{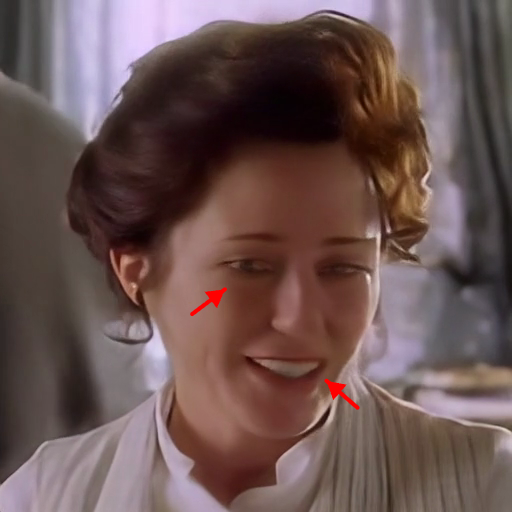} &
    \includegraphics[width=0.13\textwidth]{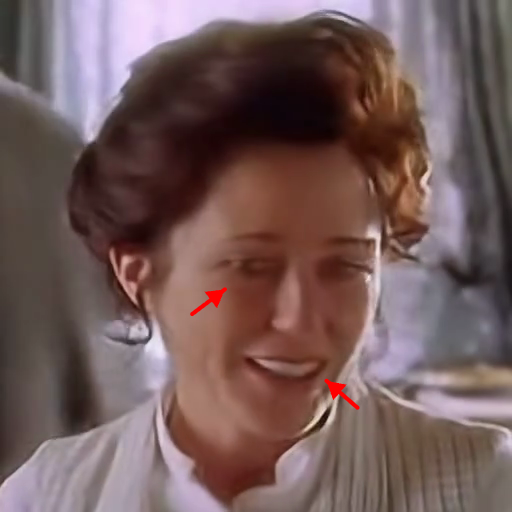} &
    \includegraphics[width=0.13\textwidth]{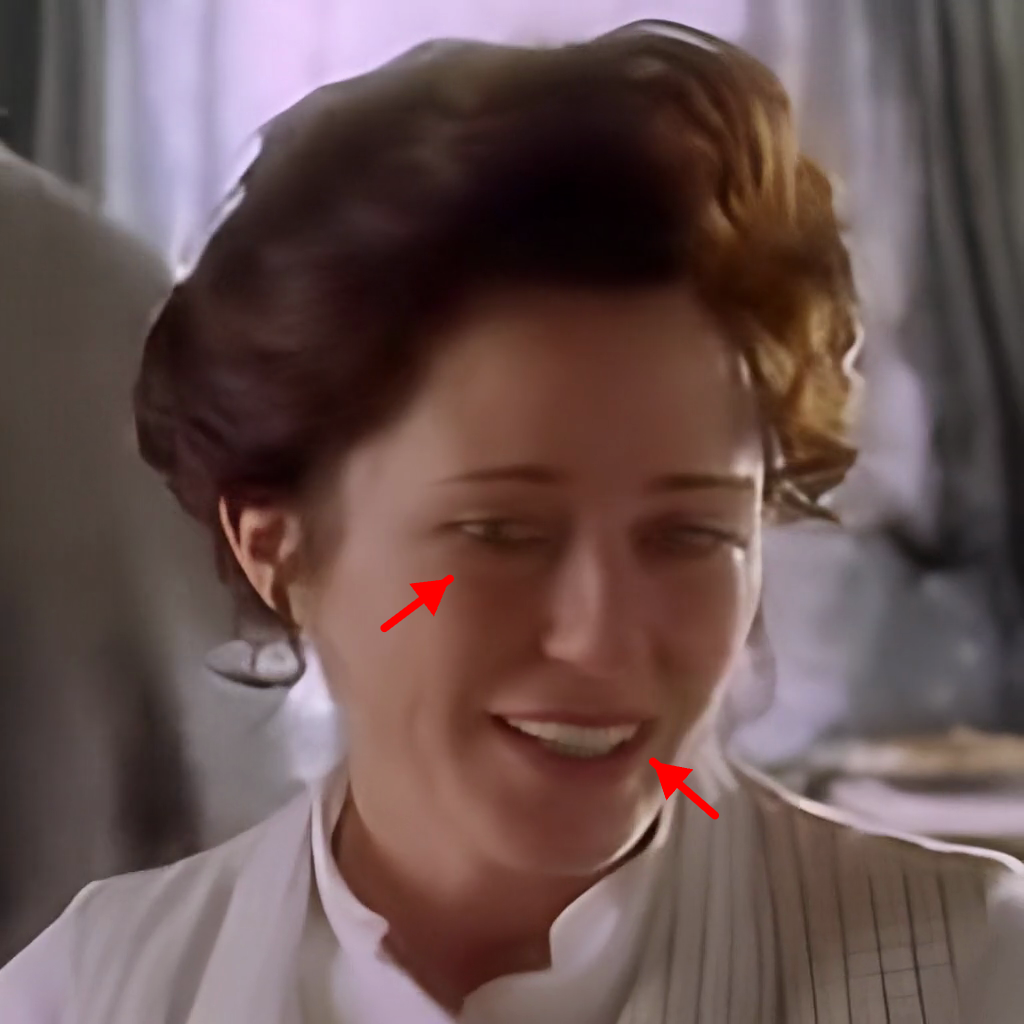} &
    \includegraphics[width=0.13\textwidth]{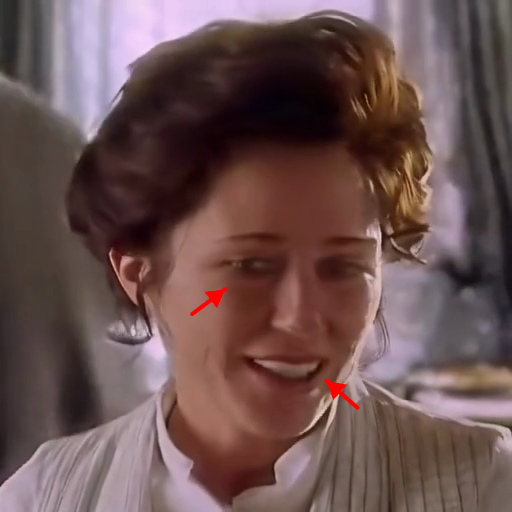} &
    \includegraphics[width=0.13\textwidth]{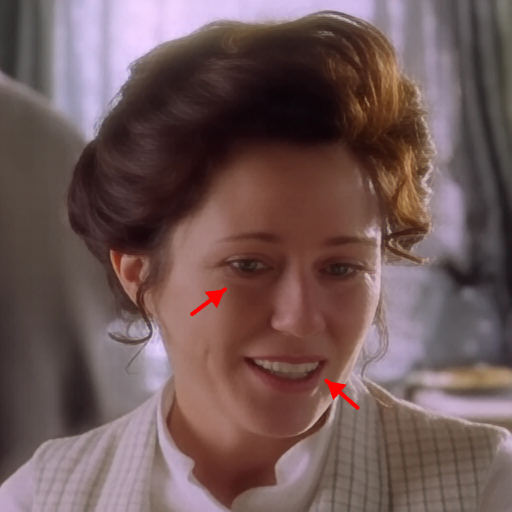} \\
    LQ (RFV-LQ) & AverNet~\cite{zhao2024avernet} & PGTFormer~\cite{xu2024beyond} &
    BFVR~\cite{wang2025efficient} & DicFace~\cite{chen2025dicface} &
    SVFR~\cite{wang2025svfr} & DVFace (ours) \\
    \end{tabular}
    \end{adjustbox}\par\vspace{0.2mm}
\begin{adjustbox}{valign=t}
    \setlength{\tabcolsep}{0.8mm}
    \footnotesize
    \begin{tabular}{ccccccc}
    \includegraphics[width=0.13\textwidth]{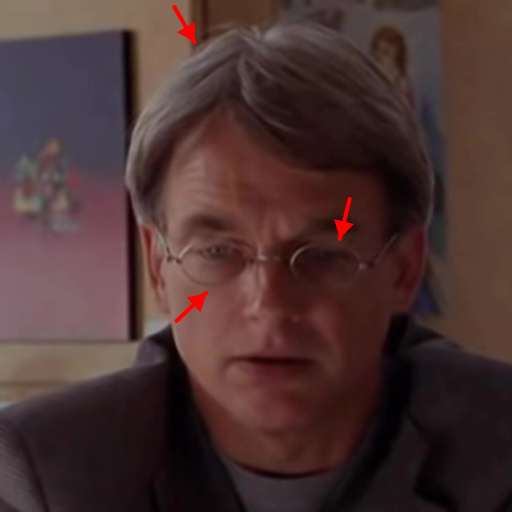} &
    \includegraphics[width=0.13\textwidth]{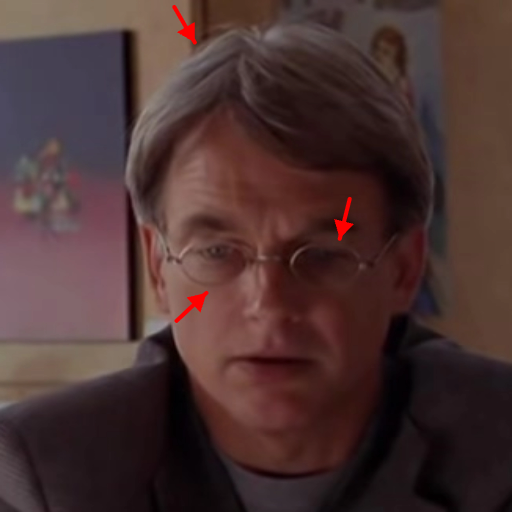} &
    \includegraphics[width=0.13\textwidth]{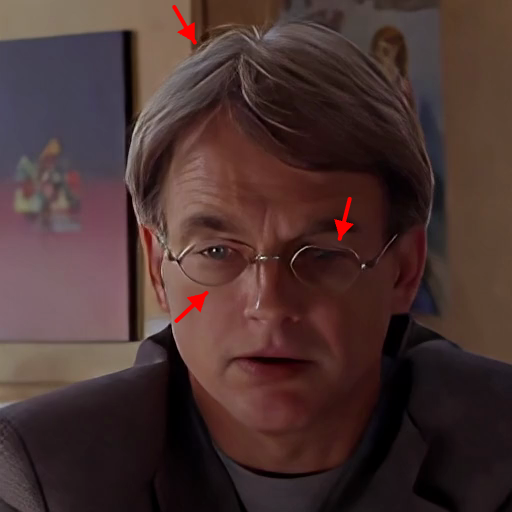} &
    \includegraphics[width=0.13\textwidth]{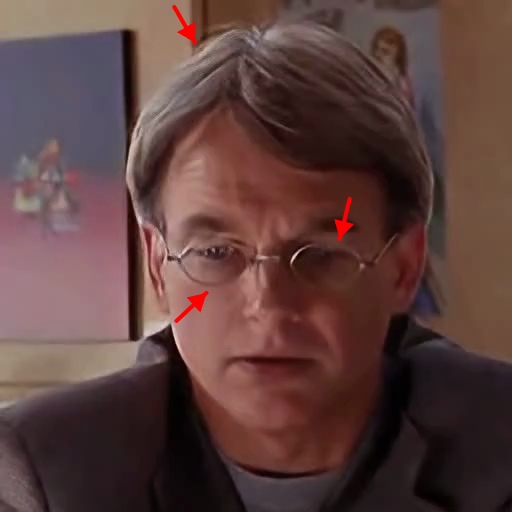} &
    \includegraphics[width=0.13\textwidth]{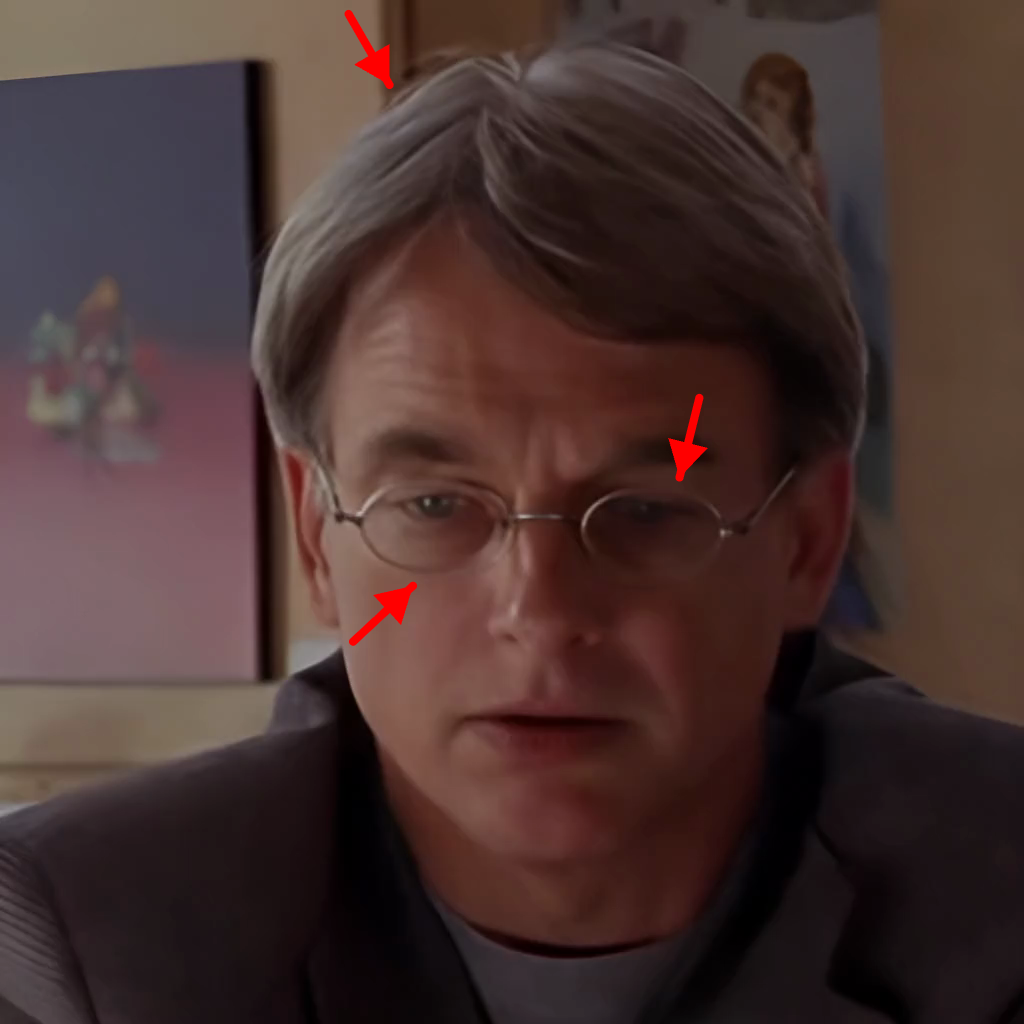} &
    \includegraphics[width=0.13\textwidth]{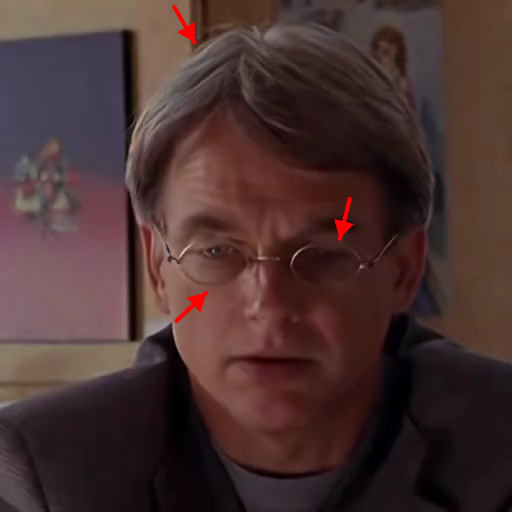} &
    \includegraphics[width=0.13\textwidth]{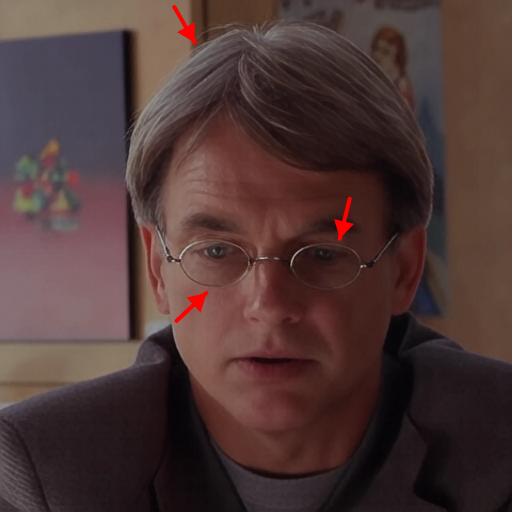} \\
    LQ (RFV-LQ) & AverNet~\cite{zhao2024avernet} & PGTFormer~\cite{xu2024beyond} &
    BFVR~\cite{wang2025efficient} & DicFace~\cite{chen2025dicface} &
    SVFR~\cite{wang2025svfr} & DVFace (ours) \\
    \end{tabular}
    \end{adjustbox}\par\vspace{0.2mm}
\begin{adjustbox}{valign=t}
    \setlength{\tabcolsep}{0.8mm}
    \footnotesize
    \begin{tabular}{ccccccc}
    \includegraphics[width=0.13\textwidth]{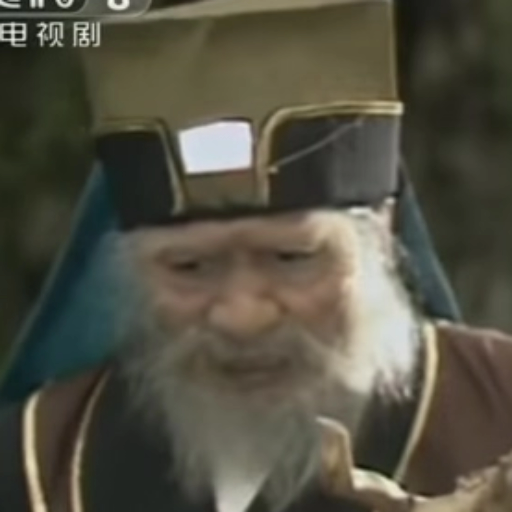} &
    \includegraphics[width=0.13\textwidth]{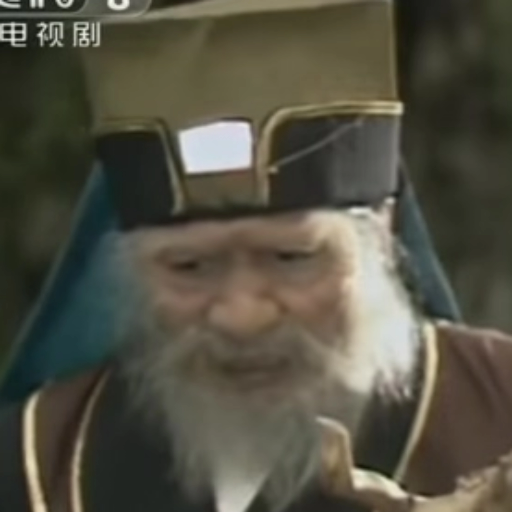} &
    \includegraphics[width=0.13\textwidth]{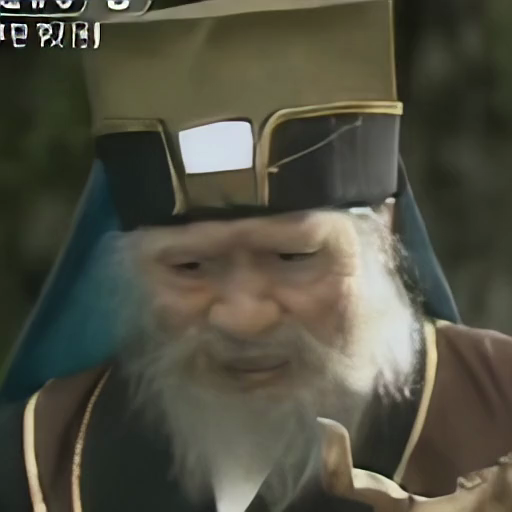} &
    \includegraphics[width=0.13\textwidth]{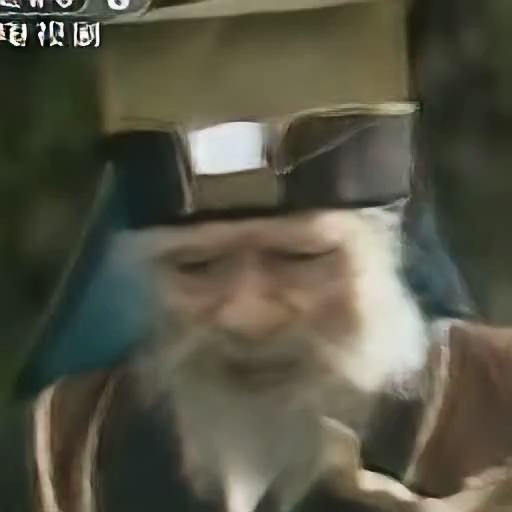} &
    \includegraphics[width=0.13\textwidth]{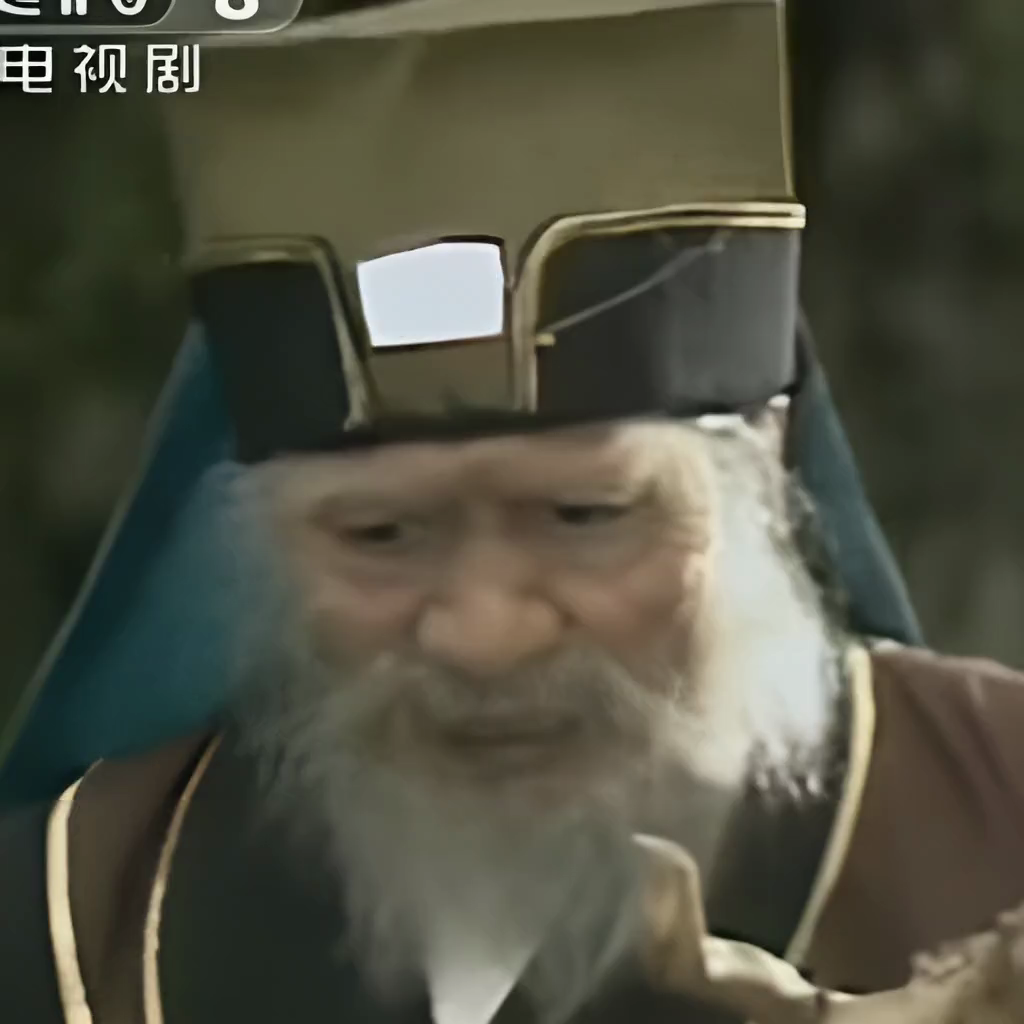} &
    \includegraphics[width=0.13\textwidth]{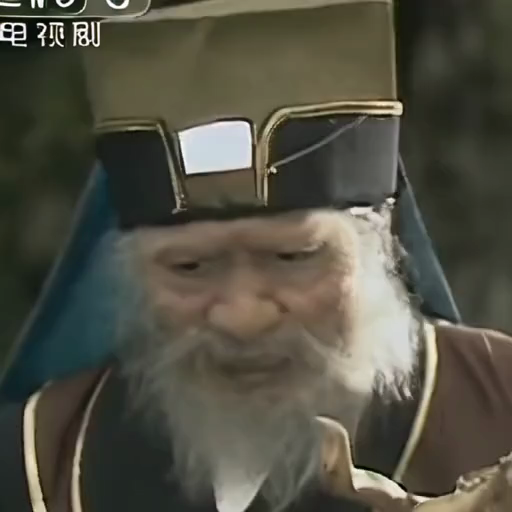} &
    \includegraphics[width=0.13\textwidth]{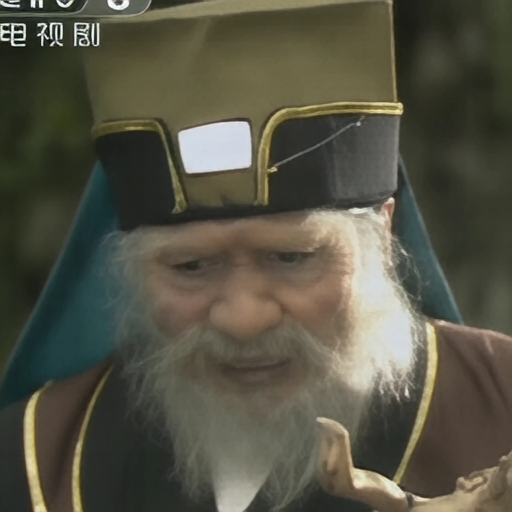} \\
    LQ (RFV-LQ) & AverNet~\cite{zhao2024avernet} & PGTFormer~\cite{xu2024beyond} &
    BFVR~\cite{wang2025efficient} & DicFace~\cite{chen2025dicface} &
    SVFR~\cite{wang2025svfr} & DVFace (ours) \\
    \end{tabular}
    \end{adjustbox}\par\vspace{0.2mm}
\begin{adjustbox}{valign=t}
    \setlength{\tabcolsep}{0.8mm}
    \footnotesize
    \begin{tabular}{ccccccc}
    \includegraphics[width=0.13\textwidth]{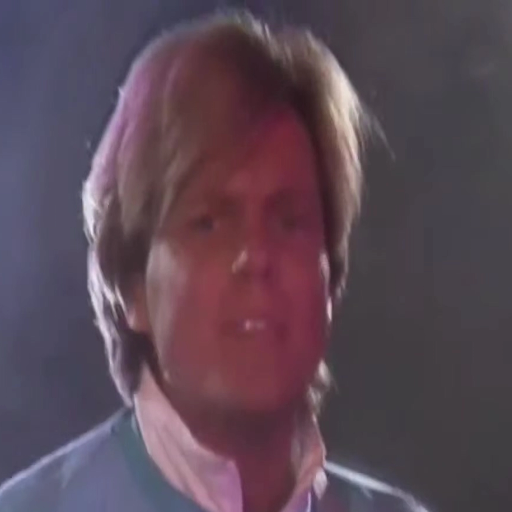} &
    \includegraphics[width=0.13\textwidth]{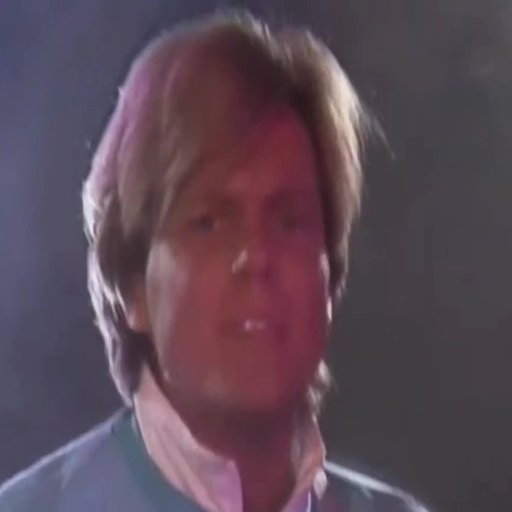} &
    \includegraphics[width=0.13\textwidth]{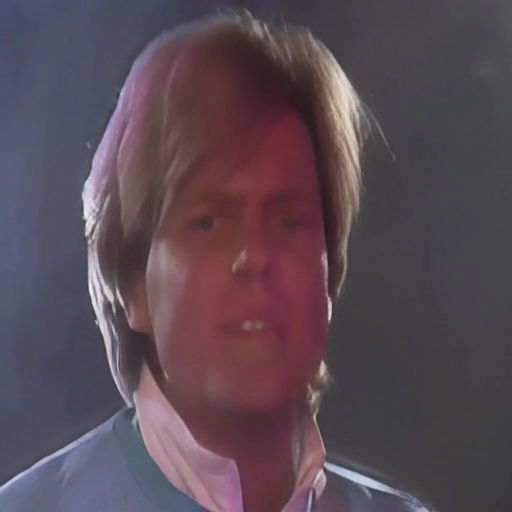} &
    \includegraphics[width=0.13\textwidth]{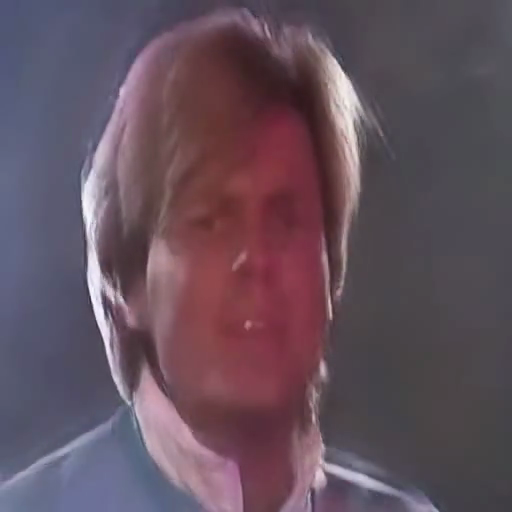} &
    \includegraphics[width=0.13\textwidth]{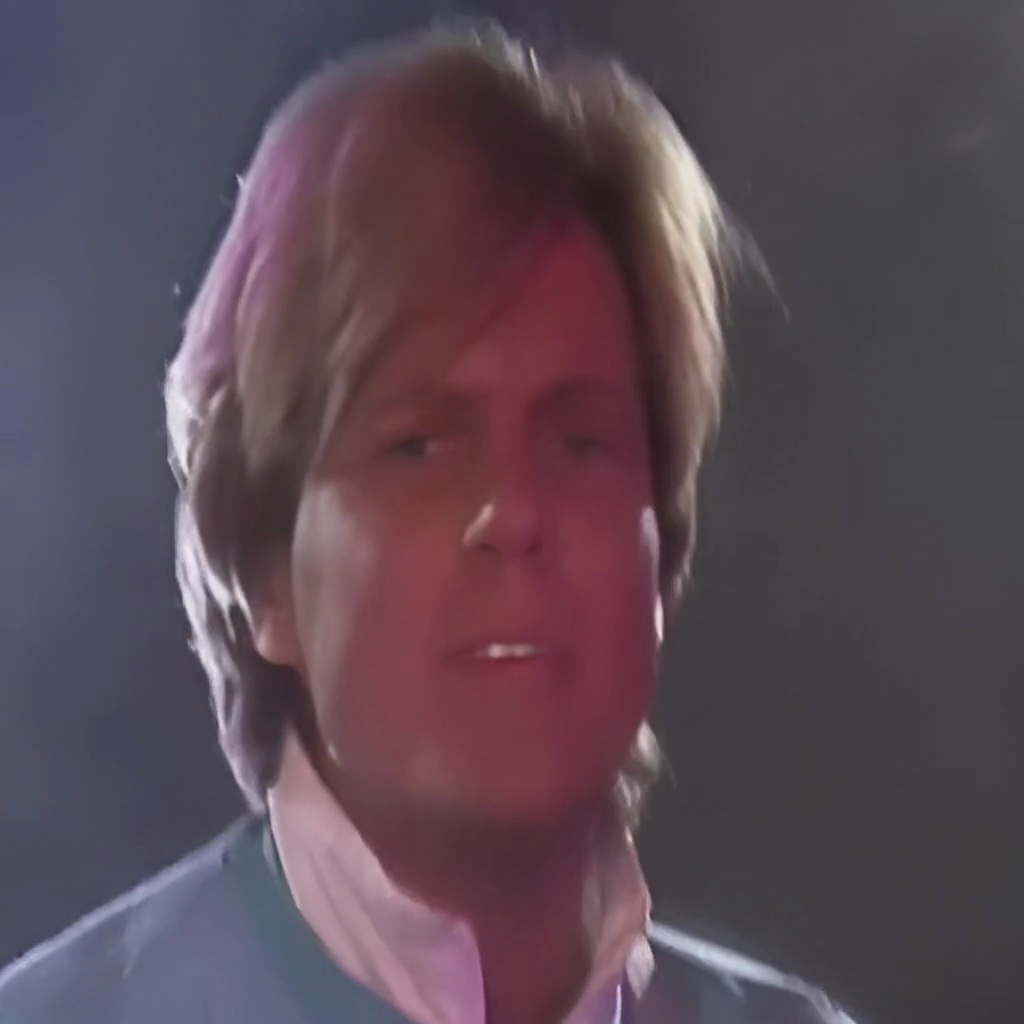} &
    \includegraphics[width=0.13\textwidth]{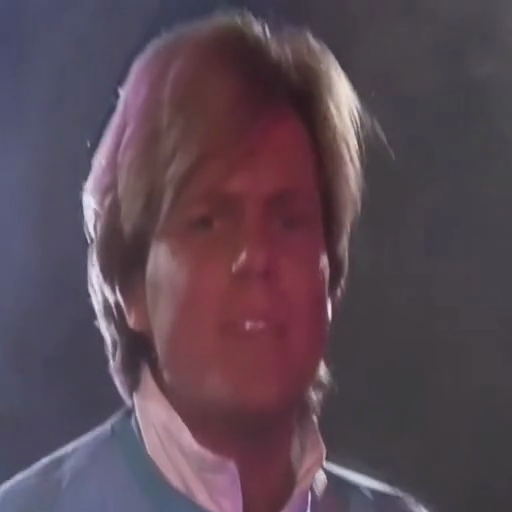} &
    \includegraphics[width=0.13\textwidth]{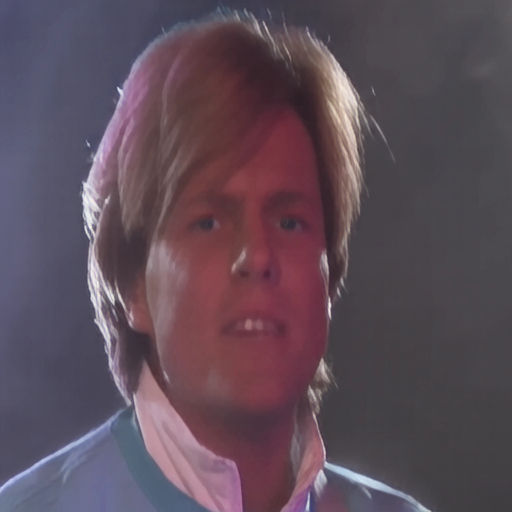} \\
    LQ (RFV-LQ) & AverNet~\cite{zhao2024avernet} & PGTFormer~\cite{xu2024beyond} &
    BFVR~\cite{wang2025efficient} & DicFace~\cite{chen2025dicface} &
    SVFR~\cite{wang2025svfr} & DVFace (ours) \\
    \end{tabular}
    \end{adjustbox}\par\vspace{0.2mm}
\begin{adjustbox}{valign=t}
    \setlength{\tabcolsep}{0.8mm}
    \footnotesize
    \begin{tabular}{ccccccc}
    \includegraphics[width=0.13\textwidth]{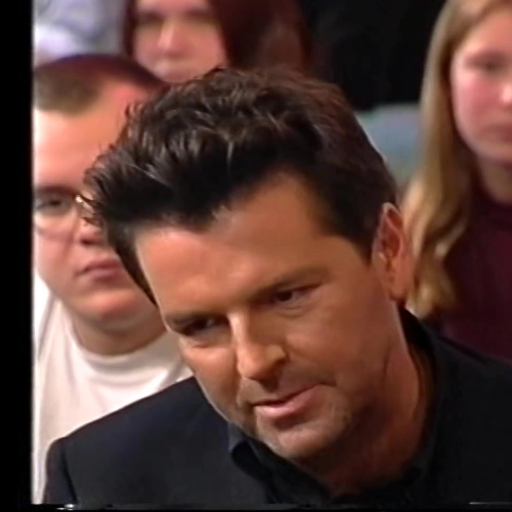} &
    \includegraphics[width=0.13\textwidth]{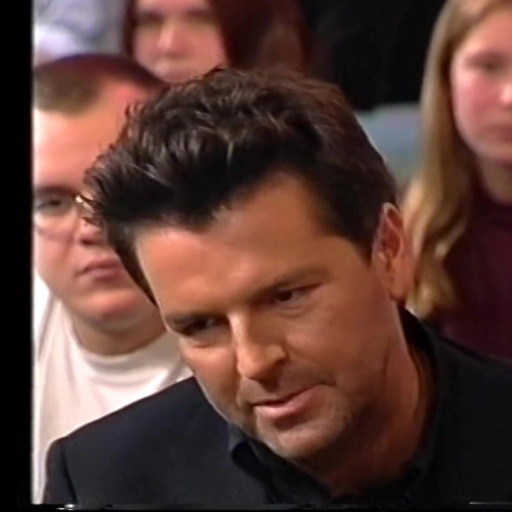} &
    \includegraphics[width=0.13\textwidth]{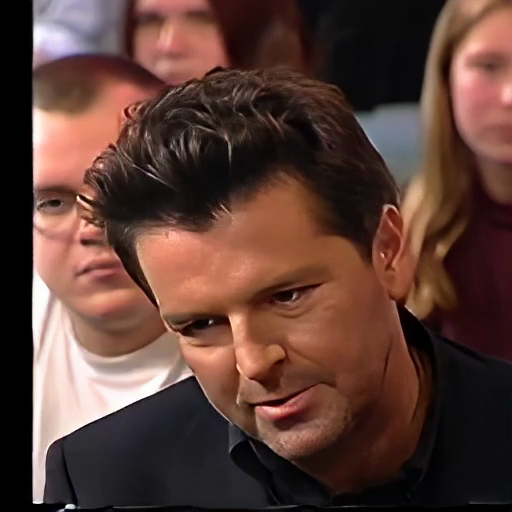} &
    \includegraphics[width=0.13\textwidth]{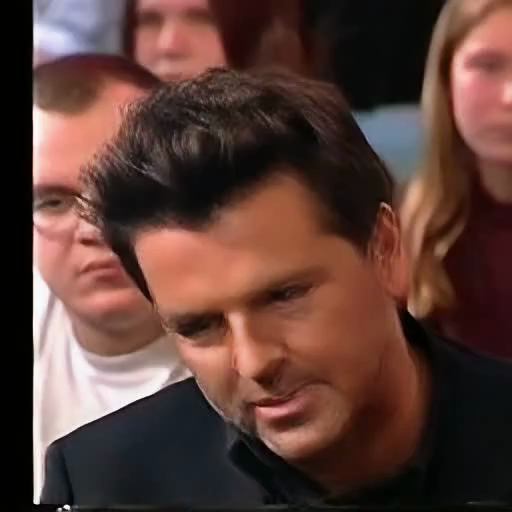} &
    \includegraphics[width=0.13\textwidth]{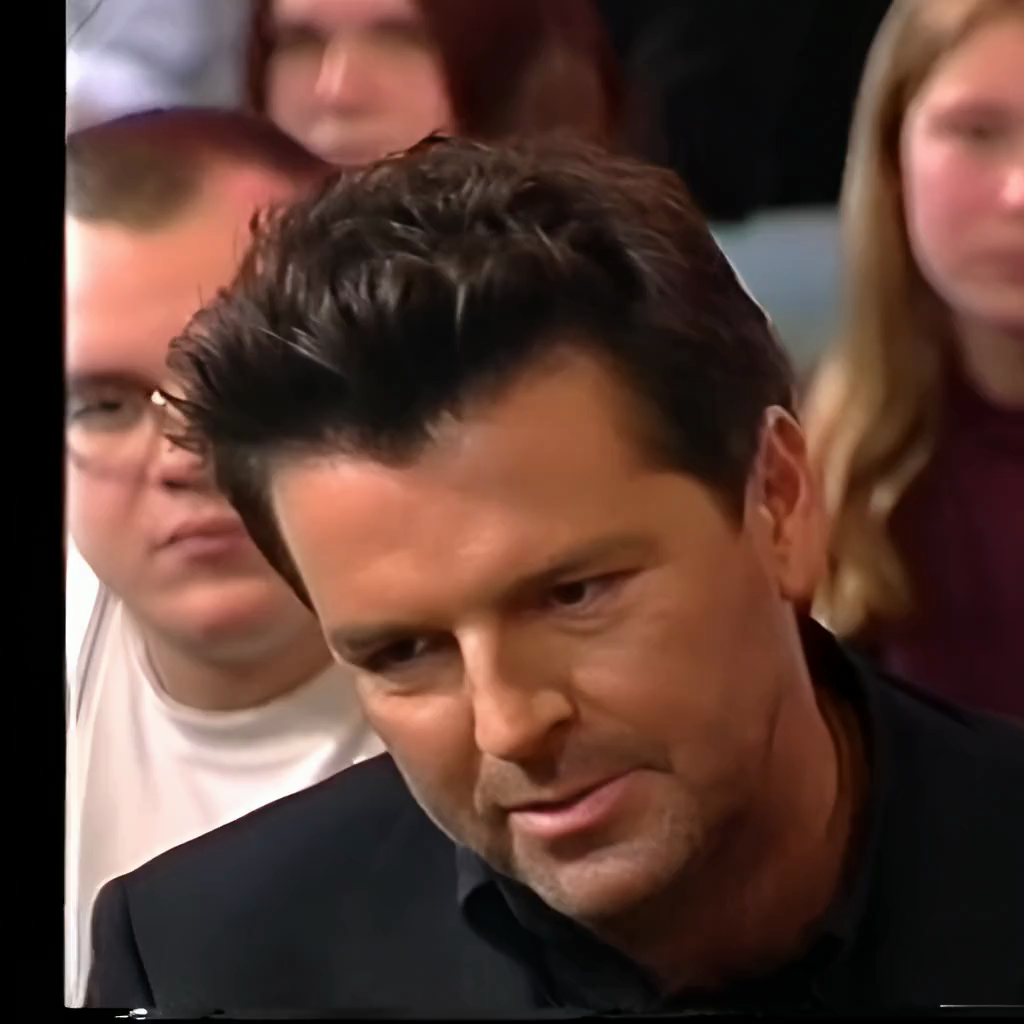} &
    \includegraphics[width=0.13\textwidth]{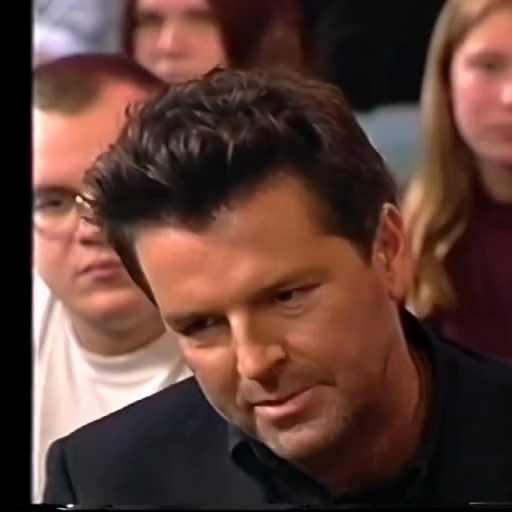} &
    \includegraphics[width=0.13\textwidth]{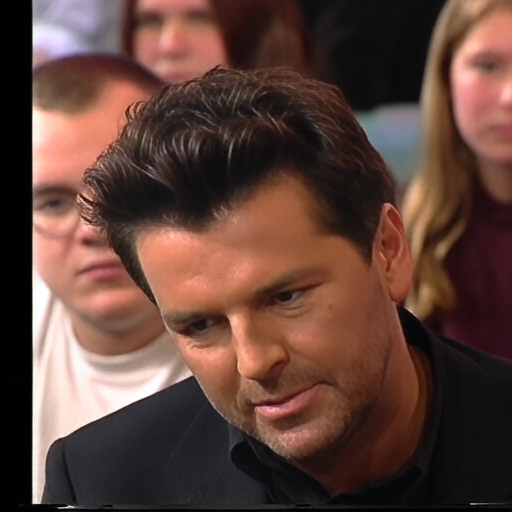} \\
    LQ (RFV-LQ) & AverNet~\cite{zhao2024avernet} & PGTFormer~\cite{xu2024beyond} &
    BFVR~\cite{wang2025efficient} & DicFace~\cite{chen2025dicface} &
    SVFR~\cite{wang2025svfr} & DVFace (ours) \\
    \end{tabular}
    \end{adjustbox}\par\vspace{0.2mm}
\vspace{-2.mm}
\caption{\added{Restoration results on real-world datasets, where HQ references are unavailable. DVFace suppresses blur and artifacts while producing natural-looking facial textures.Please zoom in for better visibility.}}
\label{fig:visual-real}
\end{figure*}

\noindent \textbf{Evaluation on Real-World Datasets.}
We further present the results on the real-world datasets RFV-LQ and Voxceleb2 in Tab.~\ref{tab:quant-real} and Fig.~\ref{fig:visual-real}. DVFace delivers leading performance on the majority of perceptual, video-quality, identity, and consistency metrics, such as CLIP-IQA, MUSIQ, LIQE, and VIDD, on both datasets. It also obtains the best MANIQA and DOVER on RFV-LQ, and the best NIQE and $E^*_{warp}$ on Voxceleb2. These results demonstrate the strong robustness of DVFace under complex real-world degradations and diverse challenging conditions. We can further observe from Fig.~\ref{fig:performance} that, in challenging examples such as the second case, other methods may suffer from blur or unstable restoration. In contrast, our method restores more realistic appearance and preserves facial characteristics more faithfully. The qualitative comparisons in Figs.~\ref{fig:visual-synthetic} and~\ref{fig:visual-real} cover synthetic examples with HQ references and real-world examples without corresponding high-quality reference frames.

\begin{table*}[t]
    \footnotesize
    \centering
    \caption{\added{Quantitative comparison with diffusion-based real-world video super-resolution (VSR) methods on VFHQ-Test. The best and second-best results are colored with \textcolor{red}{red} and blue, respectively. $\uparrow$ indicates higher is better, and $\downarrow$ indicates lower is better.}}
    \vspace{-2.mm}
    \label{tab:vsr-comparison}
    \setlength{\tabcolsep}{1.5mm}
    \begin{tabular}{l|c|c|c|c|c|c|c|c|c|c|c}
        \toprule[0.1em]
        \rowcolor{color-ours} Method & PSNR $\uparrow$ & SSIM $\uparrow$ & LPIPS $\downarrow$ & DISTS $\downarrow$ & CLIP-IQA $\uparrow$ & MUSIQ $\uparrow$ & AKD $\downarrow$ & FVD $\downarrow$ & DOVER $\uparrow$ & $E^*_{warp}$ $\downarrow$ & VIDD $\downarrow$ \\
        \midrule[0.1em]
        Upscale-A-Video~\cite{zhou2024upscale} & 27.09 & 0.8296 & 0.2578 & 0.1900 & 0.3049 & 45.22 & 0.0060 & 432.61 & 0.6211 & \textcolor{red}{0.62} & \textcolor{red}{0.3663} \\
        SeedVR~\cite{wang2025seedvr} & 26.36 & 0.8334 & 0.1359 & 0.1004 & 0.5260 & 70.07 & 0.0033 & 107.32 & 0.8568 & 1.54 & 0.5228 \\
        SeedVR2~\cite{wang2025seedvr2} & 26.11 & 0.8410 & \textcolor{blue}{0.1162} & 0.1049 & 0.5435 & 71.70 & 0.0036 & 154.44 & 0.8497 & 1.91 & 0.5426 \\
        DOVE~\cite{chen2025dove} & \textcolor{blue}{31.49} & \textcolor{red}{0.9021} & 0.0946 & \textcolor{blue}{0.0871} & \textcolor{blue}{0.5707} & \textcolor{blue}{71.81} & \textcolor{blue}{0.0027} & \textcolor{red}{49.58} & \textcolor{red}{0.8806} & 1.13 & 0.4784 \\
        DVFace (ours) & \textcolor{red}{31.81} & \textcolor{blue}{0.9007} & \textcolor{red}{0.0776} & \textcolor{red}{0.0805} & \textcolor{red}{0.6289} & \textcolor{red}{71.89} & \textcolor{red}{0.0026} & \textcolor{blue}{60.11} & \textcolor{blue}{0.8703} & \textcolor{blue}{0.91} & \textcolor{blue}{0.4339} \\
        \bottomrule[0.1em]
    \end{tabular}
\end{table*}

\begin{table*}[t]
    \footnotesize
    \centering
    \caption{\added{Quantitative comparison with image face restoration (IFR) methods on VFHQ-Test. OSDFace and AuthFace are diffusion-based IFR methods, which restore videos by processing each frame independently. The best and second-best results are colored with \textcolor{red}{red} and blue, respectively.}}
    \vspace{-2.mm}
    \label{tab:diffusion-face-comparison}
    \setlength{\tabcolsep}{7.2mm}
    \begin{tabular}{l|c|c|c|c|c|c}
        \toprule[0.1em]
        \rowcolor{color-ours} Method & Task & Steps & PSNR $\uparrow$ & LPIPS $\downarrow$ & $E^*_{warp}$ $\downarrow$ & FVD $\downarrow$ \\
        \midrule[0.1em]
        OSDFace~\cite{wang2025osdface} & Image-BFR & 1 & 24.66 & 0.1950 & 3.66 & 233.76 \\
        AuthFace~\cite{liang2025authface} & Image-BFR & 50 & \textcolor{blue}{26.99} & \textcolor{blue}{0.1782} & \textcolor{blue}{2.69} & \textcolor{blue}{152.35} \\
        DVFace (ours) & Video-BFR & 1 & \textcolor{red}{31.81} & \textcolor{red}{0.0776} & \textcolor{red}{0.91} & \textcolor{red}{60.11} \\
        \bottomrule[0.1em]
    \end{tabular}
\end{table*}

\begin{figure*}[t]
    \centering
    \includegraphics[width=\linewidth]{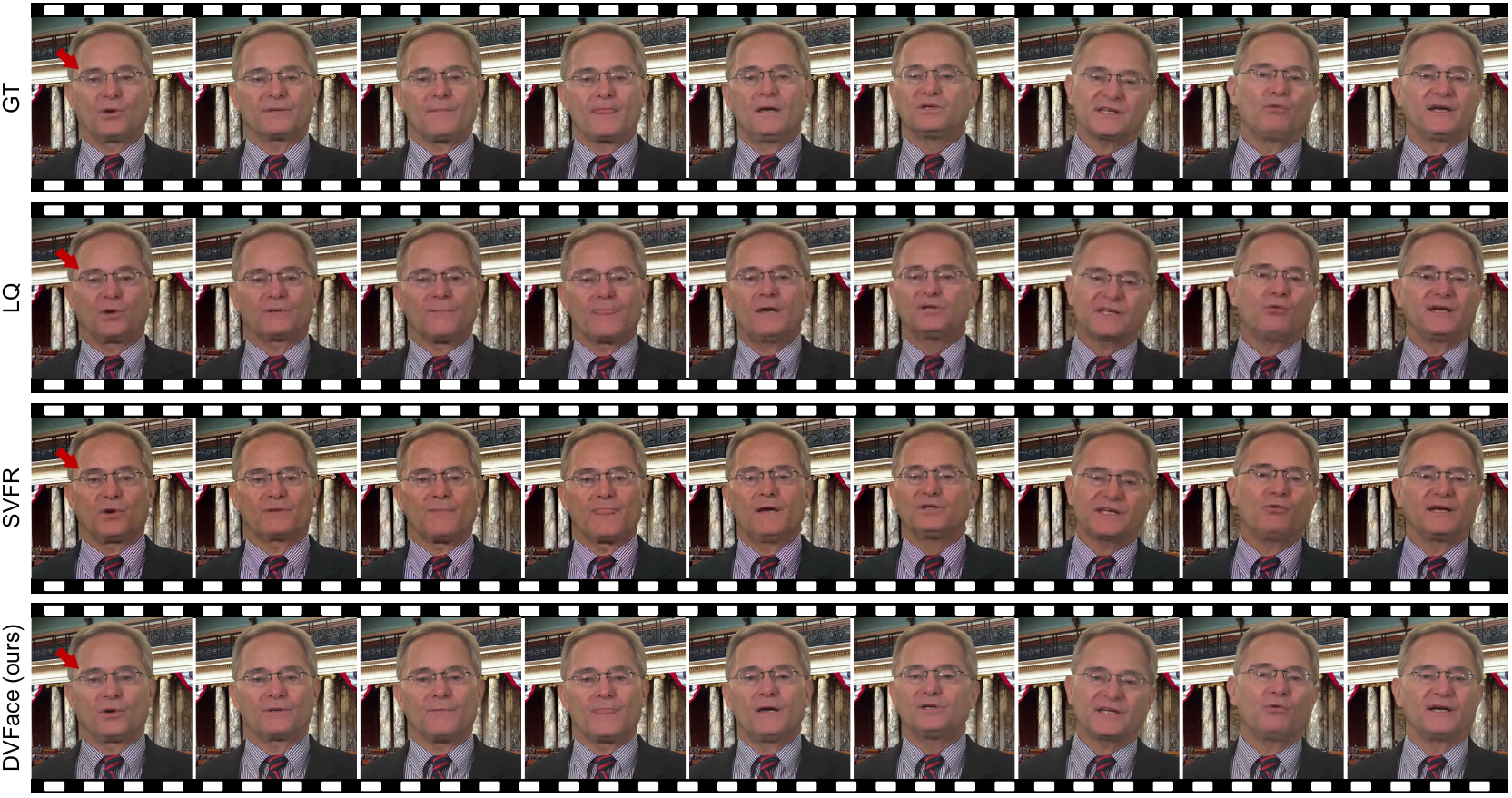}
    \vspace{-2.mm}
    \caption{\added{Multi-frame comparison on consecutive frames. DVFace preserves stable facial appearance and consistent local details, whereas competing methods exhibit texture variations across time.}}
    \label{fig:multi}
\end{figure*}

\begin{figure}[t]
    \centering
    \includegraphics[width=\columnwidth]{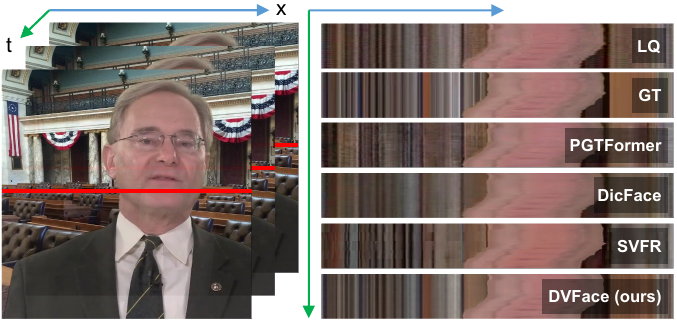}
    \vspace{-2.mm}
    \caption{Temporal consistency comparison across consecutive frames (red line stacked across frames).}
    \label{fig:temporal}
\end{figure}

\added{\noindent \textbf{Temporal Consistency.}}
\added{We analyze temporal consistency using two complementary visualizations. In the line-stack comparison of Fig.~\ref{fig:temporal}, pixels along the marked red line are stacked over time. Smooth and continuous patterns indicate stable motion, whereas distortions reveal flickering or structural changes. DVFace produces the smoothest transition and maintains coherent facial structures. The independent multi-frame comparison in Fig.~\ref{fig:multi} further shows that DVFace preserves facial appearance and local textures consistently across frames, while competing methods introduce visible detail variations. These observations agree with the quantitative results in Tabs.~\ref{tab:quant-synthetic} and~\ref{tab:quant-real}, where DVFace achieves the best or competitive $E^*_{warp}$ and VIDD across all benchmarks.}

\subsection{\added{Comparison with Related Diffusion-Based Restoration Methods}}
\vspace{1.5mm}
\added{DVFace is built upon a pretrained diffusion model. We further compare it with related diffusion-based restoration methods, including those for video super-resolution (VSR) and image face restoration (IFR).}

\added{\noindent \textbf{Comparison with Diffusion-Based VSR Methods.}
Video face restoration is a specialized video restoration task, and general-purpose VSR models can also be applied to face videos. We compare DVFace with representative diffusion-based VSR methods, including Upscale-A-Video~\cite{zhou2024upscale}, SeedVR~\cite{wang2025seedvr}, SeedVR2~\cite{wang2025seedvr2}, and DOVE~\cite{chen2025dove}. As shown in Tab.~\ref{tab:vsr-comparison}, DVFace achieves the best PSNR, LPIPS, DISTS, CLIP-IQA, MUSIQ, and AKD on VFHQ-Test. Compared with DOVE, it improves PSNR from 31.49 to 31.81 dB and reduces LPIPS from 0.0946 to 0.0776, while also obtaining lower $E^*_{warp}$ and VIDD. Although DOVE achieves better SSIM, FVD, and DOVER, DVFace offers a favorable balance between facial reconstruction quality and temporal consistency. Upscale-A-Video achieves the best temporal consistency, but it suffers from significantly degraded visual quality, as reflected by its much lower PSNR and poor perceptual metrics. Overall, these results support the benefit of guiding the restoration backbone with face-specific spatial and temporal priors extracted by our dual codebooks.}

\added{\noindent \textbf{Comparison with Diffusion-Based IFR Methods.}
IFR models can restore face videos by processing each frame independently. We compare DVFace with representative state-of-the-art diffusion-based IFR methods, OSDFace~\cite{wang2025osdface} and AuthFace~\cite{liang2025authface}, and include the video-based method SVFR~\cite{wang2025svfr} as a reference. As reported in Tab.~\ref{tab:diffusion-face-comparison}, DVFace achieves the best results on all four metrics with only one denoising step. Compared with the one-step IFR method OSDFace, DVFace improves PSNR from 24.66 to 31.81 dB and reduces LPIPS from 0.1950 to 0.0776, while substantially lowering $E^*_{warp}$ and FVD. It also outperforms AuthFace on all four metrics despite requiring only one denoising step rather than 50. These results highlight the limitations of independent frame-wise restoration and support the advantage of jointly exploiting facial appearance priors and temporal information for video face restoration.}

\subsection{\added{Efficiency Comparison}}
\added{We compare the inference efficiency of DVFace with representative diffusion-based video restoration and blind face restoration methods. All measurements are obtained on a single NVIDIA A6000 GPU using an 81-frame video at $512\times512$ resolution under the same evaluation setting.}

\begin{table*}[t]
    \footnotesize
    \centering
    \caption{\added{Efficiency comparison with representative restoration methods. Runtime is measured on NVIDIA A6000 GPU using an 81-frame $512\times512$ video.}}
    \vspace{-2.mm}
    \label{tab:efficiency-comparison}
    \setlength{\tabcolsep}{1.1mm}
    \begin{tabular}{l|c|c|c|c|c|c|c|c}
        \toprule[0.1em]
        \rowcolor{color-ours} Metric
        & Upscale-A-Video~\cite{zhou2024upscale}
        & SeedVR~\cite{wang2025seedvr}
        & SeedVR2~\cite{wang2025seedvr2}
        & DOVE~\cite{chen2025dove}
        & DicFace~\cite{chen2025dicface}
        & SVFR~\cite{wang2025svfr}
        & AuthFace~\cite{liang2025authface}
        & DVFace (ours) \\
        \midrule[0.1em]
        Inference Steps & 30 & 50 & 1 & 1 & 1 & 30 & 50 & 1 \\
        Parameters (M)  & 1092.01 & 3642.12 & 3642.12 & 5787.19
                        & \textbf{164.78} & 2254.45 & 4719.85 & 1604.16 \\
        FLOPs (T)       & 6070.53 & 6387.90 & 403.56 & 942.88
                        & 333.80 & 4156.17 & 9523.17 & \textbf{230.21} \\
        Runtime (s)     & 349.73 & 131.13 & 23.74 & 16.85
                        & 33.21 & 229.81 & 307.32 & \textbf{15.13} \\
        \bottomrule[0.1em]
    \end{tabular}
\end{table*}

\added{\noindent \textbf{Overall Efficiency.}}
\added{As shown in Tab.~\ref{tab:efficiency-comparison}, multi-step diffusion methods require 30--50 denoising steps and therefore incur substantially higher computation and latency. DVFace completes restoration in one step with 230.21 T FLOPs and 15.13 s runtime, achieving approximately $23.1\times$ and $15.2\times$ speedups over Upscale-A-Video and SVFR, respectively. Compared with one-step video restoration models, DVFace uses about $3.6\times$ fewer parameters than DOVE and is also faster than both DOVE and SeedVR2. Although DicFace has fewer parameters, DVFace requires fewer FLOPs and runs more than twice as fast under the same setting. These results demonstrate that the one-step formulation offers an effective efficiency--quality trade-off for video face restoration.}

\begin{table}[t]
    \centering
    \footnotesize
    \caption{\added{Module-wise complexity of DVFace on an 81-frame $512\times512$ input. Parentheses report the overhead relative to the baseline.}}
    \vspace{-2.mm}
    \label{tab:module-overhead}
    \setlength{\tabcolsep}{4.8mm}
    \begin{tabular}{l|c|c}
        \toprule[0.1em]
        \rowcolor{color-ours} Variant & Parameters (M) & FLOPs (T) \\
        \midrule[0.1em]
        Baseline & 1545.89 & 214.43 \\
        +Codebook & +31.04 (2.01\%) & +8.23 (3.84\%) \\
        +Temporal & +7.48 (0.48\%) & +0.011 (0.005\%) \\
        +Spatial & +19.75 (1.28\%) & +7.53 (3.51\%) \\
        DVFace & 1604.16 (+3.77\%) & 230.21 (+7.36\%) \\
        \bottomrule[0.1em]
    \end{tabular}
\end{table}

\added{\noindent \textbf{Component Complexity.}}
\added{Tab.~\ref{tab:module-overhead} decomposes the cost of the proposed components. The dual codebooks add 31.04 M parameters and 8.23 T FLOPs, while the temporal modulation branch introduces 7.48 M parameters and 0.011 T FLOPs. The spatial fusion branch adds 19.75 M parameters and 7.53 T FLOPs. Together, STDC and ASTF increase the baseline parameters and FLOPs by 3.77\% and 7.36\%, respectively, under the same evaluation setting.}

\subsection{Ablation Study}
\begingroup
\arrayrulecolor{black}

We conduct ablation studies to evaluate the effectiveness of the key designs in our methods. The studies consist of three parts. The first two analyze our two core contributions: the spatio-temporal dual-codebook (STDC) design and the asymmetric spatio-temporal fusion (ASTF) module. The third focuses on temporal consistency, a critical property for video restoration. We report PSNR and LPIPS for fidelity and perceptual quality, $E^*_{warp}$ for temporal consistency, and FVD for overall video quality. Unless otherwise specified, all experiments are conducted on VFHQ-Test.

\begin{table*}[t]
    \centering
    \footnotesize
    \arrayrulecolor{black}
    \caption{Ablations of the spatio-temporal dual-codebook (STDC) priors on VFHQ-Test. The first block compares different conditioning priors, and the second evaluates the composition of spatial and temporal priors.}
    \vspace{-2mm}
    \setlength{\tabcolsep}{6.2mm}
    \begin{tabular}{ll|cccc}
        \toprule[0.1em]
        \rowcolor{color-ours} Ablation Target & Configuration & PSNR $\uparrow$ & LPIPS $\downarrow$ & $E^*_{warp} \downarrow$ & FVD $\downarrow$ \\
        \midrule[0.1em]
        \multirow{4}{*}{Conditioning representation}
        & Null & 31.50 & 0.0826 & 0.97 & 69.83 \\
        & Fixed text & 31.52 & 0.0813 & 0.96 & 62.48 \\
        & DAPE~\cite{wu2023seesr} & 31.51 & 0.0796 & 0.92 & 63.61 \\
        & STDC priors (ours) & \textbf{31.81} & \textbf{0.0776} & \textbf{0.91} & \textbf{60.11} \\
        \midrule[0.1em]
        \multirow{4}{*}{Prior composition}
        & No prior & 31.50 & 0.0826 & 0.97 & 69.83 \\
        & Spatial only ($\boldsymbol{f}_s$) & 31.68 & 0.0791 & 0.92 & 60.45 \\
        & Temporal only ($\boldsymbol{f}_t$) & 31.71 & 0.0789 & \textbf{0.90} & 60.85 \\
        & Spatial + temporal (ours) & \textbf{31.81} & \textbf{0.0776} & 0.91 & \textbf{60.11} \\
        \bottomrule[0.1em]
    \end{tabular}
    \label{table:ablation_prior}
    \vspace{-2.mm}
\end{table*}

\begin{figure}[t]
    \centering
    \includegraphics[width=\linewidth]{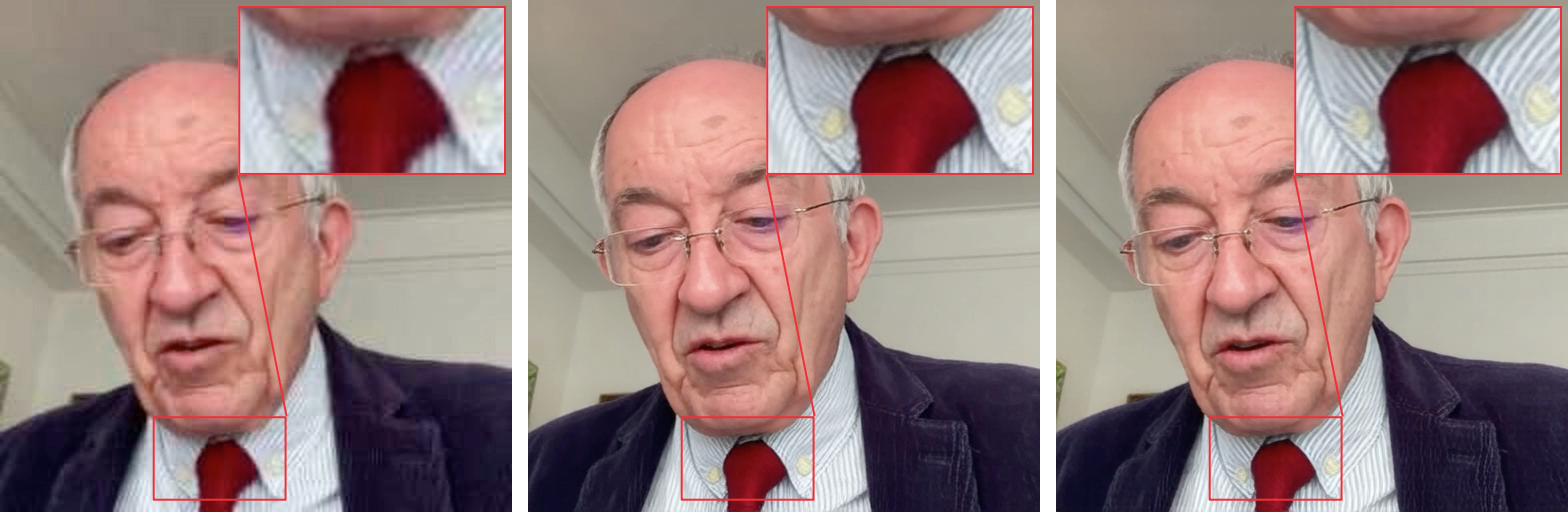}
    \vspace{2pt}
    {\footnotesize
    \makebox[\linewidth]{\makebox[0.333\linewidth]{Input}\hfill
        \makebox[0.333\linewidth]{No prior}\hfill
        \makebox[0.333\linewidth]{Ours}\hfill
        }}
    \vspace{-6mm}
    \caption{Visual comparison with and without spatio-temporal priors. DVFace restores more realistic and natural-looking textures.}
    \vspace{-5mm}
    \label{fig:ablation}
\end{figure}

\vspace{1.mm}
\noindent \textbf{Spatio-Temporal Dual-Codebook Design.}
We evaluate STDC from two complementary perspectives in Tab.~\ref{table:ablation_prior}: whether a structured codebook representation is preferable to text conditioning and whether its spatial and temporal components provide complementary information.

\vspace{1.mm}
\textbf{\textit{(1) Conditioning representation.}}
This experiment determines whether the gains arise merely from adding a condition or specifically from the proposed structured video-face priors. \textit{Null} uses an empty text prompt without an additional prior; \textit{Fixed text} applies the same quality-oriented prompt to every video; and \textit{DAPE} extracts descriptive tags from the first LQ frame and reuses them for the entire sequence. Our STDC instead extracts $\boldsymbol{f}_s$ and $\boldsymbol{f}_t$ from the LQ video; inputs of varying length are temporally interpolated to 24 frames before codebook extraction. Text conditioning improves some perceptual and video-level metrics but leaves PSNR almost unchanged. In contrast, STDC achieves the best result on all four metrics, improving PSNR by 0.31 dB and reducing LPIPS, $E^*_{warp}$, and FVD by 0.0050, 0.06, and 9.72, respectively, over Null. This consistent gain indicates that discrete facial appearance and dynamics provide more task-relevant guidance than coarse semantic descriptions.

\vspace{1.mm}
\textbf{\textit{(2) Prior composition.}}
We isolate the two codebooks by removing both priors, retaining only $\boldsymbol{f}_s$, retaining only $\boldsymbol{f}_t$, or using both. The spatial-only variant improves PSNR/LPIPS from 31.50/0.0826 to 31.68/0.0791 and reduces FVD from 69.83 to 60.45, demonstrating the benefit of spatial appearance priors. The temporal-only variant attains the lowest $E^*_{warp}$ of 0.90, highlighting the value of inter-frame dynamics. Combining both priors yields the best PSNR, LPIPS, and FVD while maintaining competitive temporal consistency (0.91 versus 0.90). Fig.~\ref{fig:ablation} further shows that using both priors produces more realistic and natural-looking textures than the variant without priors.

\begin{table*}[t]
    \centering
    \footnotesize
    \arrayrulecolor{black}
    \caption{Ablations of the asymmetric spatio-temporal fusion (ASTF) module on VFHQ-Test. Each block varies one design choice with the remaining components fixed. Direct $\boldsymbol{f}_s$ injection omits temporal pre-refinement; layer-specific modulation predicts independent $(\gamma,\beta)$ for each DiT block.}
    \vspace{-2mm}
    \setlength{\tabcolsep}{6.7mm}
    \begin{tabular}{ll|cccc}
        \toprule[0.1em]
        \rowcolor{color-ours} Ablation Target & Configuration & PSNR $\uparrow$ & LPIPS $\downarrow$ & $E^*_{warp} \downarrow$ & FVD $\downarrow$ \\
        \midrule[0.1em]
        \multirow{2}{*}{Overall fusion}
        & Without ASTF & 31.50 & 0.0826 & 0.97 & 69.83 \\
        & With ASTF (ours) & \textbf{31.81} & \textbf{0.0776} & \textbf{0.91} & \textbf{60.11} \\
        \midrule[0.1em]
        \multirow{2}{*}{Temporal modulation}
        & Layer-specific $(\gamma,\beta)$ & 31.69 & 0.0786 & 0.94 & \textbf{59.20} \\
        & Layer-shared $(\gamma,\beta)$ (ours) & \textbf{31.81} & \textbf{0.0776} & \textbf{0.91} & 60.11 \\
        \midrule[0.1em]
        \multirow{3}{*}{Spatial refinement}
        & Direct $\boldsymbol{f}_s$ injection & 31.45 & 0.0784 & \textbf{0.91} & 63.41 \\
        & $\boldsymbol{f}_s$ as Q; $\boldsymbol{f}_t$ as K/V & 31.61 & 0.0795 & 0.93 & 62.90 \\
        & $\boldsymbol{f}_t$ as Q; $\boldsymbol{f}_s$ as K/V (ours) & \textbf{31.81} & \textbf{0.0776} & \textbf{0.91} & \textbf{60.11} \\
        \midrule[0.1em]
        \multirow{3}{*}{Fusion placement}
        & First half of DiT blocks & 31.69 & 0.0801 & 0.93 & 62.48 \\
        & Second half of DiT blocks & 31.72 & 0.0795 & 0.93 & 62.13 \\
        & All DiT blocks (ours) & \textbf{31.81} & \textbf{0.0776} & \textbf{0.91} & \textbf{60.11} \\
        \bottomrule[0.1em]
    \end{tabular}
    \label{table:ablation_fusion}
    \vspace{-2.mm}
\end{table*}

\noindent \textbf{Asymmetric Spatio-Temporal Fusion Module.}
We study the necessity, branch-specific design, and placement of ASTF in Tab.~\ref{table:ablation_fusion}. Together, these experiments \added{directly} examine both the overall value of explicit prior integration and the rationale behind its asymmetric implementation.

\vspace{1.mm}
\textbf{\textit{(1) Overall fusion.}}
We first remove all ASTF modules, preventing the codebook priors from conditioning the diffusion backbone, and compare this variant with the complete model. Introducing ASTF increases PSNR from 31.50 to 31.81 dB and decreases LPIPS, $E^*_{warp}$, and FVD from 0.0826/0.97/69.83 to 0.0776/0.91/60.11. The consistent improvements \added{across all four metrics} show that explicitly integrating the extracted priors is essential for translating STDC representations into restoration gains.

\vspace{1.mm}
\textbf{\textit{(2) Temporal modulation.}}
In the temporal branch, we share the same scale-and-shift parameters across all DiT blocks instead of predicting layer-specific parameters. This design reduces parameter and computational overhead. It also improves restoration performance, as shown in Tab.~\ref{table:ablation_fusion}. Compared with layer-specific modulation, sharing improves PSNR, LPIPS, and $E^*_{warp}$ to 31.81, 0.0776, and 0.91, respectively, although layer-specific modulation obtains a slightly lower FVD (59.20 versus 60.11). Overall, the advantage on three complementary metrics supports our design.

\vspace{1.mm}
\textbf{\textit{(3) Spatial refinement.}}
We assess both the need for temporal guidance and the direction of cross-attention in spatial branch. The compared variants directly inject $\boldsymbol{f}_s$ without temporal pre-refinement, use $\boldsymbol{f}_s$ as query and $\boldsymbol{f}_t$ as key/value, or adopt our reversed assignment. Direct injection degrades PSNR and FVD to 31.45 and 63.41, indicating that unfiltered spatial codes may introduce unreliable local patterns. Moreover, using $\boldsymbol{f}_t$ as query and $\boldsymbol{f}_s$ as key/value outperforms the reversed direction on every metric. Since attention aggregates information from the value tokens ($\boldsymbol{f}_s$), using $\boldsymbol{f}_t$ as query allows temporal information to guide the injection of spatial details. This confirms that temporally guided spatial information injection is effective.

\vspace{1.mm}
\textbf{\textit{(4) Fusion placement.}}
Finally, we insert ASTF into the first half, the second half, or all DiT blocks to determine how extensively the priors should condition the backbone. Both half-depth variants improve \added{all reported metrics} substantially over the model without ASTF, and the second-half placement is marginally stronger than the first. Applying ASTF to all blocks further achieves the best result on every metric, showing that prior conditioning is beneficial throughout the denoising hierarchy.

\begin{table}[t]
    \centering
    \small
    \arrayrulecolor{black}
    \caption{\added{Temporal-consistency ablations. (a) Sensitivity to $\lambda_{temp}$ on VFHQ-Test. (b) Real-world generalization of temporal consistency on real-world datasets RFV-LQ and VoxCeleb2, measured by $E^*_{warp}\downarrow$.}}
    \vspace{-2mm}
    \label{table:ablation_other}
    \begin{minipage}{\columnwidth}
        \centering
        \scriptsize
        \textbf{(a) Temporal-loss weight sensitivity}\par\vspace{1mm}
        \begin{tabularx}{\linewidth}{l|>{\centering\arraybackslash}X>{\centering\arraybackslash}X>{\centering\arraybackslash}X>{\centering\arraybackslash}X}
            \toprule[0.1em]
            \rowcolor{color-ours} $\lambda_{temp}$ & PSNR $\uparrow$ & LPIPS $\downarrow$ & $E^*_{warp} \downarrow$ & FVD $\downarrow$ \\
            \midrule[0.1em]
            0 & 31.50 & 0.0797 & 1.10 & 60.19 \\
            0.1 (ours) & \textbf{31.81} & \textbf{0.0776} & 0.91 & \textbf{60.11} \\
            0.2 & 31.66 & 0.0788 & \textbf{0.83} & 64.69 \\
            \bottomrule[0.1em]
        \end{tabularx}
    \end{minipage}
    \par\vspace{2.5mm}
    \begin{minipage}{\columnwidth}
        \centering
        \scriptsize
        \textbf{(b) Real-World Generalization of Temporal Consistency}\par\vspace{1mm}
        \begin{tabularx}{\linewidth}{l|>{\centering\arraybackslash}X>{\centering\arraybackslash}X|>{\centering\arraybackslash}X>{\centering\arraybackslash}X}
            \toprule[0.1em]
            \rowcolor{color-ours} Dataset & STDC w/o & STDC w/ & $\mathcal{L}_{temp}$ w/o & $\mathcal{L}_{temp}$ w/ \\
            \midrule[0.1em]
            RFV-LQ & 1.00 & \textbf{0.89} & 1.07 & \textbf{0.89} \\
            VoxCeleb2 & 1.65 & \textbf{1.45} & 1.84 & \textbf{1.45} \\
            \bottomrule[0.1em]
        \end{tabularx}
    \end{minipage}
    \vspace{-1.mm}
\end{table}

\vspace{1.mm}
\noindent \textbf{Temporal Consistency Analysis.}
Both the codebook priors and the temporal loss $\mathcal{L}_{temp}$ improve temporal consistency in our method. We further study how the temporal-loss weight affects temporal consistency and whether the temporal gains obtained on synthetic training data generalize to real-world videos. The results are summarized in Tab.~\ref{table:ablation_other}.

\vspace{1.mm}
\textbf{\textit{(1) Temporal-loss weight.}}
We vary the loss weight $\lambda_{temp}$ over $\{0,0.1,0.2\}$. As shown in Tab.~\ref{table:ablation_other}(a), removing temporal supervision worsens temporal consistency, raising the warping error $E^*_{warp}$ from 0.91 to 1.10, while lowering PSNR and worsening LPIPS. Increasing the weight to 0.2 yields the lowest $E^*_{warp}$ of 0.83 but lowers PSNR, worsens LPIPS, and raises FVD to 64.69, indicating that excessive temporal regularization favors smoothness at the expense of frame-level fidelity and video realism. We therefore set $\lambda_{temp}=0.1$, which provides the best overall balance.

\vspace{1.mm}
\textbf{\textit{(2) Real-World Generalization of Temporal Consistency.}}
We independently remove STDC and $\mathcal{L}_{temp}$ and evaluate the resulting variants on the real-world datasets RFV-LQ and VoxCeleb2. As shown in Tab.~\ref{table:ablation_other}(b), removing STDC worsens temporal consistency, raising the warping error $E^*_{warp}$ from 0.89 to 1.00 on RFV-LQ and from 1.45 to 1.65 on VoxCeleb2. Removing $\mathcal{L}_{temp}$ similarly raises the warping error. Overall, both STDC and $\mathcal{L}_{temp}$ improve temporal consistency, and their benefits generalize to real-world videos.

\endgroup
\FloatBarrier
\vspace{-2.mm}
\section{\added{Discussion and Future Work}}
\added{The results indicate that explicit spatial and temporal priors provide effective guidance for adapting a pretrained one-step diffusion model to video face restoration. The dual codebooks separate stable facial appearance from inter-frame dynamics, while the asymmetric fusion injects them according to their distinct roles. This design allows DVFace to preserve fine facial details and temporal coherence while maintaining efficient inference.}

\added{Two limitations remain. Under extreme degradation, low-quality features may deviate from the high-quality facial manifold, making codebook retrieval ambiguous and potentially yielding unreliable priors. In addition, long videos require chunk-wise processing because of GPU memory constraints, which may weaken consistency across chunk boundaries. Future work will explore confidence-aware or degradation-adaptive prior selection, as well as cross-chunk temporal memory and streaming inference to improve robustness and long-video consistency.}

\vspace{-2.mm}
\section{Conclusion}
In this paper, we present DVFace, a one-step diffusion framework for real-world video face restoration. DVFace is developed from a pretrained diffusion backbone. It explicitly exploits facial prior information from degraded input videos while enabling efficient and effective one-step inference. To achieve this goal, we design a spatio-temporal dual-codebook scheme to extract spatial and temporal priors. We further develop an asymmetric spatio-temporal fusion module to incorporate these priors according to their distinct roles. With these designs, DVFace restores faithful facial details while preserving temporal coherence and identity consistency. Extensive experiments on both synthetic and real-world datasets verify the superior overall performance of DVFace \added{across all evaluated benchmarks}.

\section*{Acknowledgments}
This work is supported by the National Natural Science Foundation of China (625B2116 and 62501386).

\bibliographystyle{IEEEtran}
\bibliography{bib_conf}

\begin{IEEEbiography}[{\includegraphics[width=1in,height=1.25in, clip,keepaspectratio]{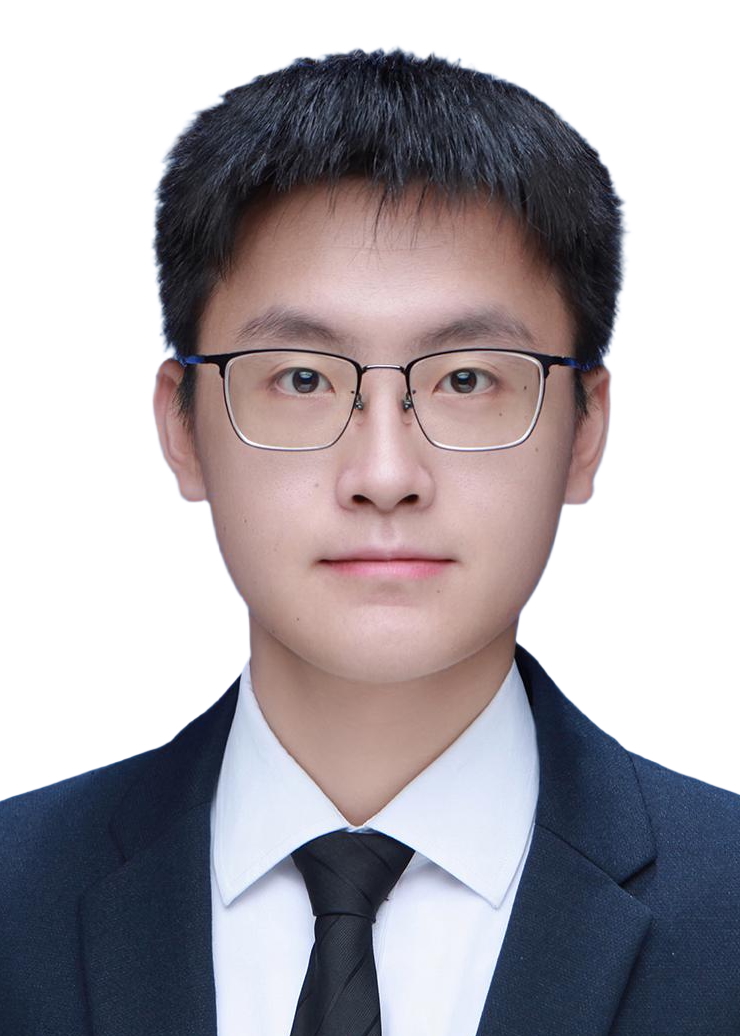}}]{Zheng Chen} is a Ph.D. student in Computer Science and Engineering at Shanghai Jiao Tong University, advised by Prof. Yulun Zhang. He received his B.S. degree from Shanghai Jiao Tong University in 2022. His research focuses on image/video restoration, model compression, and multimodal models. He has published papers in international conferences, including NeurIPS, ICCV, CVPR, ICLR, AAAI, and ICML. He has received the National Scholarship for Graduate Excellence and a grant from the NSFC Young Students Basic Research Program for Ph.D. students. He was also selected for the Young Science and Technology Talent Cultivation Program (Doctoral Student Special Project) of the China Association for Science and Technology.
\end{IEEEbiography}

\begin{IEEEbiography}[{\includegraphics[width=1in,height=1.25in,clip,keepaspectratio]{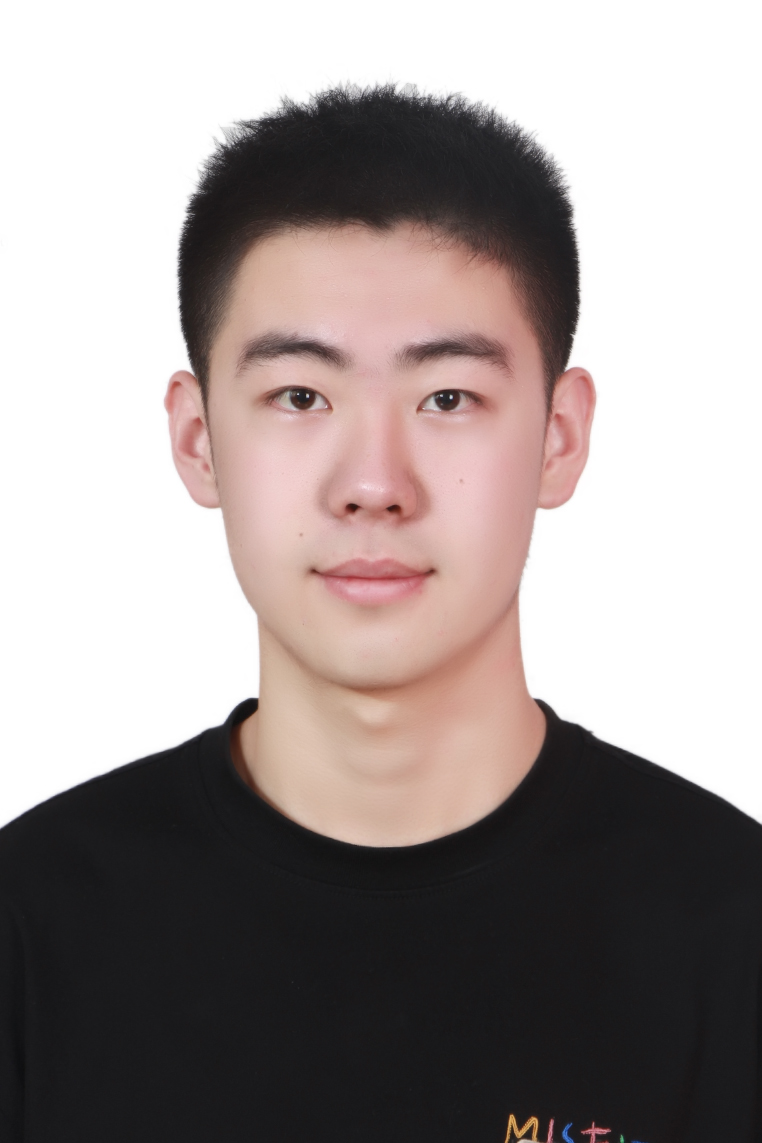}}]{Bowen Chai} received his B.E. degree from the School of Automation and Sensing, Shanghai Jiao Tong University. He is currently pursuing an M.S. degree at Shanghai Jiao Tong University under the supervision of Prof. Danping Zou in the Vision and Intelligent System Group (VISYS). During his undergraduate studies, he was a research intern in Prof. Yulun Zhang's laboratory. His research interests include video restoration, model compression, and 3D vision. He has published papers in international conferences, including AAAI and ICML.
\end{IEEEbiography}

\begin{IEEEbiography}[{\includegraphics[width=1in,height=1.25in,clip,keepaspectratio]{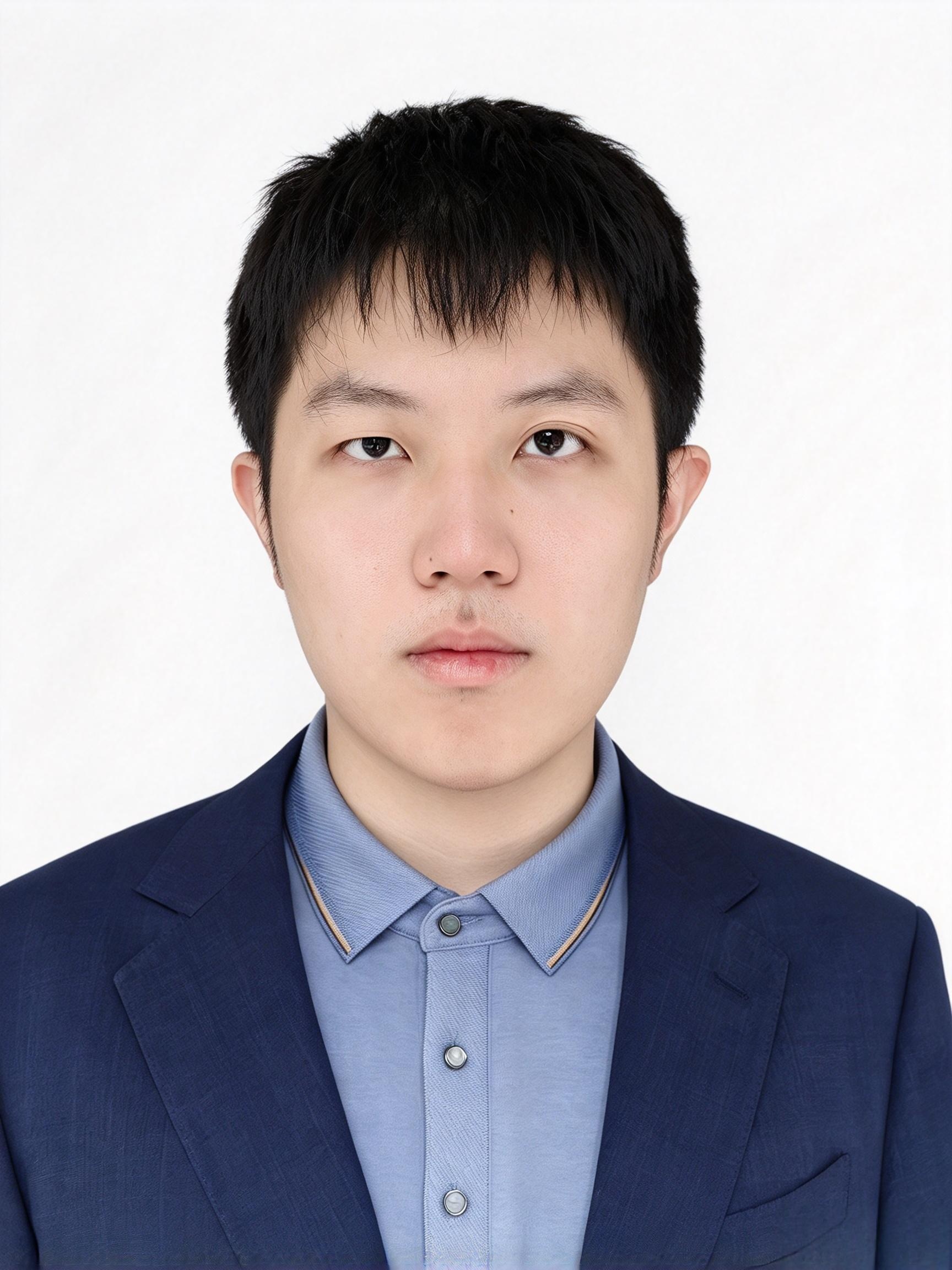}}]{Rongjun Gao}
is a B.S. student majoring in Electrical and Computer Engineering in Global College, Shanghai Jiao Tong University. During his undergraduate studies, he was a research intern in Prof. Yulun Zhang's laboratory. His research interests include image/video restoration and video generation.
\end{IEEEbiography}

\begin{IEEEbiography}[{\includegraphics[width=1in,height=1.25in,clip,keepaspectratio]{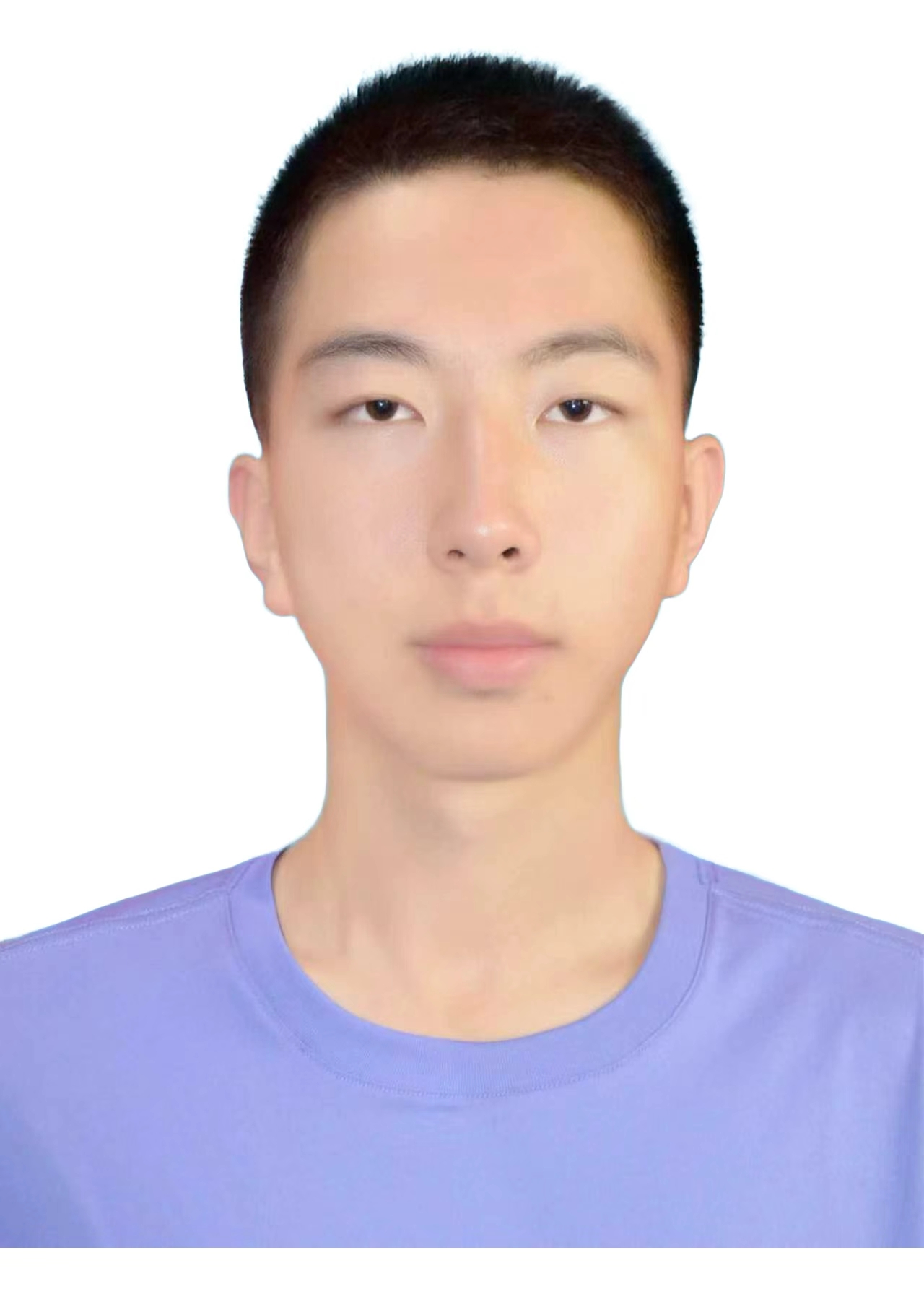}}]{Mingtao Nie}
is an undergraduate student majoring in Information Security at Shanghai Jiao Tong University, expected to receive his B.E. degree in 2027.
He is currently a research intern in Prof. Yulun Zhang's laboratory.
His research interests include video restoration, robot learning, and world models.
He has received several honors, including the Optiver Excellence Scholarship and the Finalist Award in the 2026 Mathematical Contest in Modeling (MCM).
\end{IEEEbiography}

\newpage

\begin{IEEEbiography}[{\includegraphics[width=1in,height=1.25in,clip,keepaspectratio]{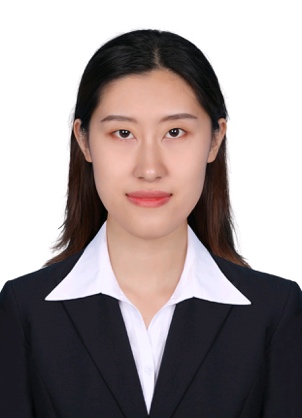}}]{Xi Li}
received the M.S. degree in software engineering from Beihang University, Beijing, China.
She is currently a researcher at Meituan, where she focuses on multimodal generation and understanding, including speech synthesis and content understanding algorithms.
Her research interests include generative models, multimodal learning, and post-training techniques for large models.
\end{IEEEbiography}

\begin{IEEEbiography}[{\includegraphics[width=1in,height=1.25in,clip,keepaspectratio]{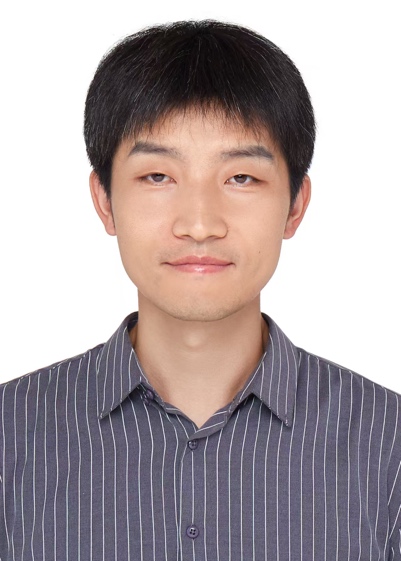}}]{Bingnan Duan}
received the B.S. degree in computer science and technology from Hebei GEO University in 2006.
He is currently a technical lead of the video architecture team at Meituan, where he leads the development of the company's audio/video infrastructure for live streaming and video-on-demand, supporting the technical needs of numerous business lines.
He has extensive experience in building Internet-scale audio/video systems and in stability engineering for large-scale systems.
\end{IEEEbiography}

\begin{IEEEbiography}[{\includegraphics[width=1in,height=1.25in,clip,keepaspectratio]{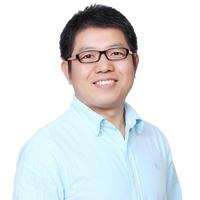}}]{Jianping Fang}
is currently the head of the Online Entertainment Technology Department at Meituan.
He has nearly ten years of experience in building distributed backend architectures for large-scale Internet systems.
\end{IEEEbiography}

\newpage

\begin{IEEEbiography}[{\includegraphics[width=1in,height=1.25in,clip,keepaspectratio]{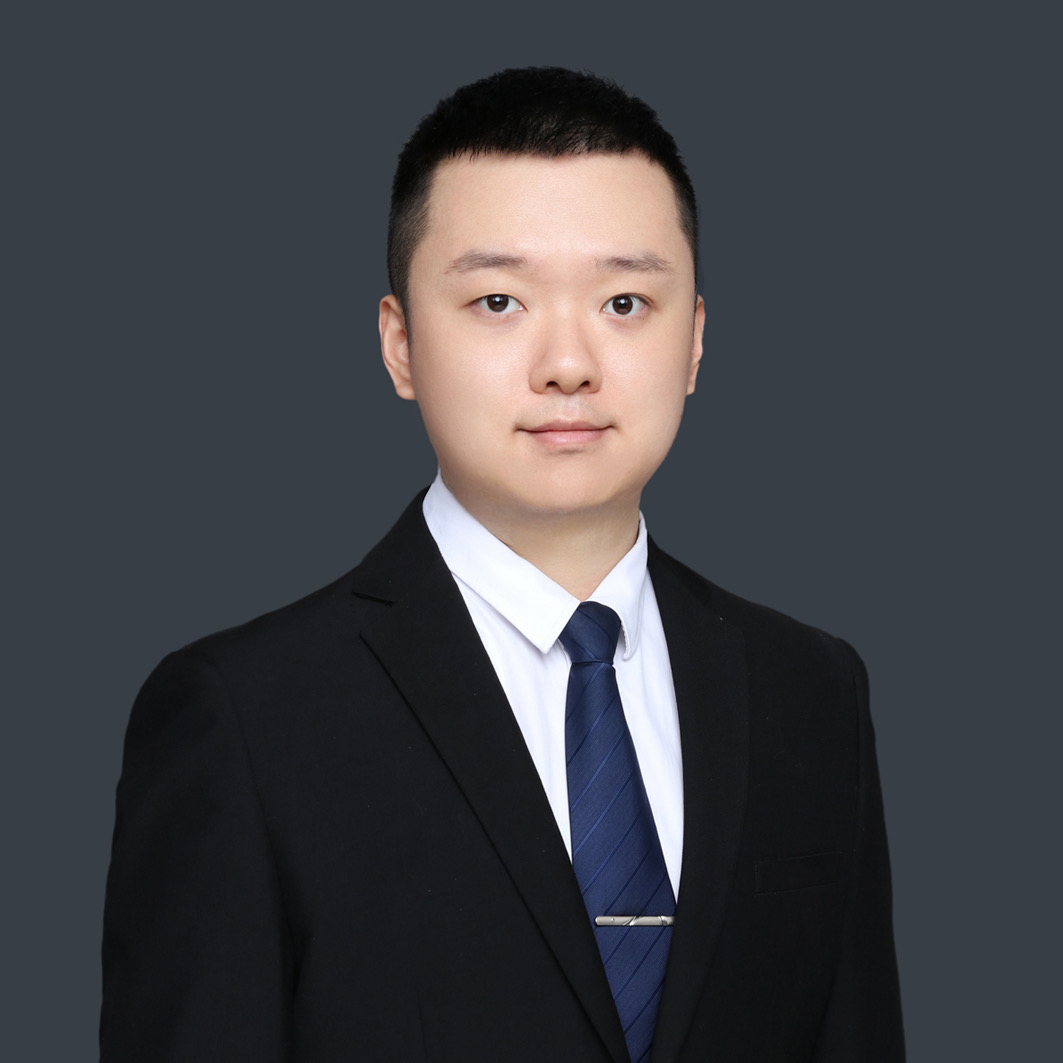}}]{Xiaohong Liu}
is an associate professor at the John Hopcroft Center for Computer Science, Shanghai Jiao Tong University. He received his Ph.D. from McMaster University, Canada. His research focuses on computer vision and multimedia, and he has published more than 80 papers.
He was named a Microsoft Research Asia StarTrack Scholar in 2024, was selected for the Shanghai Pujiang Program in 2022, and received the Chinese Government Award for Outstanding Self-Financed Students Abroad in 2021 and a Borealis AI Fellowship in 2020.
His work has received support from the young scientist programs of the National Natural Science Foundation of China and the Natural Science Foundation of Sichuan Province, as well as from industry. He serves as an associate editor of \emph{ACM Transactions on Multimedia Computing, Communications, and Applications} (TOMM). Website: \url{https://jhc.sjtu.edu.cn/~xiaohongliu/}.
\end{IEEEbiography}

\begin{IEEEbiography}[{\includegraphics[width=1in,height=1.25in,clip,keepaspectratio]{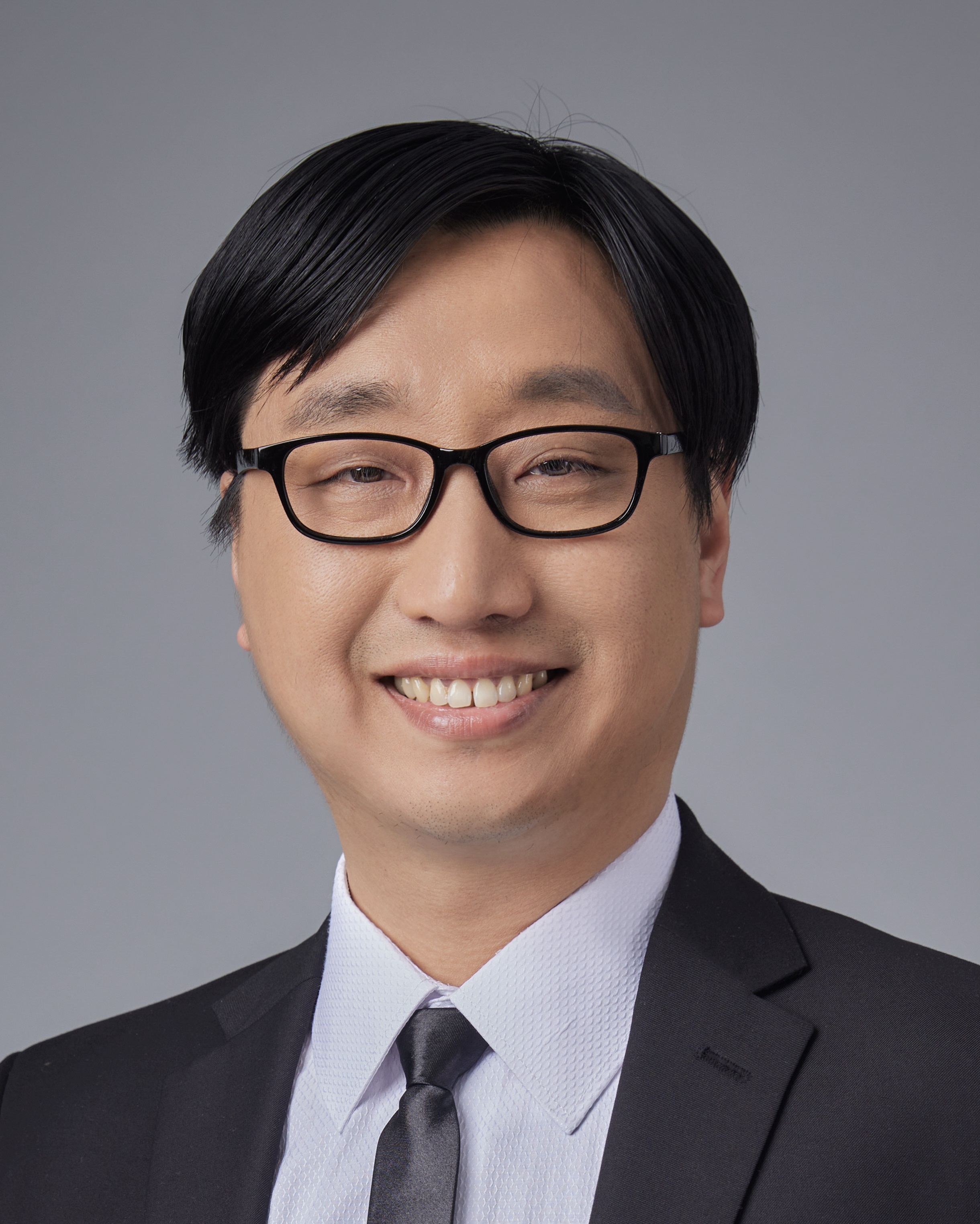}}]{Linghe Kong}
(Fellow, IEEE) received a B.Eng. in automation from Xidian University in 2005, a master's degree in telecommunications from Télécom SudParis in 2007, and a Ph.D. in computer science from Shanghai Jiao Tong University in 2013. He is a professor in the School of Computer Science at Shanghai Jiao Tong University.
He held postdoctoral appointments at Columbia University, McGill University, and the Singapore University of Technology and Design. His research interests include the Internet of Things, 5G, blockchain, and mobile computing.
\end{IEEEbiography}

\begin{IEEEbiography}[{\includegraphics[width=1in,height=1.25in,clip,keepaspectratio]{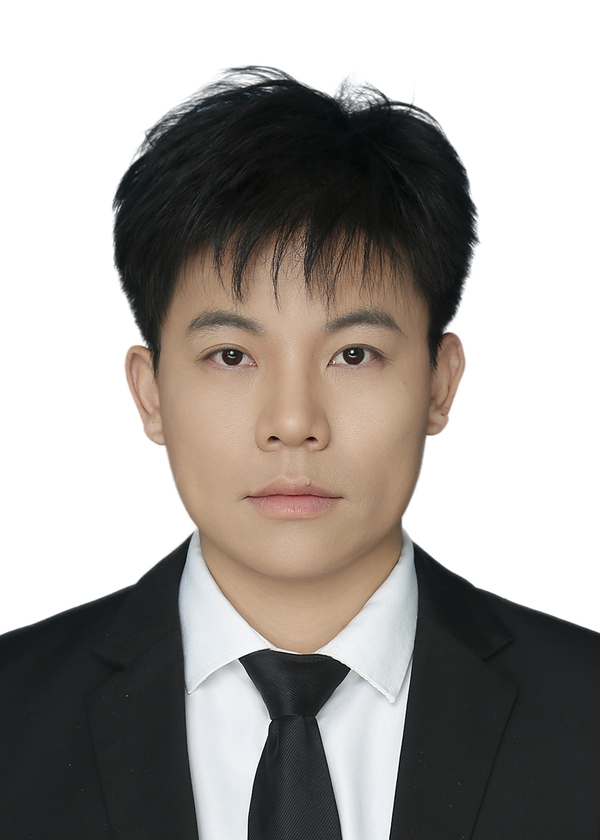}}]{Yulun Zhang} received a B.E. degree from the School of Electronic Engineering, Xidian University, China, in 2013, an M.E. degree from the Department of Automation, Tsinghua University, China, in 2017, and a Ph.D. degree from the Department of ECE, Northeastern University, USA, in 2021. He is an associate professor at Shanghai Jiao Tong University, Shanghai, China. He was a postdoctoral researcher at the Computer Vision Lab, ETH Zürich, Switzerland. His research interests include image/video restoration and synthesis, biomedical image analysis, model compression, multimodal computing, large language models, and computational imaging. He has served as an Area Chair for CVPR, ICCV, ECCV, NeurIPS, ICML, ICLR, IJCAI, and ACM MM, and as a Senior Program Committee (SPC) member for IJCAI and AAAI.
\end{IEEEbiography}

\end{document}